\documentclass[letterpaper]{article} 
\usepackage{aaai25}  
\usepackage{times}  
\usepackage{helvet}  
\usepackage{courier}  
\usepackage[hyphens]{url}  
\usepackage{graphicx} 
\usepackage{natbib}  
\usepackage{caption} 
\usepackage{algorithm}

\usepackage{graphicx}
\usepackage{amsmath}
\usepackage{amsthm}
\usepackage{booktabs}
\usepackage[noend]{algpseudocode}
\usepackage[switch]{lineno}
\usepackage{subcaption}
\usepackage{adjustbox}
\usepackage{multirow}
\usepackage{xcolor}
\usepackage{amssymb}
\usepackage{placeins}
\usepackage{sidecap}

\newtheorem{proposition}{Proposition}

\usepackage{newfloat}
\usepackage{listings}
\DeclareCaptionStyle{ruled}{labelfont=normalfont,labelsep=colon,strut=off} 
\floatstyle{ruled}
\newfloat{listing}{tb}{lst}{}
\floatname{listing}{Listing}
\title{The Distributional Reward Critic Framework for\\Reinforcement Learning Under Perturbed Rewards}
\author{
    Xi Chen,
    Zhihui Zhu,
    Andrew Perrault
}
\affiliations{
    The Ohio State Univeristy\\
    chen.10183@osu.edu, zhu.3440@osu.edu, perrault.17@osu.edu
}

\begin{document}

\maketitle

\begin{abstract}
The reward signal plays a central role in defining the desired behaviors of agents in reinforcement learning (RL). Rewards collected from realistic environments could be perturbed, corrupted, or noisy due to an adversary, sensor error, or because they come from subjective human feedback. Thus, it is important to construct agents that can learn under such rewards. Existing methodologies for this problem make strong assumptions, including that the perturbation is known in advance, clean rewards are accessible, or that the perturbation preserves the optimal policy. We study a new, more general, class of unknown perturbations, and introduce a distributional reward critic framework for estimating reward distributions and perturbations during training. Our proposed methods are compatible with any RL algorithm. Despite their increased generality, we show that they achieve comparable or better rewards than existing methods in a variety of environments, including those with clean rewards. Under the challenging and generalized perturbations we study, we win/tie the highest return in 44/48 tested settings (compared to 11/48 for the best baseline). Our results broaden and deepen our ability to perform RL in reward-perturbed environments. If necessary, please check the full paper (\url{https://arxiv.org/abs/2401.05710}), including the Appendix.
\end{abstract}

%
\begin{links}
    \link{Code}{https://github.com/cx441000319/DRC}
\end{links}

\section{Introduction}

The use of reward as an objective is a central feature of reinforcement learning (RL) that has been hypothesized to constitute a path to general intelligence~\cite{silver2021reward}. The reward is also the cause of a substantial amount of human effort associated with RL, from engineering to reduce difficulties caused by sparse, delayed, or misspecified rewards~\cite{ng1999policy,hadfield2017inverse,qian2023learning} to gathering large volumes of human-labeled rewards used for tuning large language models (LLMs)~\cite{ouyang2022training,bai2022training}. Thus, the ability of RL algorithms to tolerate noisy, perturbed, or corrupted rewards is of general interest~\cite{romoff2018reward,wang2020reinforcement,everitt2017reinforcement,moreno2006noisy,corazza2022reinforcement,zheng2014robust}.

\begin{table}[t]
  \setlength{\intextsep}{1pt}
  \centering
  \begin{adjustbox}{center}
  \resizebox{1.0\columnwidth}{!}{
  \fontsize{19}{1.2\baselineskip}\selectfont
  \renewcommand{\arraystretch}{1.8}
  \begin{tabular}{c | ccc | ccc | c}
    \toprule
     & \multicolumn{3}{c|}{Required Info} & \multicolumn{3}{c|}{Reward Estimator} & Perturbation \\
    \cmidrule(lr){2-4} \cmidrule(lr){5-7} \cmidrule(lr){8-8}
     Method & \multirow{2}{*}{$\tilde{R}$} & \multirow{2}{*}{$n_r$} & Reward & Discrete & Leverages & Reduces & Optimal-Policy- \\
     & & & Range & Rewards & $(s,a,s\prime)$ & Variance & Changing \\
    \midrule
    \multirow{2}{*}{Standard} & \multirow{2}{*}{\textcolor{blue}{$\times$}} & \multirow{2}{*}{\textcolor{blue}{$\times$}}  & \multirow{2}{*}{\textcolor{blue}{$\times$}} & \multirow{2}{*}{\textcolor{blue}{$\times$}} & \multirow{2}{*}{\textcolor{orange}{$\times$}} & \multirow{2}{*}{\textcolor{orange}{$\times$}} & \multirow{2}{*}{\textcolor{orange}{$\times$}} \\ 
    & & & & & & \\
    \multirow{2}{*}{SR\_W} & \multirow{2}{*}{\textcolor{orange}{$\checkmark$}} & \multirow{2}{*}{\textcolor{orange}{$\checkmark$}}  & \multirow{2}{*}{\textcolor{orange}{$\checkmark$}} & \multirow{2}{*}{\textcolor{orange}{$\checkmark$}} & \multirow{2}{*}
    {\textcolor{orange}{$\times$}} & \multirow{2}{*}{\textcolor{orange}{$\times$}}  & \multirow{2}{*}{\textcolor{blue}{$\checkmark$}} \\ 
    & & & & & & \\
    \multirow{2}{*}{SR} & \multirow{2}{*}{\textcolor{blue}{$\times$}} & \multirow{2}{*}{\textcolor{orange}{$\checkmark$}}  & \multirow{2}{*}{\textcolor{orange}{$\checkmark$}} & \multirow{2}{*}{\textcolor{orange}{$\checkmark$}} & \multirow{2}{*}{\textcolor{orange}{$\times$}} & \multirow{2}{*}{\textcolor{orange}{$\times$}} & \multirow{2}{*}{\textcolor{blue}{$\checkmark$}} \\ 
    & & & & & & \\
    \multirow{2}{*}{RE} & \multirow{2}{*}{\textcolor{blue}{$\times$}} & \multirow{2}{*}{\textcolor{blue}{$\times$}}  & \multirow{2}{*}{\textcolor{blue}{$\times$}} & \multirow{2}{*}{\textcolor{blue}{$\times$}} & \multirow{2}{*}{\textcolor{blue}{$\checkmark$}} & \multirow{2}{*}{\textcolor{blue}{$\checkmark$}} & \multirow{2}{*}{\textcolor{orange}{$\times$}} \\ 
    & & & & & &  \\
    \midrule
    \multirow{2}{*}{\textbf{DRC}} & \multirow{2}{*}{\textcolor{blue}{$\times$}} & \multirow{2}{*}{\textcolor{orange}{$\checkmark$}}  & \multirow{2}{*}{\textcolor{orange}{$\checkmark$}} & \multirow{2}{*}{\textcolor{blue}{$\times$}} & \multirow{2}{*}{\textcolor{blue}{$\checkmark$}} & \multirow{2}{*}{\textcolor{blue}{$\checkmark$}} & \multirow{2}{*}{\textcolor{blue}{$\checkmark$}} \\ 
    & & & & & &  \\
    \multirow{2}{*}{\textbf{GDRC}} & \multirow{2}{*}{\textcolor{blue}{$\times$}} & \multirow{2}{*}{\textcolor{blue}{$\times$}}  & \multirow{2}{*}{\textcolor{blue}{$\times$}} & \multirow{2}{*}{\textcolor{blue}{$\times$}} & \multirow{2}{*}{\textcolor{blue}{$\checkmark$}} & \multirow{2}{*}{\textcolor{blue}{$\checkmark$}} & \multirow{2}{*}{\textcolor{blue}{$\checkmark$}} \\ 
    & & & & & & \\
    \bottomrule
  \end{tabular}
  }
  \end{adjustbox}
  \caption{The methods proposed in this paper (in bold) have key advantages relative to past methods, which are limited by the amount of Required Info about the perturbation that is needed, the structural properties of the environment such as discrete rewards, and assumptions on the effect of the perturbation on the optimal policy. Leveraging $(s,a,s')$ for reward estimation and variance reduction (of the estimated reward vs. true reward) are advantages of methods that construct explicit perturbation models. Blue is an advantage, and orange is a limitation.}
  \label{table:method_comp}
\end{table}

Different reward correction methods have been proposed to enhance the performance of RL algorithms under reward perturbations by estimating the expectation of perturbed rewards~\cite{romoff2018reward} or learning the perturbation model~\cite{wang2020reinforcement}. However, these existing approaches rely on strong assumptions. For instance, the Reward Estimation (RE) method~\cite{romoff2018reward} assumes that the perturbation does not impact the optimal policy, a condition satisfied in limited cases, such as when the reward undergoes a positive affine transformation. On the other hand, the Surrogate Reward (SR) method~\cite{wang2020reinforcement} can handle perturbations beyond affine transformations, but it presupposes that there are a finite number of possible reward values and a specific perturbation structure (see Section~\ref{related_work} for a more detailed discussion). Thus, the methods are individually limited in where they can be applied, and no method exists for certain problem-perturbation combinations (i.e., continuous reward environments where the optimal policy is altered). Indeed, it is not clear that such a method can be constructed due to the structural difference between perturbations on a discrete-valued reward (e.g., binary success/failure) and those on a continuous-valued reward (e.g., velocity).

We show that an adaptive reward modeling approach, with appropriate design choices, can be used to unify the two scenarios, creating a universal framework for learning under perturbed rewards. We propose a family of \emph{distributional} reward critic methods that redefine the problem of reward estimation as a classification task and infer the unknown perturbation structures without prior knowledge. Our contributions (Table~\ref{table:method_comp}) are summarized in three parts as follows.

\paragraph{The Distributional Reward Critic (DRC) Framework} To recover from a changing optimal policy on the perturbed rewards, we begin by turning the reward regression into a classification problem using ordinal cross-entropy for estimating the reward distribution under the assumption of knowing the structure of the perturbation, which we call Generalized Confusion Matrix (GCM) perturbations. GCM perturbations generalize past reward perturbation distributions studied in the literature and allow for perturbations on both discrete- and continuous-valued rewards that alter the optimal policy, and, in addition, can approximate any distribution. However, DRC is limited by its assumed foreknowledge of the structural properties of the GCM perturbation.


\paragraph{General Distributional Reward Critic (GDRC)} We devise a General Distributional Reward Critic (GDRC) method to simultaneously estimate the structure of the GCM during training. GDRC works by training an ensemble of DRCs and using the metrics from the training process to intelligently select the most credible DRC. We show in simulation that GDRC is effective at inferring the structure of unknown GCM perturbations.

\paragraph{Experimental Evaluation} Using an array of tasks and reward perturbation distributions, we compare DRC and GDRC to state-of-the-art methods for perturbed reward reinforcement learning. Under GCM perturbations, we win/tie ($95\%$ of the winner) the highest return in 44/48 sets (compared to 11/48 for the best baseline). Even under continuous perturbations with strict assumptions, we find that our methods are on par with existing ones that leverage strong assumptions about perturbations. Together, our results imply that GDRC is a strong tool for learning under broad perturbations. Moreover, we set up a series of studies to verify every decision we make when developing our methods.

\section{Related Work}
\label{related_work}

Reward perturbations in RL have been extensively studied~\cite{romoff2018reward,wang2020reinforcement,rakhsha2020policy,pattanaik2017robust,pinto2017robust,choromanski2020provably,NEURIPS2023_29ef811e,corazza2022reinforcement,zhuang2021no,hu2022distributional,ring2011delusion,hutter2005universal,amodei2016concrete}. We describe two of the most related approaches  \cite{romoff2018reward,wang2020reinforcement} in detail.

\paragraph{Reward Prediction By The Reward Estimation (RE) Method}
\label{related_work_continuous_noise}
\cite{romoff2018reward} focus on variance reduction in the case of continuous perturbations that increase reward variance but do not change the optimal policy. This occurs, for example, when the reward perturbation applies a positive affine transformation, i.e., $\mathbb{E}[\tilde{R}(s,a)] = \omega_0 R(s,a) + \omega_1$ and $\omega_0 > 0$. In this setting, the perturbations can slow down the training and even destroy it when the critic is not trained with enough samples. They propose the Reward Estimation~(RE) method by introducing a reward critic to predict rewards $\tilde{R}(s,a)$ that aims to reduce the variance.

\paragraph{Perturbation Modeling By The Surrogate Reward (SR) Method}
\label{related_work_confusion_matrix_noise}
\cite{wang2020reinforcement} study the setting where the rewards are discrete and perturbed by a confusion matrix $C$, where $C(i,j)$ represents the probability that reward $R_i$ is perturbed into $R_j$. Wang et al.~propose the Surrogate Reward~(SR) method by inverting the confusion matrix $\hat{R}=C^{-1}\cdot R$, where vectorized $R$ and $\hat{R}$ represent clean and predicted rewards, replacing each observed reward $R_i$ with an unbiased estimate of the true reward $\hat{R_i}$. As their method assumes $C$ to be estimated separately, they cannot leverage the structure of the reward signal in state-action-state space (i.e., that similar state-action pairs may have similar rewards.

In brief, our method addresses the limitations of both RE and SR, by handling a broader class of perturbations without assuming foreknowledge of its structure.

\paragraph{Distributional RL}
\label{related_work_distributional_RL}
Our approaches are inspired by distributional RL~\cite{dabney2018distributional,bellemare2017distributional,dabney2018implicit,rowland2018analysis}, where the value function is modeled distributionally. We find that fixed-width regression ~\cite{bellemare2017distributional} is more appropriate than fixed-ratio for reward modeling for technical reasons. However, unlike standard fixed-width methods, we control the size and location of the fixed-width bins adaptively (which are standardly treated as hyperparameters in distributional RL). The result is a generic method that does not require tuning hyperparameters across tasks or perturbations, which is commonly a weakness of fixed-width compared to fixed-ratio methods.

\section{Problem Statement}
\label{problem_statement}

In Section~\ref{problem_statement:sub1}, we describe the standard extension of MDPs to perturbed rewards. In Section~\ref{problem_statement:sub2}, we introduce generalized confusion matrix (GCM) perturbations and show their significance and universality.


\subsection{Perturbed Reward MDP}
\label{problem_statement:sub1}

Let $\left\langle S, A, R, P, \gamma, \beta \right\rangle$ be a Markov Decision Process (MDP)~\cite{puterman2014markov}, where $S$ is the state space, $A$ is the action space, $R: S \times A  \times S \rightarrow \mathbb{R}$ is the reward function, $P: S \times A \rightarrow \Delta S$ is the transition function, $\beta \in \Delta S$ is the initial state distribution, and $\gamma \in [0,1]$ is the discount factor. We denote the state at timestep $t$ as $s_t$, the action as $a_t$, and the reward as $r_t = R(s_t, a_t, s_{t+1})$.

We define a \emph{perturbed reward MDP} of the form $\left\langle S, A, R, \tilde{R}, P, \gamma, d_0 \right\rangle$. The agent perceives perturbed rewards from $\tilde{R}$ instead of true rewards from $R$.
Following the prior work~\cite{romoff2018reward,wang2020reinforcement},  
we assume that the perturbed reward is random and that its distribution depends on the true reward, i.e., $\tilde{r}_t \sim \tilde{R}(R(s_t,a_t,s_{t+1}))$. The objective is the same as in a standard MDP: we seek a policy $\pi: S \rightarrow \Delta A$ that maximizes the return, i.e., $\pi^* \in \arg\max_\pi \mathbb{E}_\pi[\sum_{t=0}^{\infty }\gamma^{t} R(s_t,a_t,s_{s+1})]$, but we only receive perturbed rewards.

\subsection{Generalized Confusion Matrix (GCM) Perturbations}
\label{problem_statement:sub2}

\begin{figure}[t]
  \centering
  \includegraphics[width=0.474\textwidth]{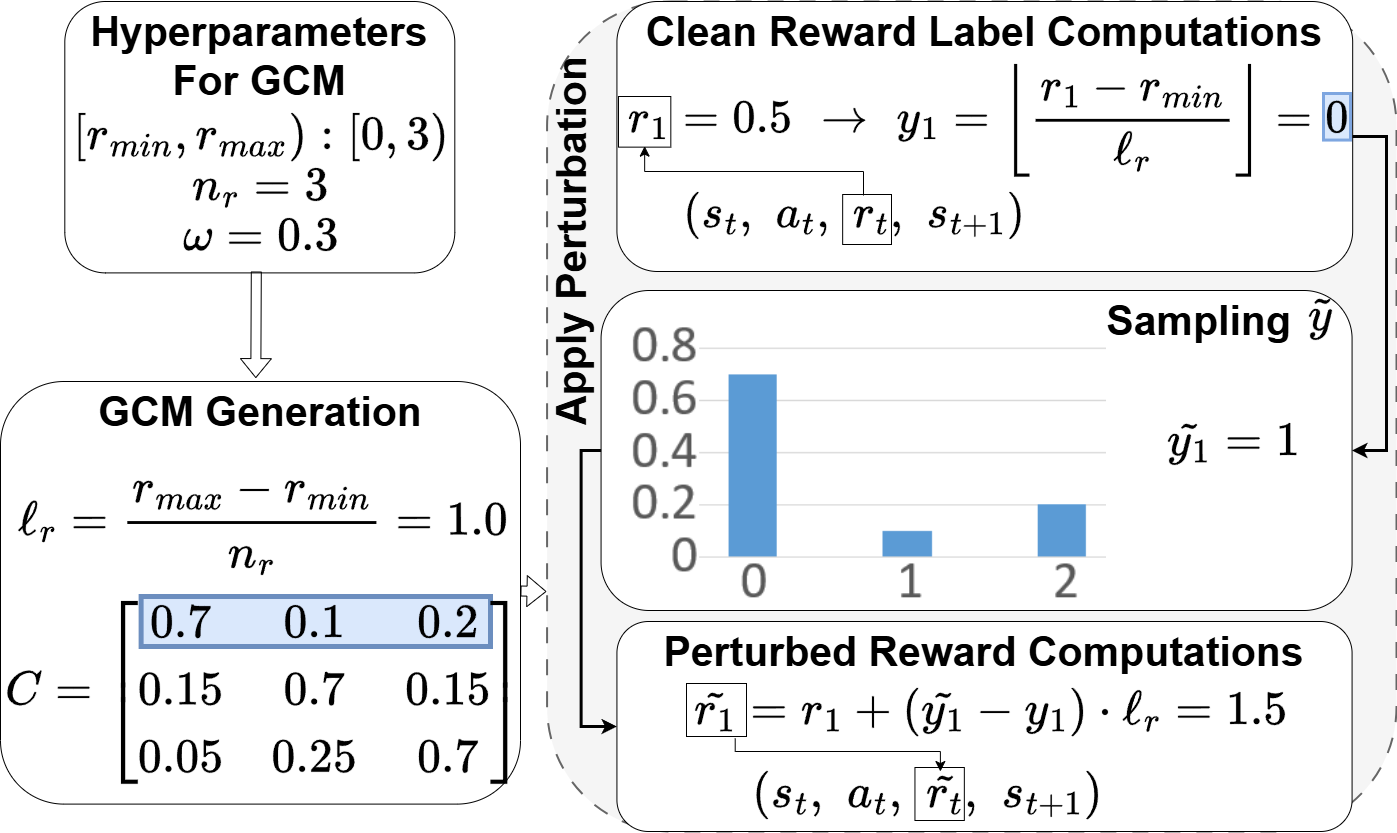}
  \caption{An example of GCM perturbation.}
  \label{fig:concrete_example}
\end{figure}

To effectively handle perturbed rewards, we rely on an appropriate model to capture the perturbations. The Surrogate Reward (SR) method~\cite{wang2020reinforcement} models the reward perturbation by perturbing discrete rewards with a confusion matrix $C$, where $C(i,j)$ represents the probability that reward $R_i$ is perturbed into $R_j$. We develop a perturbation model that preserves the continuous nature of the reward space. Specifically, we propose a generalized confusion matrix (GCM) perturbation model and provide a motivating example in Fig.~\ref{fig:concrete_example}. The parameters of a GCM are the reward interval length $\ell_r=(r_\textrm{max}-r_\textrm{min})/n_r$, the minimum and maximum rewards $r_\textrm{min}$ and $r_\textrm{max}$, and the associated confusion matrix $C$. To sample from a GCM with input reward $r_t$, we convert $r_t$ into its reward label $y_t$ with $y_t=\lfloor (r_t-r_{min})/l_r \rfloor$. Then, we use the probability distribution $C(y_t,\cdot)$ to sample perturbed label $\tilde{y_t}$ and shift $r_t$ by the signed distance between $y_t$ and $\tilde{y}_t$, i.e., $\tilde{r}_t = r_t + \ell_r(\tilde{y}_t - y_t)$. 

We study GCM perturbations for three reasons. First, GCM perturbations naturally handle continuous rewards without sparsifying the perturbed rewards (i.e., the perturbed rewards are themselves continuous). Second, GCM can represent perturbations that can change the optimal policy, i.e., the optimal policy for the perturbed rewards is different than the optimal policy for the clean rewards. This expressivity comes from their ability to represent perturbations with arbitrary probability density functions (PDF) for each reward interval---each row is a perturbation PDF. Third, GCMs can approximate any continuous perturbation with a bounded error that diminishes as reward interval number $n_r$ as shown in Proposition~\ref{thm:0}, whose proof is in Appendix~\ref{appendix:a}.
\begin{proposition}
\label{thm:0}
Consider continuous perturbations that for each reward $r\in[r_{\min},r_{\max})$, it can be perturbed to $\bar r \in[r_{\min},r_{\max})$. Our GCM represents $\bar r$ with $\tilde r$ that satisfies $|\tilde r - \bar r|\le \frac{r_{\max} - r_{\min}}{n_r}$.
\end{proposition}
We now turn our attention to developing a method that can learn effectively under a GCM-perturbed reward signal.

\section{The Distributional Reward Critic Architecture}
\label{method}
We begin in Sec.~\ref{2C} with the simpler case where the GCM interval length, number of intervals, and minimum and maximum are all known (i.e., all the GCM parameters except for the confusion matrix $C$). Then, we study the more general case where these parameters must be inferred during training in Sec.~\ref{G2C}.

\subsection{Distributional Reward Critic~(DRC)}
\label{2C}

To learn effectively under a GCM perturbation, we want to introduce a network to model the reward distribution for each state-action-state. We view the reward estimation as a classification problem by turning the reward range $[r_{\textrm{min}},r_{\textrm{max}}]$ into $n_r$ intervals. Therefore, the problem we want to resolve could be expressed as: given input $(s_t,a_t, s_{t+1},\tilde{r}_t)$, predict the distribution of label $\tilde{y}_t=\textrm{floor}\big(\left(\tilde r_t - r_{\textrm{min}}\right)/\ell_r\big)$ where $\ell_r=(r_{\textrm{max}} - r_{\textrm{min}})/n_r$ is the reward interval length. We do so by training a classification model $\hat{R_\theta}: S \times A \times S \rightarrow \Delta Y$ where $Y=(0,1,\ldots,n_r-1)$.

When we train a classification model like $\hat{R_\theta}$, the most common loss is cross-entropy (CE). However, CE discards information about the order of reward labels. This is especially critical in RL because, as the reward distribution shifts during training, it is critical to be able to quickly identify that certain rare reward values are superior to those seen thus far. When CE is used as a loss function, $\hat{R_\theta}$ tends to classify unseen samples as belonging to the most common class that is seen thus far (as the loss incurred by predicting any incorrect class is equivalent). We thus propose ordinal cross-entropy (OCE) instead\footnotemark: 
\begin{equation}
\begin{split}
OCE=\sum_t(1+\frac{\left| \hat{y}_t - \tilde{y}_t \right|}{n_r-1})\cdot H(\tilde{y}_t, \hat{R}_\theta (s_t,a_t,s_{t+1})), \\
 \text{where}~\hat{y}_t = {\arg\max}_{y \in Y} \hat{R}_\theta(s_t,a_t,s_{t+1})
\end{split} 
\end{equation}

\begin{figure}[tb]
    \centering
    \includegraphics[width=0.474\textwidth]{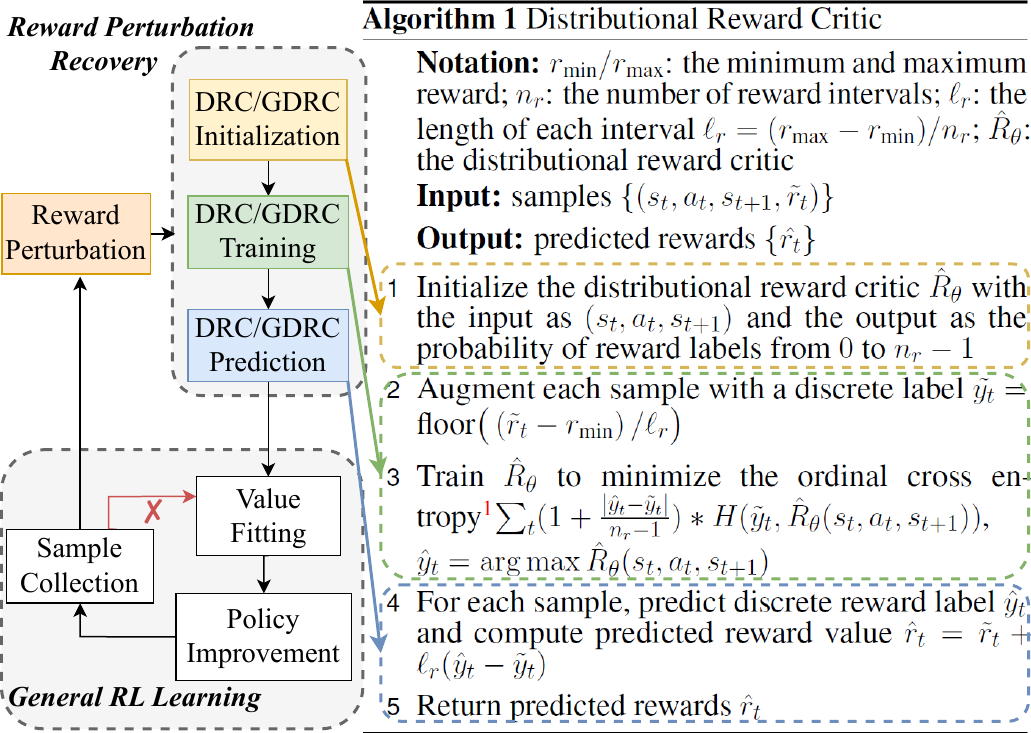}
    \caption{The pipeline of the whole process and the pseudocode of distributional reward methods.}
    \label{fig:pipe2}
\end{figure}

Using OCE as the loss function, the order among rewards is preserved naturally when we are in the discrete label space. In addition, when the observed reward label $\tilde{y}_t$ is far from the predicted reward label $\hat{y}_t$, the cross-entropy term $H(\tilde{y}_t, \hat{R}_\theta (s_t,a_t,s_{t+1}))$ will be assigned more weight based on their distance. Therefore, misclassified rare samples can receive a large weight during training if the distance is large.

We propose our Distributional Reward Critic~(DRC) method in Fig.~\ref{fig:pipe2}. DRC estimates the reward distribution online during RL training using OCE as the loss. The rewards that are received by the RL algorithm are replaced with the current predictions of the reward critic after they are updated based on the rewards observed for the current epoch. In experiments, we see that DRC is an effective approach across environments and that the use of OCE rather than CE is a critical choice.

However, DRC will only be practically applicable in environments with a known structure of GCM. Next, we study the simultaneous estimation of all GCM parameters.

\footnotetext{With slight abuse of notation, $H(\tilde{y}_t, \hat{R}_\theta (s_t,a_t,s_{t+1}))$ denotes the cross entropy between $\hat{R}_\theta (s_t,a_t,s_{t+1})$ and the distribution with all zero probabilities except for the $\tilde{y}_t$-th being 1 (i.e., the one-hot vector version of $\tilde y_t$).}

\subsection{General Distributional Reward Critic~(GDRC)}
\label{G2C}
In this section, we first study the influence of the discretization parameter $n_o$ of DRC on the reward prediction error, an unobservable metric termed Reconstruction Error, which provides insights into what $n_o$ is preferred. Then, we study how to use the observable ordinary cross-entropy loss to guide the selection of $n_o$. At last, we show how to estimate the reward range online while allowing for reward distribution shifts during training.


\subsubsection{Impact Of $n_o$ On Reconstruction Error For DRC}
\label{re_error}


Depending on the number of bins $n_o$ used in the reward critic, it may be impossible to recover the true reward even in the infinite sample case. We define \emph{Reconstruction Error} as $\textsc{Error}_r(\hat{R}_\theta) = \frac{1}{|T|} \sum_{t \in T} | \hat{r}_t - r_t |$, where $T$ represents the number of samples. Reconstruction Error prevents exact recovery of the reward when $n_o \neq a \cdot n_r (a\in \mathbb{Z}^+)$ as there is now an irreducible source of error caused by the misalignment of the intervals in the reward critic compared to the perturbation.

\begin{figure}[tb]
  \centering
  \begin{subfigure}[b]{0.23\textwidth}
    \includegraphics[width=\textwidth]{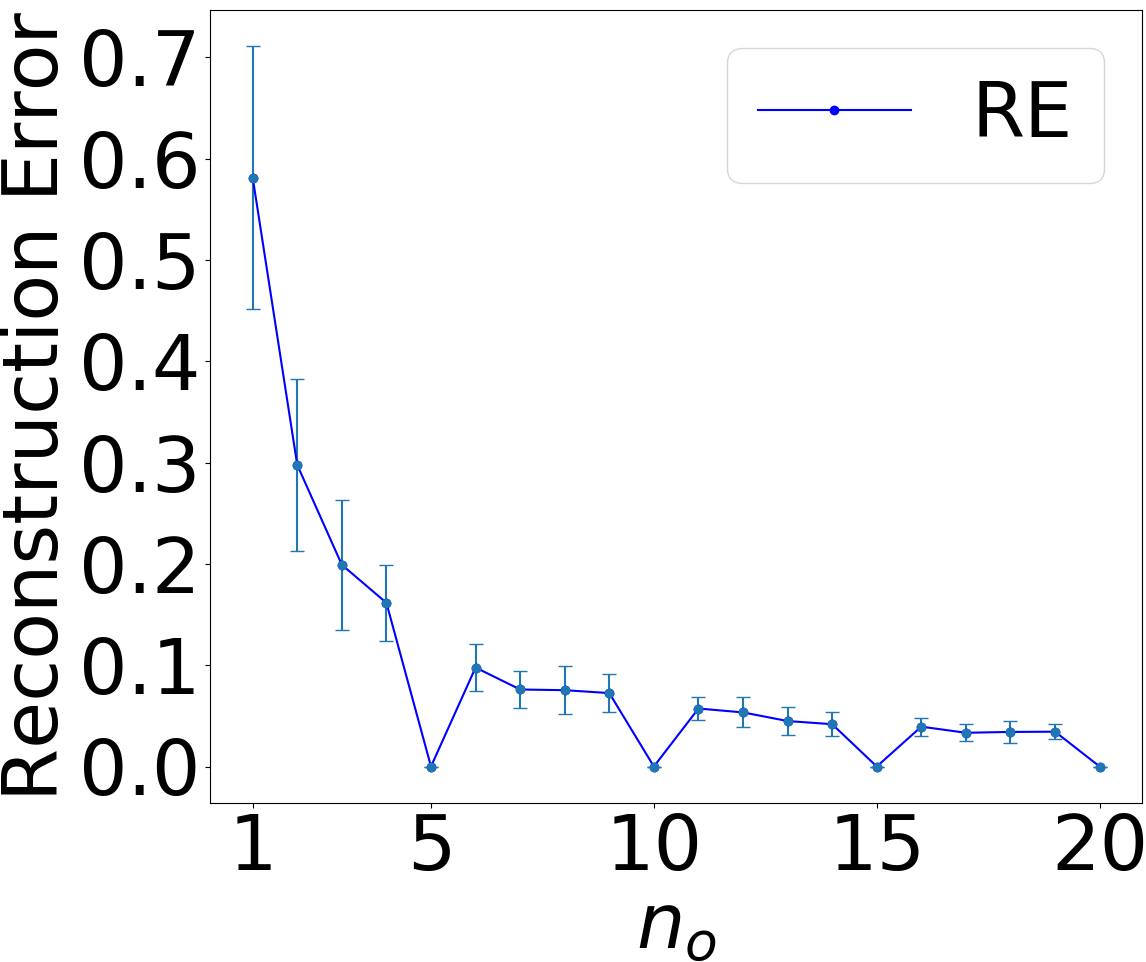}
    \caption{The reconstruction error initially decreases as $n_o$ increases, reaches 0 at $n_o=n_r$, and then oscillates.}
    \label{fig:r_error}
  \end{subfigure}
  \hfill
  \begin{subfigure}[b]{0.23\textwidth}
    \includegraphics[width=\textwidth]{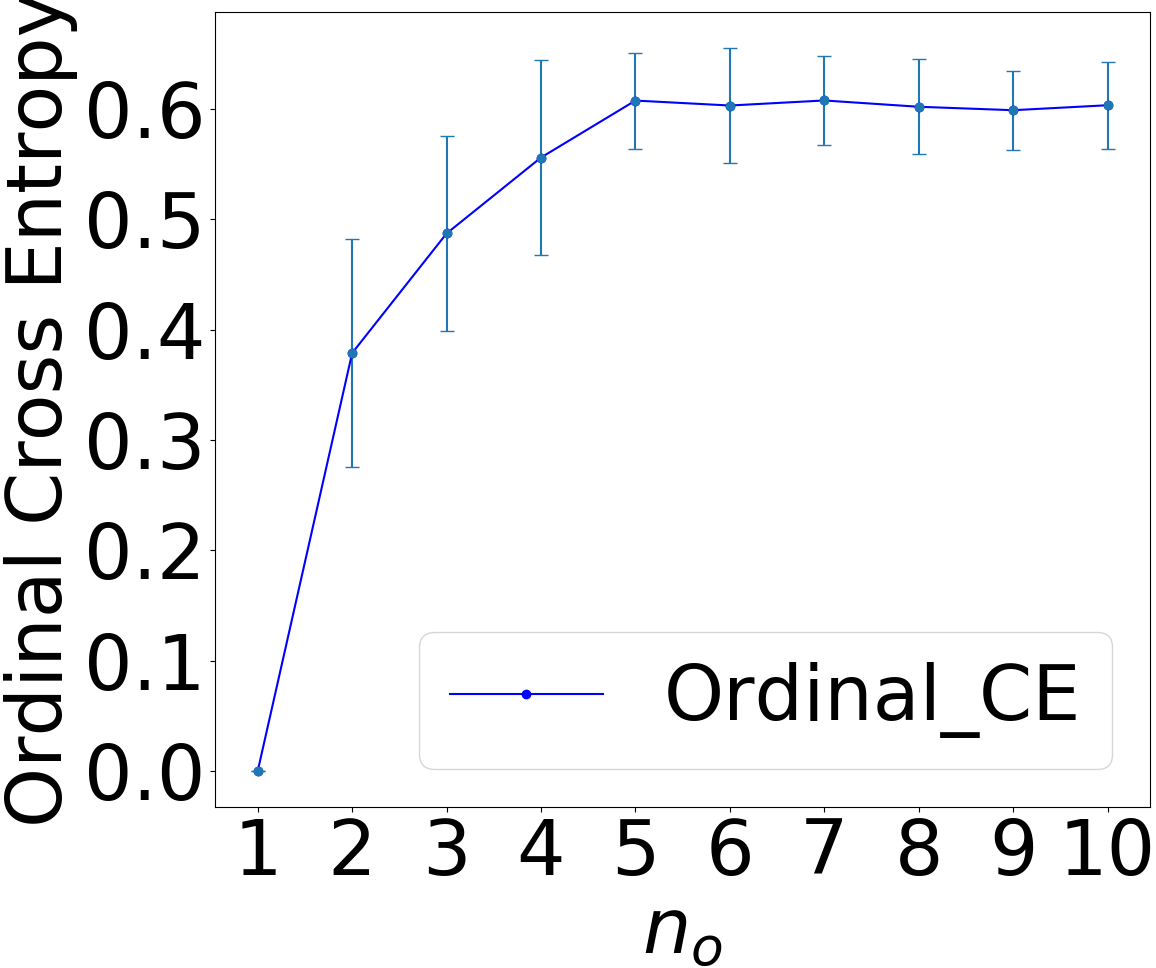}
    \caption{The minimum ordinal cross-entropy of the reward critic increases as $n_o$ increases until $n_o$ reaches $n_r$.}
    \label{fig:kl}
  \end{subfigure}
  \caption{Illustration of reconstruction error and cross entropy as $n_o$ varies in simulation environments where $n_r=5$.}
  \label{fig:side_by_side}
\end{figure}

To illustrate Reconstruction Error intuitively, we divide the range $[r_\textrm{min}, r_\textrm{max}]$ into $n_o$ intervals of length $\ell_o$. Given an observed reward $\tilde r = r+ \ell_r(\tilde{y} - y)$ perturbed from $r$, we compute its predicted reward $\hat{r} = \tilde{r} + \ell_o (\hat{y}_o - \tilde{y}_o)$, where $\hat{y}_o = \arg\max_{y_o \in Y_o} \hat{R_\theta}(s,a, s')$. Even if $\hat R_\theta$ is sufficiently expressive and trained well, $\hat r$ is not a correct prediction of $r$ due to the misalignment between $\ell_r 
 (\tilde{y} - y)$ and $\ell_o (\hat{y}_o - \tilde{y}_o)$. 
 In general, when $n_o< n_r$, Reconstruction Error becomes more pronounced as $n_o$ decreases because of the large granularity of $\ell_o$. When $n_o>n_r$, Reconstruction Error still exists except for the case that $n_o$ is a multiple of $n_r$, but is less significant. Fig.~\ref{fig:r_error} shows the impact $n_o$ on Reconstruction Error when a large number of samples are available for training. A detailed discussion is in Appendix~\ref{appendix:b}.
 
With infinite samples, a large $n_o$ is preferred to achieve a small Reconstruction Error. Without this condition, there is a tradeoff---a larger $n_o$ leads to more overfitting because of limited samples, but a small $n_o$ results in more Reconstruction Error. We aim to set $n_o=n_r$ as it achieves zero Reconstruction Error and requires the least samples, and we show the ordinal cross-entropy is an accessible metric to help infer it in the next part.

\subsubsection{Leveraging The Ordinal Cross-Entropy To Select $n_o$}
\label{KL}

During training, we can view the ordinal cross-entropy (OCE) loss of the distributional reward critic. We turn our attention to the dynamics of this loss as $n_o$ changes, and we will show that it can be used to estimate $n_r$.

We begin with a motivating example of intuition. For simplicity, we study the relation between cross-entropy (CE) and $n_o$ because the best prediction of the reward critic would be the reward label distribution using CE, and that relation also applies to OCE.

For example, $n_r=3$, $[r_{\textrm{min}},r_{\textrm{max}})=[0,3)$, and we have many samples $(s_t,a_t, s_{t+1},\tilde{r}_t)$ whose true rewards are all the same value $r\in[0,1)$. We define the probability of their perturbed rewards $\tilde{r}$ as $\mathbb{P}(\tilde{r})$, where $\tilde{r}=r+m$, $m\in(0,1,2)$ and $\mathbb{P}(\tilde{r})=C(0,\cdot)$. The probability of perturbed reward labels discretized by $n_r$ intervals is denoted as $q$, where $q_m=\mathbb{P}(\tilde r=r+m)$. Then we use $n_o$ intervals to fit $q$, whose reward label distribution is represented by $p$. Here, the question turns into how the entropy $H(p)$ changes as $n_o$ changes. There are two circumstances shown in Fig.~\ref{fig:k_m}:

\begin{figure}[tb]
  \centering
  \begin{subfigure}[b]{0.23\textwidth}
    \includegraphics[width=\textwidth]{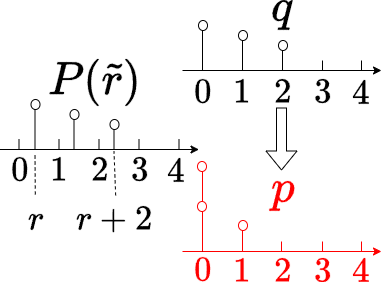}
    \caption{How $p$ fits $q$ when $n_o<n_r$}
    \label{fig:k<=m}
  \end{subfigure}
  \hfill
  \begin{subfigure}[b]{0.23\textwidth}
    \includegraphics[width=\textwidth]{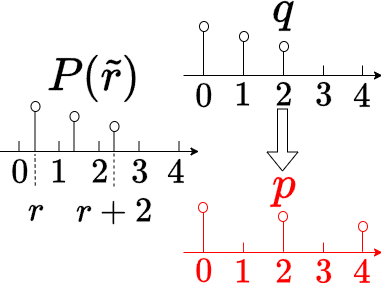}
    \caption{How $p$ fits $q$ when $n_o\geqslant n_r$}
    \label{fig:k>m}
  \end{subfigure}
  \caption{Illustration of cross-entropy for different $n_o$.}
  \label{fig:k_m}
\end{figure}

\begin{itemize}
    \item When $n_o<n_r$, more $q_m$ get combined together for smaller $n_o$, resulting in a smaller entropy $H(p) = -\sum_{k=1}^{n_o} p_k \log p_k$. For instance, combing $q_0, q_1 >0$ together leads to a reduction as $-(q_0+q_1)\log(q_0+q_1) < -(q_0\log q_0+q_1\log q_1)$.
    \item When $n_o>n_r$, there are $n_r$ dimensions with non-zero probabilities and $n_o-n_r$ dimensions with zero probabilities. Hence, $H(p) = -\sum_{k=1}^{n_o} p_k \log p_k = -\sum_{m=1}^{n_r} q_m \log q_m$ remains constant with respect to $n_o$
\end{itemize}

The combination and the separation of impulses with CE are also the case for OCE. Therefore, our intuition is that OCE also increases first when $n_o<n_r$ and then stays still when $n_o\geq n_r$. We verify the intuition by running a simple simulation test where we use a large number of samples to predict the perturbed reward label distribution $q$ using OCE in the cases $n_r=5$ and $\omega=0.1$. The result for each $n_o$ is averaged over 20 trails with different perturbed reward label distributions. As shown in Fig.~\ref{fig:kl}, the minimum OCE first increases when $n_o<n_r$ until it reaches $n_r$, which verifies our intuition above.

We run experiments using our Distributional Reward Critic (DRC) method with different $n_o$ to confirm the relation between OCE and $n_o$ experimentally. According to the results in Fig.~\ref{fig:kl_n}, we can clearly tell with the same TotalEnvInteracts (training steps), OCE first increases as $n_o$ increases and then oscillates when $n_o\geq n_r$.

Based on our analysis, we conclude there is a significant relation between OCE and $n_o$: OCE first increases as $n_o$ increases and then oscillates when $n_o\geq n_r$. As we aim to set $n_o$ to be $n_r$ to achieve the smallest Reconstruction Error, we propose our General Distributional Reward Critic~(GDRC) method, which trains an ensemble of reward critics $\left\{\hat{R}^{(n_o)}_\theta\right\}$ with uniform perturbation discretizations $n_o \in N_o$ to identify $n_r$ by referring to ordinal cross-entropy. We use these reward critics to vote on where the increasing rate of ordinal cross-entropy starts increasing. We use the critic who has received the most votes with a discount factor for reward prediction in each epoch.

\subsubsection{Handling The Unknown Reward Range}
\label{unknown_range}

For the case of an unknown reward range and an unknown number of intervals, we use the GDRC from the previous part plus an addition that updates the intervals based on the observed rewards seen in the latest 20 epochs. We create variables $r_\textrm{emin}$ and $r_\textrm{emax}$ to store the 1\% percentile and 99\% of the observed rewards across the samples in the latest 20 epochs as the minimum and maximum rewards, which excludes the possible influence of outlier perturbed rewards because of long-tail perturbations. This choice turns out to be important---keeping samples for too long slows training under reward distribution shift. We study the impact of this choice in Sec.~\ref{exp:moving_windown}.

We provide pseudocode for GDRC in Appendix~\ref{appendix:c}.

\makeatletter
\@addtoreset{theorem}{section}
\makeatother

\makeatletter
\@addtoreset{proposition}{section}
\makeatother

\section{Experimental Results}
\label{experiement}

In this section, we demonstrate that DRC and GDRC methods outperform existing approaches by attaining higher clean rewards and exhibiting broader applicability. Sec.~\ref{exp_setup} introduces the setups first. In Sec.~\ref{eval_cm} and \ref{eval_continuous}, we experiment under Generalized Confusion Matrix (GCM) and continuous perturbations. The influence of $n_o$ on Reconstruction Error and ordinal cross-entropy is substantiated in Sec.~\ref{eval_g2C}. In Sec.~\ref{exp:oce} and \ref{exp:moving_windown}, we do ablation studies to the critical decisions while developing DRC and GDRC. We include hyperparameters and training curves in Appendix~\ref{appendix:f}.

\subsection{Experimental Setup}
\label{exp_setup}

\paragraph{Algorithms} The methods introduced for perturbed rewards in this paper and previous work can be applied to any RL algorithm. Thus, we compare all methods as applied to some popular algorithms such as PPO~\cite{schulman2017proximal}, DDPG~\cite{lillicrap2015continuous}, and DQN~\cite{mnih2013playing}, covering on-policy and off-policy algorithms. The baseline methods other than the original algorithms are the Reward Estimation (RE)~\cite{romoff2018reward} method and Surrogate Reward~(SR/SR\_W)~\cite{wang2020reinforcement} methods (SR estimates the confusion matrix and SR\_W receives the true confusion matrix as input). Our proposed methods are Distributional Reward Critic~\textbf{(DRC)} and General Distributional Reward Critic~\textbf{(GDRC)}. The hyperparameters are provided in Appendix~\ref{appendix:d}.

\paragraph{Environments} First, we consider complex Mujoco environments: Hopper, HalfCheetah, Walker2d, and Reacher~\cite{todorov2012mujoco} (the environments tested by RE), where SR cannot be applied due to the complexity of the state-action space, but our methods can, demonstrating their broader applicability. Then we consider two discrete control tasks, Pendulum and CartPole (the environments tested by SR and SR\_W), and show that our performance is also dominant in these settings. 

\paragraph{Perturbations}

We test two kinds of perturbations: Generalized Confusion Matrix~(GCM) and continuous perturbations. For GCM perturbations, we vary two parameters: the number of intervals $n_r$ and the perturbation ratio $\omega$. An $\omega$ proportion of samples in the interval containing the true reward are perturbed into other intervals at random. For continuous perturbations, we test the same perturbation as \cite{romoff2018reward}. These are zero-mean additive Gaussian noise, the ``Uniform'' perturbation where the reward is sampled uniformly from $[-1,1]$ with a probability of $\omega$ and is unaltered otherwise, and the ``Reward Uniform'' perturbation, adjusting the range of the uniform distribution to $r_\textrm{min}$ and $r_\textrm{max}$ (the minimum and maximum reward achievable in each environment). Each method has its episodic reward averaged over 50 and 20 seeds under GCM and continuous perturbations respectively, and +/- shows the standard error in Appendix~\ref{appendix:e}.

\subsection{Under GCM Perturbations}
\label{eval_cm}

\begin{figure}[tb]
    \centering
    \includegraphics[width=0.474\textwidth]{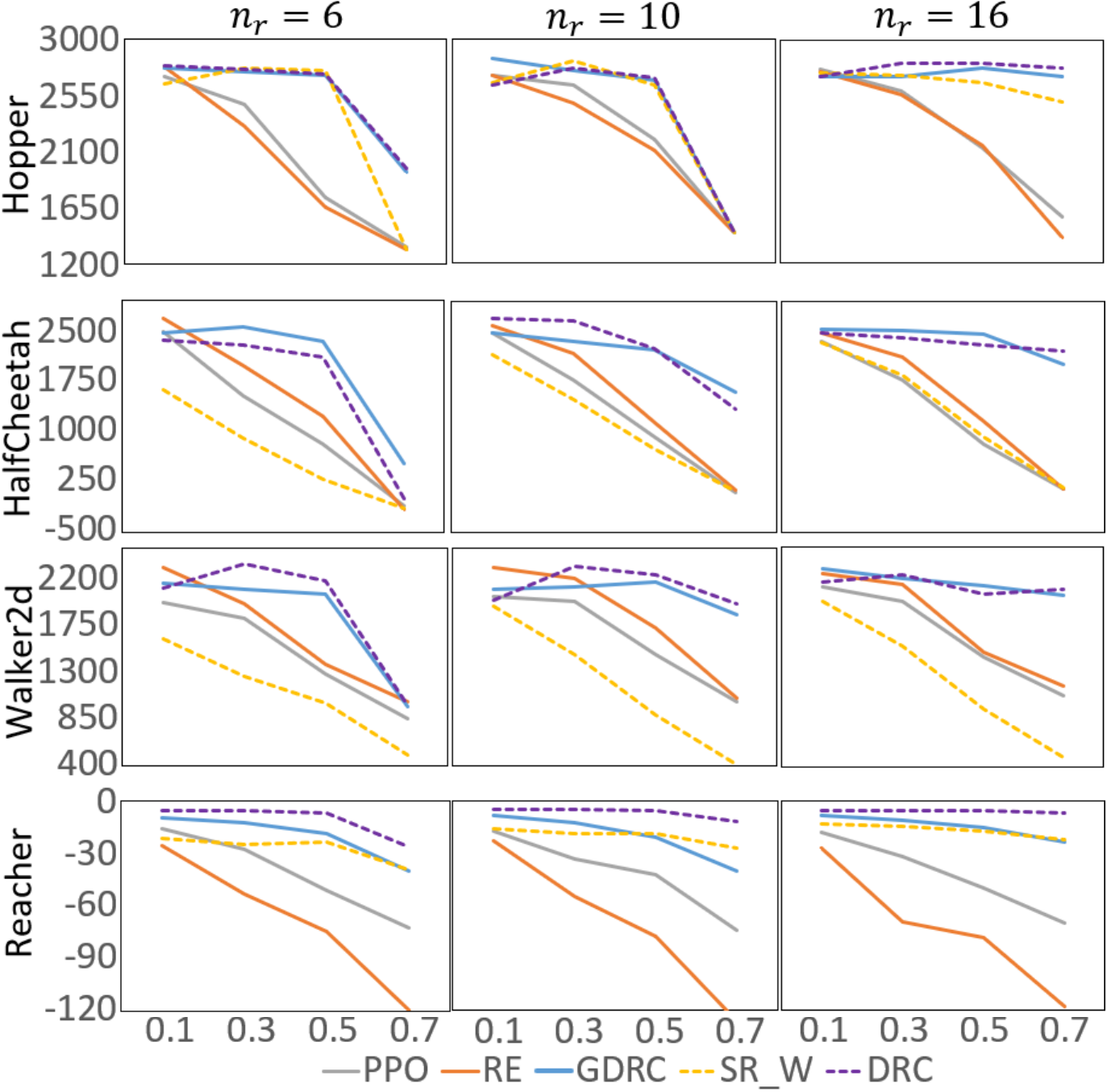}
    \caption{ The results of Mujoco environments under GCM perturbations. Solid line methods, which are of greater interest, can be applied without any prior information. {\color{violet}\textbf{DRC}} and {\color{cyan}\textbf{GDRC}} are our methods. The $x$-axis represents perturbation ratio $\omega$, and the $y$-axis represents performance.}
    \label{fig:cm}
\end{figure}

\paragraph{Mujoco Environments} There are two kinds of methods in Fig.~\ref{fig:cm}, the methods (dashed lines) requiring prior information about perturbations and the ones (solid lines) without any prior knowledge, leading to two comparison groups: DRC vs.\ PPO, RE, and SR\_W, and GDRC vs.\ PPO and RE. Overall, we can clearly see the outperformance of our DRC and GDRC compared with their baseline methods. To summarize, DRC wins/ties ($95\%$ of the winning performance) the best performance in 44/48 sets (compared to 11/48 for the best baseline RE), and GDRC wins/ties in 44/48 sets (compared to 12/44 for the best baseline RE), of which the second comparison is our focus. Both DRC and GDRC demonstrate comprehensive robustness across environments, $n_r$, and $\omega$.

All methods perform worse as we increase $\omega$. Other than that, varying $n_r$ also influences their performance. There is a structural reason for this: GCMs with smaller $n_r$ have less evenly distributed rewards outside the clean interval, making it harder to denoise the signal.

SR\_W performs well in Hopper and Reacher, but worse than PPO in the HalfCheetah and Walker2d. This is perhaps surprising as it has access to the ground truth reward perturbation. We hypothesize two reasons why it can be beaten.
First, the estimate of reward it provides $\hat{R}=C^{-1}\cdot R$ is conditioned only on the observed reward. This means it is not able to do additional denoising using $(s,a,s')$. Second, they introduce a hyperparameter rescaling the estimates, which is not tunable and has a large impact on performance across environments.

We find that RE's performance is generally similar on average to PPO in these experiments. While RE can reduce reward variance, it also has to learn the reward structure. We hypothesize that the cost of learning the reward structure is sometimes overcome by the benefit of variance reduction and sometimes not.

\begin{figure}[tb]
  \centering
  \includegraphics[width=0.45\textwidth]{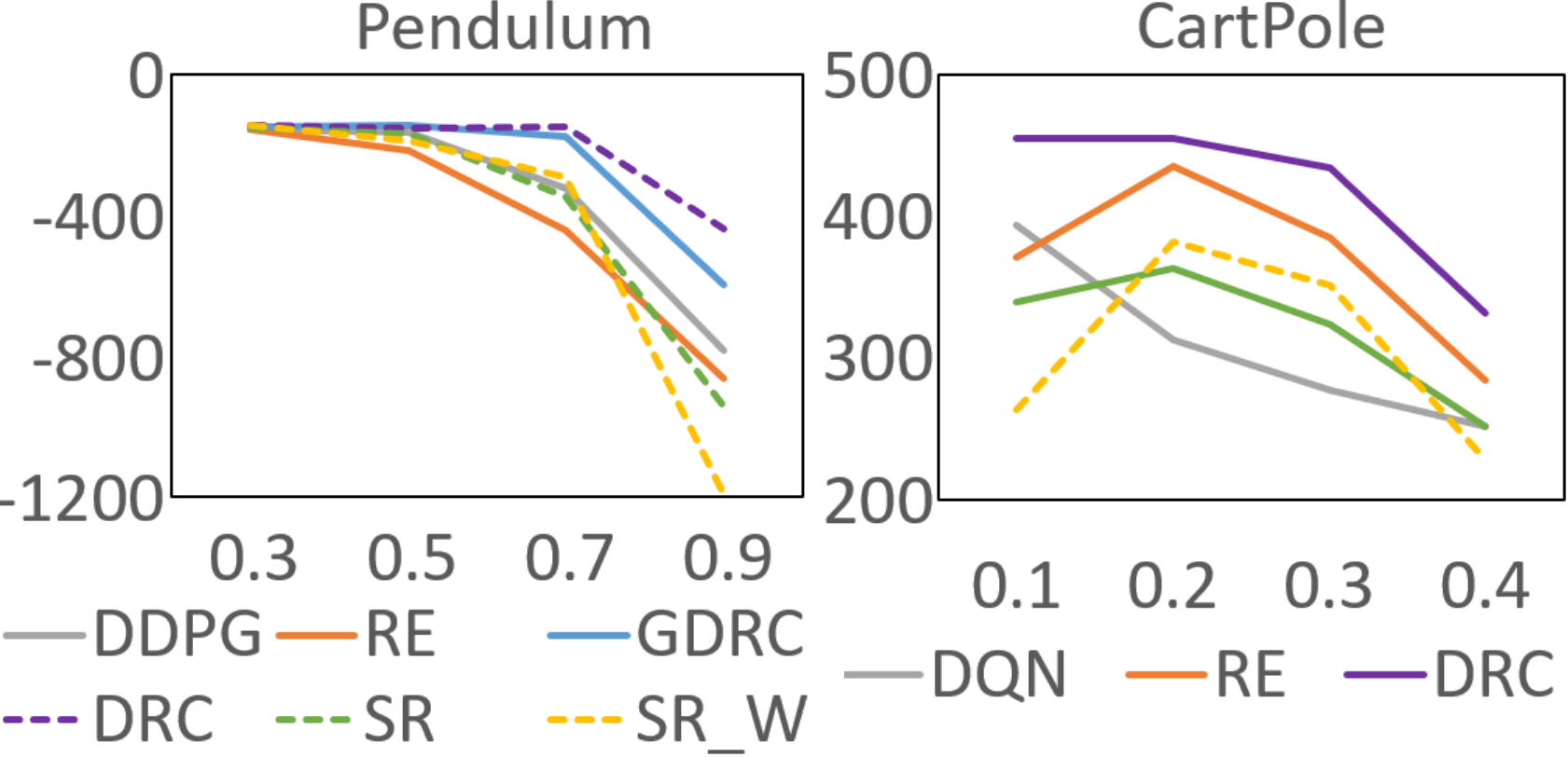}
  \caption{The results of discrete control tasks. Solid line methods, which are of greater interest, can be applied without any prior information. {\color{violet}\textbf{DRC}} and {\color{cyan}\textbf{GDRC}} are our methods. The $x$-axis represents perturbation ratio $\omega$ and the $y$-axis represents performance.}
  \label{fig:control}
\end{figure}

\paragraph{Discrete Control Tasks} We run experiments on two discrete control tasks in Fig.~\ref{fig:control}.
Following the settings in \cite{wang2020reinforcement}, the reward range $[-17, 0)$ is discretized into $n_r=17$ bins in Pendulum. In CartPole, apart from $+1$ rewards, $-1$ rewards are introduced for perturbations by \cite{wang2020reinforcement}. As CartPole is an environment with discrete rewards, DRC and SR can work without prior knowledge. As depicted in Fig.~\ref{fig:control}, DRC is always the best performing for all levels of perturbations, with GDRC as the strongest performer among the methods without prior knowledge in Pendulum across all $\omega$.

\subsection{Impact of $n_o$ on Reconstruction Error And Ordinal Cross-Entropy}
\label{eval_g2C}

\begin{figure}[t]
  \centering
  \begin{subfigure}[b]{0.23\textwidth}
    \includegraphics[width=\textwidth]{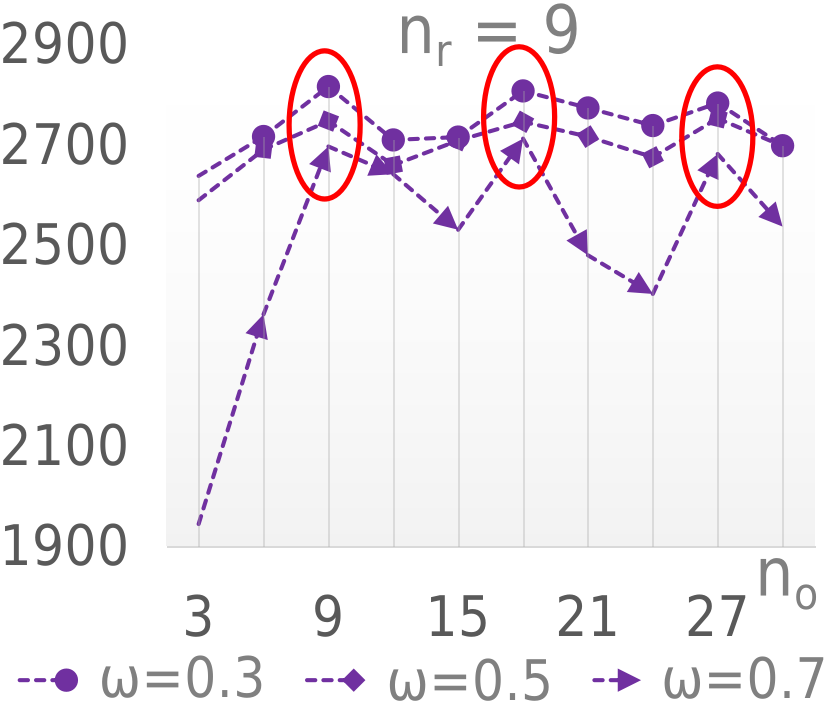}
    \caption{The performance of DRC for different $n_o$ in Hopper, where $n_r=9$.}
    \label{fig:2c_n}
  \end{subfigure}
  \hfill
  \begin{subfigure}[b]{0.23\textwidth}
    \includegraphics[width=\textwidth]{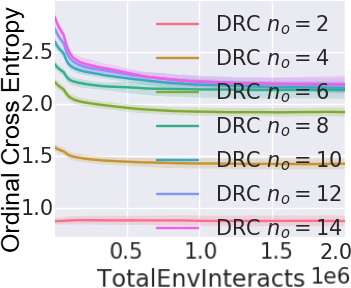}
    \caption{OCE during the training using DRC with different $n_o$, where $n_r=10$.}
    \label{fig:kl_n}
  \end{subfigure}
  \caption{Performance and OCE as $n_o$ varies in Hopper.}
  \label{fig:side_by_side2}
\end{figure}

In Fig.~\ref{fig:side_by_side2}, we study the impact of $n_o$ on Reconstruction Error (reflected as performance) and ordinal cross-entropy (OCE), experimentally verifying our propositions about the relation between $n_o$ and Reconstruction Error and OCE in Fig.~\ref{fig:r_error},~\ref{fig:kl} and Sec.~\ref{G2C}, and thus supporting our design choices for GDRC.

Fig.~\ref{fig:2c_n} compares the performance of DRC as $n_o$ varies in Hopper when $n_r=9$. As we analyze in Sec.~\ref{re_error}, Reconstruction Error turns zero when $n_o$ is a multiple of $n_r$. For $n_o=a \cdot n_r (a\in \mathbb{Z}^+)$ and any $\omega$, we can observe the peak performance of DRC as circled. Fig.~\ref{fig:2c_n} supports our strategy of aiming for $n_o=n_r$ of GDRC using ordinal cross-entropy.

In Fig.~\ref{fig:kl_n}, we study the empirical impact of $n_o$ on ordinal cross-entropy (OCE) throughout the training. If we read Fig.~\ref{fig:kl_n} from the bottom to the top at a certain training step, we can tell OCE increases rapidly for small $n_o$ and stops increasing around $n_o=n_r$, replicating the simulation results in Fig.~\ref{fig:kl} in real experiments. This suggests that OCE is a reliable metric for selecting $n_o$ as discussed in Sec.~\ref{KL}.

\subsection{Ordinal Cross-Entropy Matters For DRC}
\label{exp:oce}

\begin{table}[b]
\centering
\begin{adjustbox}{center}
\resizebox{1.0\columnwidth}{!}{
\begin{tabular}{c|cccc} \toprule
    Method  & Hopper &  HalfCheetah & Walker2d &  Reacher \\ \midrule
    DRC\_CE  & 2198.8 & 851.5 & 1747.6 & -12.2   \\
    DRC & \textbf{2677.5}  & \textbf{2127.3} & \textbf{2058.6}  & \textbf{-7.8}  \\ \bottomrule
\end{tabular}
}
\end{adjustbox}
\caption{The average performance of DRC using OCE and DRC\_CE using CE over $n_r=6,10,16$ and $\omega=0.1,0.3,0.5,0.7$}
\label{table:oce}
\end{table}

In Table~\ref{table:oce}, we compare the performance of DRC using OCE and DRC\_CE using CE, which is the only difference between the two methods. The settings are the same as the ones in Sec.~\ref{eval_cm}. DRC always performs better than DRC\_CE, and the gap is especially large in HalfCheetah and Reacher. In Hopper and Walker2d, the episode is cut when the current state is unhealthy (for example, upside down), but this is not the case in HalfCheetah and Reacher, where the episode always continues to the step limit. The result is that ``good'' reward samples are very rare in early episodes of HalfCheetah and Reacher. We hypothesize that CE overfits to the dominant samples, but in OCE, the rare samples receive higher weights and are thus predicted more accurately. Fig.~\ref{fig:DRC_CE} in Appendix~\ref{appendix:z} confirms this in HalfCheetah. The dominant reward bin $y_t=2$ is predicted well by both DRC and DRC\_CE. However, DRC\_CE struggles to classify any other samples correctly, whereas DRC quickly learns to distinguish $y_t=1,2,3$, despite the low frequencies of $y_t=1,3$ initially.



\begin{table}[tb]
\centering
\begin{adjustbox}{center}
\resizebox{0.999\columnwidth}{!}{
\fontsize{18}{1.4\baselineskip}\selectfont
\renewcommand{\arraystretch}{1.2}
\begin{tabular}{c|c|cccc} \toprule
    Perturbation & Method  & Hopper &  HalfCheetah & Walker2d &  Reacher \\ \midrule
             & PPO  & \textbf{2699.2} & 2736.3 & 2048.8 & \textbf{-10.1}   \\
    Gaussian & RE & \textbf{2708.7} & \textbf{2871.6} & \textbf{2084.3} & -17.0  \\ 
             & GDRC  & \textbf{2822.9} & \textbf{2798.9} & \textbf{2101.2} & -10.9   \\ \midrule
             & PPO  & \textbf{2743.9} & \textbf{2778.6} & 1955.0 & -7.7   \\
    Uniform & RE & 2575.3 & \textbf{2658.6} & \textbf{2211.3} & -7.2  \\ 
             & GDRC  & \textbf{2736.8} & 2332.2 & \textbf{2274.5} & \textbf{-6.2}  \\ \midrule
    Reward  & PPO  & \textbf{2648.6} & 2030.4 & 1929.2 & -31.4  \\
    Range & RE & 2571.6 & \textbf{2323.6} & \textbf{2235.3} & -43.5  \\ 
    Uniform         & GDRC  & \textbf{2721.4} & \textbf{2387.7} & \textbf{2237.9} & \textbf{-8.4}   \\ \bottomrule
\end{tabular}
}
\end{adjustbox}
\caption{The performance under continuous perturbations.}
\label{table:cont}
\end{table}

\subsection{The Reward Range Matters For GDRC}
\label{exp:moving_windown}

A key design choice for GDRC is how to estimate the reward range. Perhaps the most intuitive choice is to store all the perturbed rewards in a buffer and choose some statistics of the buffer (e.g., percentiles) to estimate the range. However, the challenge is the reward distribution can shift dramatically across epochs, so perhaps old samples could be detrimental by not allowing the range to shift up as the policy improves. Our strategy is to store the samples in a sliding window manner, meaning we only use the samples collected in the latest 20 epochs to decide $[r_{emin},r_{emax}]$. The performance of GDRC over 50 seeds in HalfCheetah with clean reward of storing all samples vs.\ storing the latest 20 epochs of samples is \textbf{1667.9} and \textbf{2698.0}. The dynamics of the maximum of the reward range estimate in HalfCheetah are shown in Fig.~\ref{fig:r_emax} in Appendix~\ref{appendix:z}. Early training generates a large number of low-reward samples (as episodes always reach the step limit). These low-reward samples must be discarded to allow the estimated max reward to increase.



\subsection{Under Continuous Perturbations}
\label{eval_continuous}

In Table~\ref{table:cont}, we compare GRDC to PPO and RE under continuous perturbations  (SR\_W and DRC cannot work outside GCM settings). We test $\sigma=1.0, 1.5, 2.0$ for Gaussian perturbations and $\omega=0.1,0.2,0.3,0.4$ for Uniform and Reward Uniform perturbations. GRDC wins/ties in 10/12, whereas RE wins/ties in 7/12 cases. Even under non-GCM perturbations, GDRC has a small edge over RE, especially targeting this kind of perturbation by making the stringent assumption that $\mathbb{E}(\tilde{r})=\omega_0\cdot r+\omega_1(\omega_0 > 0)$. We attribute GDRC's advantage to its potential ability to decide the best $n_o$ to differentiate the true rewards according to Lipschitz for different continuous perturbations. We also notice that the GDRC is more stable across perturbations and environments than baselines.


\subsection{In Clean Environments}

In Table~\ref{table:clean}, we compare GDRC with PPO and RE in clean environments. All three methods perform comparably, with PPO somewhat worse on Hopper and RE slighly worse on Reacher. This shows that the reward signal can be learned in a way that does not appear to interfere with policy learning. In Hopper, a mild amount of reward noise appears to help PPO explore---performance is better in Fig.~\ref{fig:cm} and Table~\ref{table:cont} (this phenomena has been observed consistently by other researchers including~\cite{wang2020reinforcement}).

\begin{table}[tb]
\centering
\begin{adjustbox}{center}
\resizebox{1.0\columnwidth}{!}{
\begin{tabular}{c|cccc} \toprule
    Method  & Hopper &  HalfCheetah & Walker2d &  Reacher \\ \midrule
    PPO  & 2122.6 & \textbf{2621.2} & \textbf{2222.3} & \textbf{-4.9}   \\
    RE   & \textbf{2689.2} & \textbf{2736.3} & \textbf{2237.2} & -5.5   \\ 
    GDRC & \textbf{2753.7} & \textbf{2698.0} & \textbf{2128.8} & \textbf{-4.9}   \\     \bottomrule
\end{tabular}
}
\end{adjustbox}
\caption{The average performance in clean environments.}
\label{table:clean}
\end{table}



\section*{Acknowledgements}

We thank the Ohio Supercomputer Center~\cite{OhioSupercomputerCenter1987} for providing the computational resources for this work.

\bibliography{aaai25}

\begin{thebibliography}{32}
\providecommand{\natexlab}[1]{#1}

\bibitem[{Amodei et~al.(2016)Amodei, Olah, Steinhardt, Christiano, Schulman, and Man{\'e}}]{amodei2016concrete}
Amodei, D.; Olah, C.; Steinhardt, J.; Christiano, P.; Schulman, J.; and Man{\'e}, D. 2016.
\newblock Concrete problems in AI safety.
\newblock \emph{arXiv preprint arXiv:1606.06565}.

\bibitem[{Bai et~al.(2022)Bai, Jones, Ndousse, Askell, Chen, DasSarma, Drain, Fort, Ganguli, Henighan et~al.}]{bai2022training}
Bai, Y.; Jones, A.; Ndousse, K.; Askell, A.; Chen, A.; DasSarma, N.; Drain, D.; Fort, S.; Ganguli, D.; Henighan, T.; et~al. 2022.
\newblock Training a helpful and harmless assistant with reinforcement learning from human feedback.
\newblock \emph{arXiv preprint arXiv:2204.05862}.

\bibitem[{Bellemare, Dabney, and Munos(2017)}]{bellemare2017distributional}
Bellemare, M.~G.; Dabney, W.; and Munos, R. 2017.
\newblock A distributional perspective on reinforcement learning.
\newblock In \emph{International conference on machine learning}, 449--458. PMLR.

\bibitem[{Center(1987)}]{OhioSupercomputerCenter1987}
Center, O.~S. 1987.
\newblock Ohio Supercomputer Center.

\bibitem[{Choromanski et~al.(2020)Choromanski, Pacchiano, Parker-Holder, Tang, Jain, Yang, Iscen, Hsu, and Sindhwani}]{choromanski2020provably}
Choromanski, K.; Pacchiano, A.; Parker-Holder, J.; Tang, Y.; Jain, D.; Yang, Y.; Iscen, A.; Hsu, J.; and Sindhwani, V. 2020.
\newblock Provably robust blackbox optimization for reinforcement learning.
\newblock In \emph{Conference on Robot Learning}, 683--696. PMLR.

\bibitem[{Corazza, Gavran, and Neider(2022)}]{corazza2022reinforcement}
Corazza, J.; Gavran, I.; and Neider, D. 2022.
\newblock Reinforcement learning with stochastic reward machines.
\newblock In \emph{Proceedings of the AAAI Conference on Artificial Intelligence}, volume~36, 6429--6436.

\bibitem[{Dabney et~al.(2018{\natexlab{a}})Dabney, Ostrovski, Silver, and Munos}]{dabney2018implicit}
Dabney, W.; Ostrovski, G.; Silver, D.; and Munos, R. 2018{\natexlab{a}}.
\newblock Implicit quantile networks for distributional reinforcement learning.
\newblock In \emph{International conference on machine learning}, 1096--1105. PMLR.

\bibitem[{Dabney et~al.(2018{\natexlab{b}})Dabney, Rowland, Bellemare, and Munos}]{dabney2018distributional}
Dabney, W.; Rowland, M.; Bellemare, M.; and Munos, R. 2018{\natexlab{b}}.
\newblock Distributional reinforcement learning with quantile regression.
\newblock In \emph{Proceedings of the AAAI Conference on Artificial Intelligence}, volume~32.

\bibitem[{Everitt et~al.(2017)Everitt, Krakovna, Orseau, Hutter, and Legg}]{everitt2017reinforcement}
Everitt, T.; Krakovna, V.; Orseau, L.; Hutter, M.; and Legg, S. 2017.
\newblock Reinforcement learning with a corrupted reward channel.
\newblock \emph{arXiv preprint arXiv:1705.08417}.

\bibitem[{Hadfield-Menell et~al.(2017)Hadfield-Menell, Milli, Abbeel, Russell, and Dragan}]{hadfield2017inverse}
Hadfield-Menell, D.; Milli, S.; Abbeel, P.; Russell, S.~J.; and Dragan, A. 2017.
\newblock Inverse reward design.
\newblock \emph{Advances in neural information processing systems}, 30.

\bibitem[{Hu et~al.(2022)Hu, Sun, Chen, Huang, Chang, Sun et~al.}]{hu2022distributional}
Hu, J.; Sun, Y.; Chen, H.; Huang, S.; Chang, Y.; Sun, L.; et~al. 2022.
\newblock Distributional Reward Estimation for Effective Multi-Agent Deep Reinforcement Learning.
\newblock \emph{Advances in Neural Information Processing Systems}, 35: 12619--12632.

\bibitem[{Hutter(2005)}]{hutter2005universal}
Hutter, M. 2005.
\newblock \emph{Universal artificial intelligence: Sequential decisions based on algorithmic probability}.
\newblock Springer Science \& Business Media.

\bibitem[{Lillicrap et~al.(2015)Lillicrap, Hunt, Pritzel, Heess, Erez, Tassa, Silver, and Wierstra}]{lillicrap2015continuous}
Lillicrap, T.~P.; Hunt, J.~J.; Pritzel, A.; Heess, N.; Erez, T.; Tassa, Y.; Silver, D.; and Wierstra, D. 2015.
\newblock Continuous control with deep reinforcement learning.
\newblock \emph{arXiv preprint arXiv:1509.02971}.

\bibitem[{Mnih et~al.(2013)Mnih, Kavukcuoglu, Silver, Graves, Antonoglou, Wierstra, and Riedmiller}]{mnih2013playing}
Mnih, V.; Kavukcuoglu, K.; Silver, D.; Graves, A.; Antonoglou, I.; Wierstra, D.; and Riedmiller, M. 2013.
\newblock Playing atari with deep reinforcement learning.
\newblock \emph{arXiv preprint arXiv:1312.5602}.

\bibitem[{Moreno et~al.(2006)Moreno, Mart{\'\i}n, Soria, Magdalena, and Mart{\'\i}nez}]{moreno2006noisy}
Moreno, A.; Mart{\'\i}n, J.~D.; Soria, E.; Magdalena, R.; and Mart{\'\i}nez, M. 2006.
\newblock Noisy reinforcements in reinforcement learning: some case studies based on gridworlds.
\newblock In \emph{Proceedings of the 6th WSEAS international conference on applied computer science}, 296--300.

\bibitem[{Ng, Harada, and Russell(1999)}]{ng1999policy}
Ng, A.~Y.; Harada, D.; and Russell, S. 1999.
\newblock Policy invariance under reward transformations: Theory and application to reward shaping.
\newblock In \emph{Icml}, volume~99, 278--287. Citeseer.

\bibitem[{Ouyang et~al.(2022)Ouyang, Wu, Jiang, Almeida, Wainwright, Mishkin, Zhang, Agarwal, Slama, Ray et~al.}]{ouyang2022training}
Ouyang, L.; Wu, J.; Jiang, X.; Almeida, D.; Wainwright, C.; Mishkin, P.; Zhang, C.; Agarwal, S.; Slama, K.; Ray, A.; et~al. 2022.
\newblock Training language models to follow instructions with human feedback.
\newblock \emph{Advances in Neural Information Processing Systems}, 35: 27730--27744.

\bibitem[{Pattanaik et~al.(2017)Pattanaik, Tang, Liu, Bommannan, and Chowdhary}]{pattanaik2017robust}
Pattanaik, A.; Tang, Z.; Liu, S.; Bommannan, G.; and Chowdhary, G. 2017.
\newblock Robust deep reinforcement learning with adversarial attacks.
\newblock \emph{arXiv preprint arXiv:1712.03632}.

\bibitem[{Pinto et~al.(2017)Pinto, Davidson, Sukthankar, and Gupta}]{pinto2017robust}
Pinto, L.; Davidson, J.; Sukthankar, R.; and Gupta, A. 2017.
\newblock Robust adversarial reinforcement learning.
\newblock In \emph{International Conference on Machine Learning}, 2817--2826. PMLR.

\bibitem[{Puterman(2014)}]{puterman2014markov}
Puterman, M.~L. 2014.
\newblock \emph{Markov decision processes: discrete stochastic dynamic programming}.
\newblock John Wiley \& Sons.

\bibitem[{Qian, Weng, and Tan(2023)}]{qian2023learning}
Qian, J.; Weng, P.; and Tan, C. 2023.
\newblock Learning Rewards to Optimize Global Performance Metrics in Deep Reinforcement Learning.
\newblock \emph{arXiv preprint arXiv:2303.09027}.

\bibitem[{Rakhsha et~al.(2020)Rakhsha, Radanovic, Devidze, Zhu, and Singla}]{rakhsha2020policy}
Rakhsha, A.; Radanovic, G.; Devidze, R.; Zhu, X.; and Singla, A. 2020.
\newblock Policy teaching via environment poisoning: Training-time adversarial attacks against reinforcement learning.
\newblock In \emph{International Conference on Machine Learning}, 7974--7984. PMLR.

\bibitem[{Ring and Orseau(2011)}]{ring2011delusion}
Ring, M.; and Orseau, L. 2011.
\newblock Delusion, survival, and intelligent agents.
\newblock In \emph{Artificial General Intelligence: 4th International Conference, AGI 2011, Mountain View, CA, USA, August 3-6, 2011. Proceedings 4}, 11--20. Springer.

\bibitem[{Romoff et~al.(2018)Romoff, Henderson, Pich{\'e}, Francois-Lavet, and Pineau}]{romoff2018reward}
Romoff, J.; Henderson, P.; Pich{\'e}, A.; Francois-Lavet, V.; and Pineau, J. 2018.
\newblock Reward estimation for variance reduction in deep reinforcement learning.
\newblock In \emph{Proceedings of The 2nd Conference on Robot Learning}.

\bibitem[{Rowland et~al.(2018)Rowland, Bellemare, Dabney, Munos, and Teh}]{rowland2018analysis}
Rowland, M.; Bellemare, M.; Dabney, W.; Munos, R.; and Teh, Y.~W. 2018.
\newblock An analysis of categorical distributional reinforcement learning.
\newblock In \emph{International Conference on Artificial Intelligence and Statistics}, 29--37. PMLR.

\bibitem[{Schulman et~al.(2017)Schulman, Wolski, Dhariwal, Radford, and Klimov}]{schulman2017proximal}
Schulman, J.; Wolski, F.; Dhariwal, P.; Radford, A.; and Klimov, O. 2017.
\newblock Proximal policy optimization algorithms.
\newblock \emph{arXiv preprint arXiv:1707.06347}.

\bibitem[{Silver et~al.(2021)Silver, Singh, Precup, and Sutton}]{silver2021reward}
Silver, D.; Singh, S.; Precup, D.; and Sutton, R.~S. 2021.
\newblock Reward is enough.
\newblock \emph{Artificial Intelligence}, 299: 103535.

\bibitem[{Todorov, Erez, and Tassa(2012)}]{todorov2012mujoco}
Todorov, E.; Erez, T.; and Tassa, Y. 2012.
\newblock Mujoco: A physics engine for model-based control.
\newblock In \emph{2012 IEEE/RSJ international conference on intelligent robots and systems}, 5026--5033. IEEE.

\bibitem[{Wang, Liu, and Li(2020)}]{wang2020reinforcement}
Wang, J.; Liu, Y.; and Li, B. 2020.
\newblock Reinforcement learning with perturbed rewards.
\newblock In \emph{Proceedings of the AAAI conference on artificial intelligence}, volume~34, 6202--6209.

\bibitem[{Zheng, Liu, and Ni(2014)}]{zheng2014robust}
Zheng, J.; Liu, S.; and Ni, L.~M. 2014.
\newblock Robust bayesian inverse reinforcement learning with sparse behavior noise.
\newblock In \emph{Proceedings of the AAAI Conference on Artificial Intelligence}, volume~28.

\bibitem[{Zhong, Wu, and Si(2023)}]{NEURIPS2023_29ef811e}
Zhong, J.; Wu, R.; and Si, J. 2023.
\newblock A Long N-step Surrogate Stage Reward for Deep Reinforcement Learning.
\newblock In Oh, A.; Neumann, T.; Globerson, A.; Saenko, K.; Hardt, M.; and Levine, S., eds., \emph{Advances in Neural Information Processing Systems}, volume~36, 12733--12745. Curran Associates, Inc.

\bibitem[{Zhuang and Sui(2021)}]{zhuang2021no}
Zhuang, V.; and Sui, Y. 2021.
\newblock No-regret reinforcement learning with heavy-tailed rewards.
\newblock In \emph{International Conference on Artificial Intelligence and Statistics}, 3385--3393. PMLR.

\end{thebibliography}

\clearpage

\appendix

\section{Important Figures}
\label{appendix:z}

\begin{figure}[ht]
  \centering
  \includegraphics[width=0.46\textwidth]{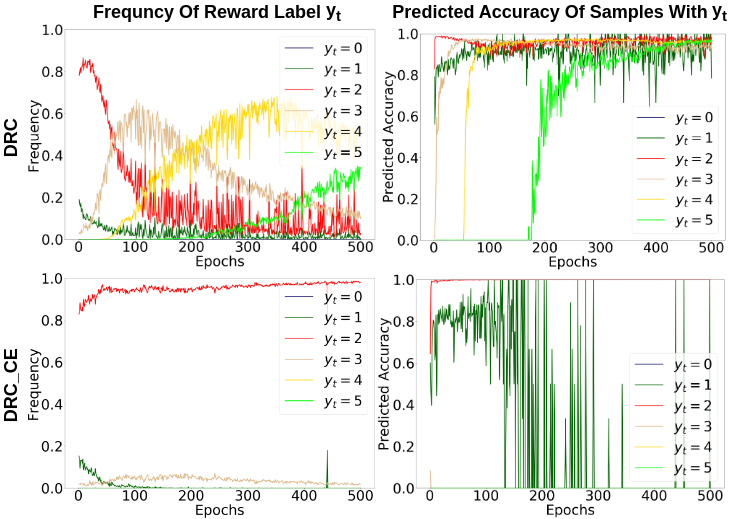}
  \caption{The frequency and the predicted accuracy of 
  each reward class $y_t$ in HalfCheetah under GCM perturbations where $n_r=6$ and $\omega=0.1$.}
  \label{fig:DRC_CE}
\end{figure}

\begin{figure}[ht]
  \centering
  \begin{subfigure}[b]{0.22\textwidth}
    \includegraphics[width=\textwidth]{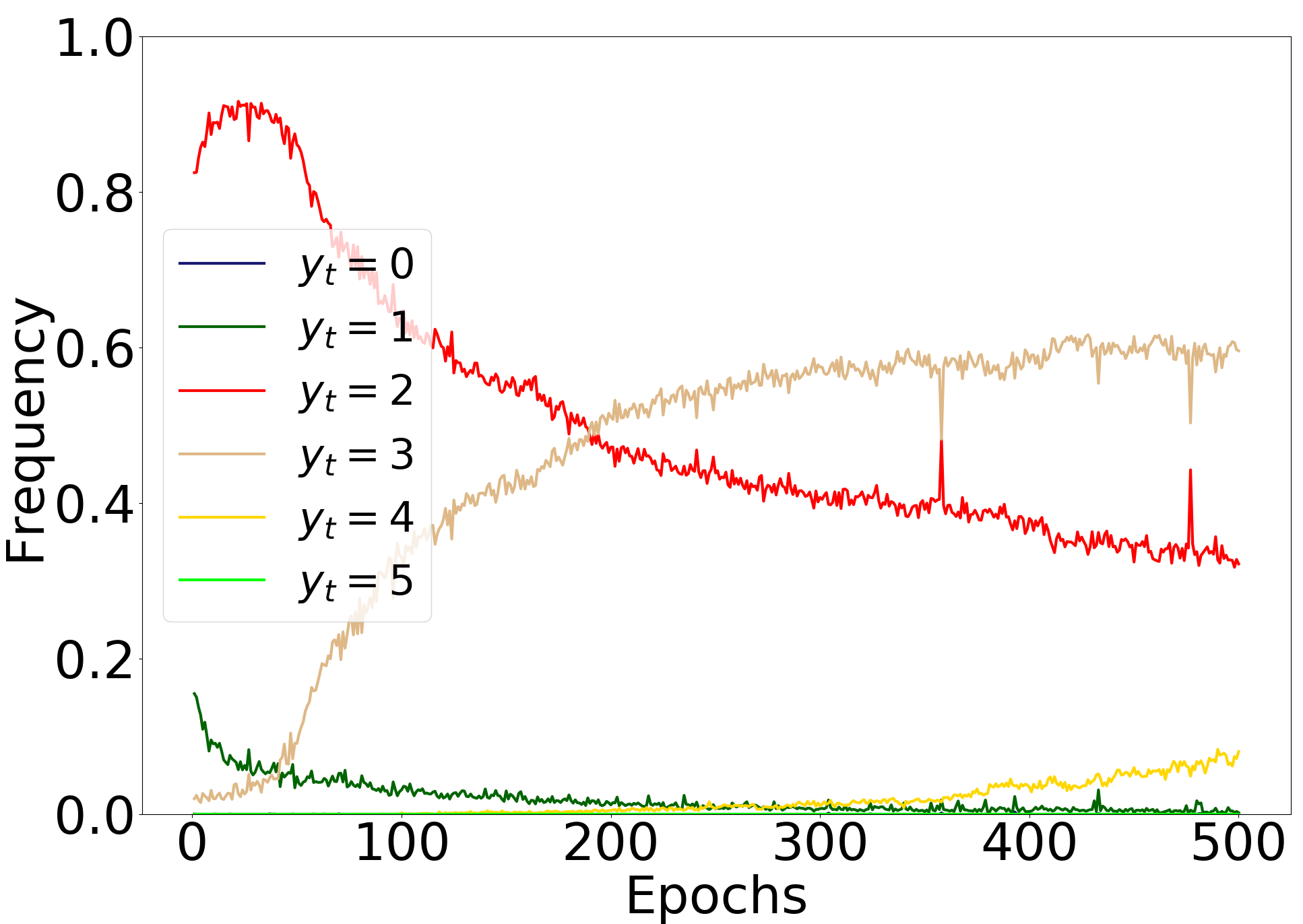}
    \caption{The frequency of the samples by storing all the samples.}
    \label{fig:r_emax_freq}
  \end{subfigure}
  \hfill
  \begin{subfigure}[b]{0.22\textwidth}
    \includegraphics[width=\textwidth]{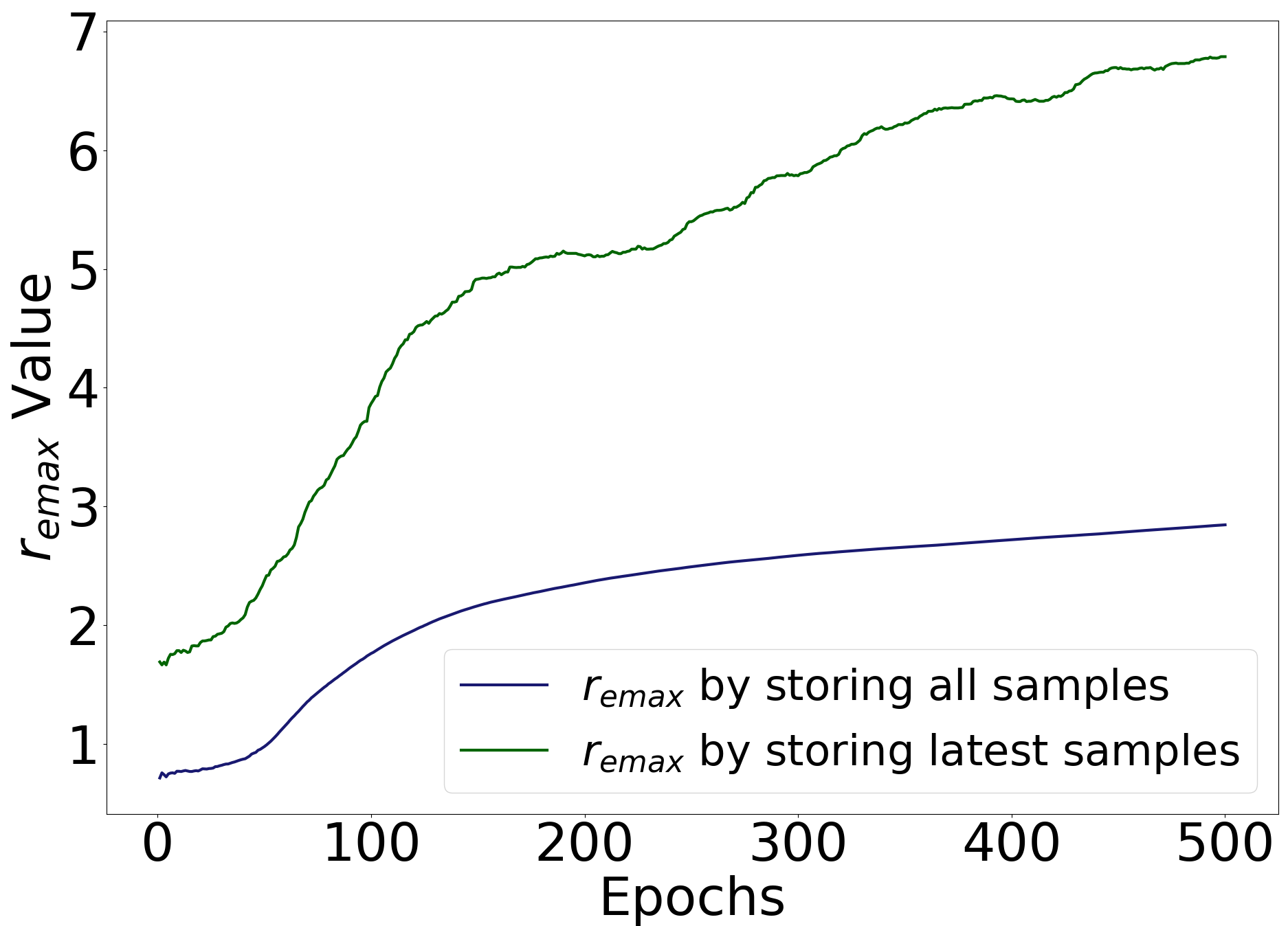}
    \caption{The dynamics of $r_{emax}$ with different strategy.}
    \label{fig:r_emax_value}
  \end{subfigure}
  \caption{The influence of how to store samples on deciding $r_{max}$.}
  \label{fig:r_emax}
\end{figure}

\section{Proof}
\label{appendix:a}

\subsection{Proposition 1}
\label{appendix:a_1}

\begin{proposition}
\label{thm:0_2}

Consider continuous perturbations that for each reward $r\in[r_{\min},r_{\max})$, it can be perturbed to $\bar r \in[r_{\min},r_{\max})$. Our GCM represents $\bar r$ with $\tilde r$ that satisfies $|\tilde r - \bar r|\le \frac{r_{\max} - r_{\min}}{n_r}$.
\end{proposition}

\begin{proof}
This result follows naturally from the proposed generalized confusion matrix perturbation framework. Recall that
we discretize the reward range into $n_r$ intervals, with an equal length $\ell_r=(r_\textrm{max}-r_\textrm{min})/n_r$. If a reward $r \in[r_{\min},r_{\max})$ is perturbed to $\bar r \in[r_{\min},r_{\max})$, we can represent the perturbed reward by $\tilde r = r + (\tilde y - y)\cdot \ell_r$, where $\tilde y - y$ captures the different intervals that $\bar r$ and $r$ fall within. Thus, $\bar r$ and $\tilde r$ belong to the same interval, implying $|\tilde r - \bar r|\le \frac{r_{\max} - r_{\min}}{n_r}$.
\end{proof}

\section{More Analysis of Reconstruction Error}
\label{appendix:b}

The theoretical discussion on Reconstruction Error, $\textsc{Error}_r$, is detailed in Section~\ref{re_error}, and empirical evidence from simulation experiments is depicted in Fig.\ref{fig:r_error}. We want to extract several key observations from Fig.~\ref{fig:r_error} and correlate them with the theoretical discussions, as follows:

\begin{itemize}
    \item In the cases presented in Fig.~\ref{fig:r_error}, $n_r$ is set to 5. It is apparent that Reconstruction Error converges to zero for $n_o=5,10,15,20$ with a standard deviation of zero, suggesting the perturbed rewards can be reverted to the clean rewards exactly when $n_o$ is a multiple of $n_r$ under the premise of infinite samples.
    \item The value of $\textsc{Error}_r$ is lower when $n_o<n_r$ in comparison to when $n_o>n_r$. This suggests that choosing a larger value for $n_o$, without knowledge of $n_r$, may be preferable. However, an overly large value for $n_o$ could lead to overfitting in practical cases, as detailed in Section~\ref{re_error}.
    \item In conjunction with Fig.~\ref{fig:kl}, it is evident that ordinal cross-entropy provides a clear indication of an optimal point when $n_o=n_r$. In this situation, concerns about overfitting are minimized, and we concurrently achieve a Reconstruction Error of zero.
\end{itemize}

\section{Pseudocode of GDRC}
\label{appendix:c}

Alg.~\ref{alg2} presents the pseudocode for implementing GDRC as discussed in Section~\ref{KL}.

\begin{algorithm*}
\caption{General Distributional Reward Critic (GDRC)}
\label{alg2}
\begin{algorithmic}[1]

\Procedure{GDRC Initialization}{}
    \State \textbf{Notations:} $[n_1,n_2,...,n_k]$: the quantities of outputs from the set of DRC under consideration
    \State Initialize $k$ reward critics, denoted as ${\hat{R}_{{\theta}_n}}$, with output quantities $n_1, n_2, \dots, n_k$.
    \State \textbf{Input:} $(s,a,s')$ (for notation simplicity later, we omit the $t$ in the subscript)
    \State \textbf{Output:} $\left[0,1\right]^{n},n\in[n_1,n_2,\dots,n_k]$
    \[\]
\EndProcedure

\Procedure{GDRC Training}{}
    \State \textbf{Notations:} $\tilde y_n$: perturbed reward label regarding $n$ of ${\hat{R}_{{\theta}_n}}$; Sliding\_Buf: the buffer to store the rewards of the latest 20 epochs; $r_{\textrm{emin}}$: the $1\%$ percentile of the collected rewards; $r_{\textrm{emax}}$: the $99\%$ percentile of the collected rewards; $\ell_n$: the length of each reward interval regarding $n$, $\ell_n=(r_{\textrm{emax}} - r_{\textrm{emin}})/n$
    \State \textbf{Input:} samples $(s, a, s', \tilde r)$ collected in an epoch
    \State \textbf{Objective:} parallel training of the reward critics and execution of voting
    \State Time the voted numbers by a discount factor
    \State Store $(\tilde{r})$ into Sliding\_Buf and set $r_{\textrm{emin}}$ and $r_{\textrm{emax}}$ to $1\%$ and $99\%$ percentiles respectively.
    \For{$n$ \textbf{in} $[n_1,n_2,...,n_k]$}
        \State Calculate $\ell_n$: $\ell_n=\big(\left(r_{\textrm{emax}} - r_{\textrm{emin}}\right)/n\big)$
        \State Convert $\tilde r$ to $\tilde y_n$ by applying: $\tilde y_n = floor\big(\left(\tilde r - r_{\textrm{emin}}\right)/\ell_n\big)$
        \While{training iterations threshold not reached}
            \State Train $\textsc{DRC}_n$ using inputs $(s,a,s')$ and labels $\tilde y_n$
        \EndWhile
    \EndFor
    \For{$n$ \textbf{in} $[n_1,n_2,...,n_k]$}
        \State Perform prediction using ${\hat {R_\theta}}_n$
        \State Compute ordinal cross-entropy $OCE=\sum(1+\frac{\left| \hat{y}_n - \tilde{y}_n \right|}{n-1})\cdot H(\tilde{y}_n, {\hat{R}_{{\theta}_n}}(s,a,s'))~\text{where}~\hat{y}_n = {\arg\max} {\hat{R}_{{\theta}_n}}(s,a,s')$

        \State Compute $dH_n = H_n - H_{n-1}$
        \If{$dH_n>dH_{n-1}$}
            \State Cast vote for $n-1$
            \State Break the for-loop
        \EndIf
    \EndFor
    \State Select ${\hat{R}_{{\theta}_n}}$ that received the maximum discounted votes, with $n$ denoting the number of discretization
    \[\]
\EndProcedure

\Procedure{GDRC Predicting}{}
    \State \textbf{Notations:} $r_{\textrm{emin}}$: the $1\%$ percentile of the collected rewards; $r_{\textrm{emax}}$: the $99\%$ percentile of the collected rewards; reward critic ${\hat{R}_{{\theta}_n}}$
    \State \textbf{Input:} $(s, a, s', \tilde{r})$
    \State \textbf{Output:} $(s, a, s', \hat{r})$
    \State Select ${\hat{R}_{{\theta}_n}}$ that received the maximum discounted votes, with $n$ denoting the number of discretization
    \State Compute the length of each reward interval $\ell_n=(r_{\textrm{emax}} - r_{\textrm{emin}})/n$
    \State Associate each sample with a discrete label $\tilde{y}_n = \textrm{floor}\big(\left(\tilde r - r_{\textrm{emin}}\right)/\ell_n\big)$
    \State Input $(s, a, s')$ to ${\hat{R}_{{\theta}_n}}$, and obtain ${\hat{R}_{{\theta}_n}}(s,a, s')$.
    \State Determine the predicted reward label $\hat{y}_n$: $\hat{y}_n=\operatorname*{argmax} {\hat{R}_{{\theta}_n}}(s,a, s')$
    \State Compute the predicted reward value $\hat{r}$: $\hat r=\tilde r + (\hat{y}_n-\tilde{y}_n) \cdot \ell_n$
\EndProcedure

\end{algorithmic}
\end{algorithm*}

\section{Experimental Hyperparameters}
\label{appendix:d}

First, we detail the important hyperparameters used by each method specifically. Second, our experimentation is conducted via both the Spinning Up and Stable Baselines3 frameworks, for which we detail their important hyperparameter configurations separately.

\paragraph{Methods} For Surrogate Reward (SR) methods, including SR and SR\_W, there is an important hyperparameter used to rescale the surrogate rewards $\hat{R}=C^{-1}\cdot R$ as discussed in Sec.~\ref{eval_cm}. Theoretically, the optimal value of this hyperparameter depends on environments and perturbation ratios, making it not tunable without accessing the true rewards. For Pendulum and CartPole they experiment with, we just follow their provided values. For Mujoco environments not tested by them, we just set it to 1. For the Reward Estimation (RE) method, following the hyperparameters provided by the authors, we introduce a three-layer neural network with 64 hidden neurons for reward estimation and set the learning rate and training iterations to be $1e-3$ and 40. For our proposed Distributional Reward Critic (DRC) and General Distributional Reward Critic (GDRC) methods, we do not do any hyperparameter tuning at all. Instead, we just follow the ones used by RE.

\paragraph{Spinning Up} The environments of Hopper-v3, HalfCheetah-v3, Walker2d-v3, and Reacher-v2 are trained using PPO-associated algorithms. Adam optimizers are employed for all neural network training processes. A total of 4,000 steps per epoch is configured, and the agents are trained across 500 epochs, resulting in an aggregate of 2,000,000 interactions. The maximum episodic length is designated as 1,000. Learning rates of 3e-4 and 1e-3 are applied to the policy and the value function respectively, with training conducted over 80 iterations per update. The clipping ratio used in the policy objective is set at 0.2. The GAE-Lambda parameter is set at 0.97. For methods incorporating the Surrogate Reward (SR) method, the reward critic has a learning rate of 1e-3 and trains over 80 iterations. For those involving a Distributional Reward Critic (DRC), the reward critic adopts a learning rate of 1e-3 and trains over 40 iterations per update.

Pendulum-v1 is trained via DDPG-associated algorithms. Adam optimizers are employed for all neural network training processes. Configured for 200 steps per epoch, the agents are trained over 750 epochs, resulting in a total of 150,000 interactions. The maximum episodic length is set at 200. Learning rates of 1e-3 are assigned to both the policy and the Q-function, with updates occurring every 200 steps. The size of the replay buffer is 10e5 and the batch size is set at 200. When the Surrogate Reward (SR) method is employed, the hyperparameters are aligned with~\cite{wang2020reinforcement}. The hyperparameter settings for DRC-related methods remain unchanged.

\paragraph{Stable Baselines3} The hyperparameters utilized in Stable Baselines3 are derived from RL Baselines3 Zoo. CartPole-v1 is trained with DQN-associated algorithms. Adam optimizers are utilized for all neural network training. The agent undergoes a total of 50,000 steps in training. The learning rate for the optimizer is set at 0.0023. The fraction of the total training period during which the exploration rate diminishes is set at 0.16, with the exploration probability dropping from 1.0 to 0.04. The buffer size is sufficiently large to accommodate all samples. When SR or DRC-related methods are applied, the hyperparameters remain unchanged as we discuss previously.

\section{Experiment Result Tables}
\label{appendix:e}

We provide Table~\ref{table:n1_sr}, \ref{table:n2_sr}, \ref{table:n1_2C}, and \ref{table:n2_continuous} of results in Sec.~\ref{eval_cm} and Sec.~\ref{eval_continuous}.

\section{Training Curves}
\label{appendix:f}

\begin{figure*}
 \centering

 \begin{minipage}{0.05\textwidth}
        \centering
        \rotatebox{90}{$\omega=0.1$}
    \end{minipage}%
    \begin{minipage}{0.95\textwidth}
  \begin{subfigure}{0.2\textwidth}
    \centering
    \includegraphics[width=1.4\linewidth]{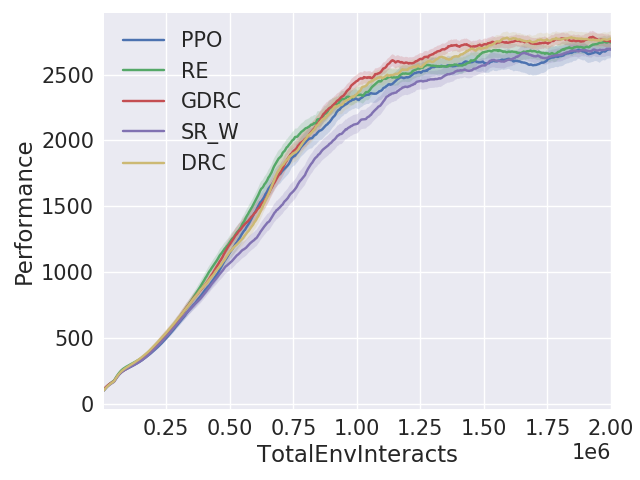}
    \label{fig:image1}
  \end{subfigure}
  \hspace{0.03\textwidth}
  \begin{subfigure}{0.2\textwidth}
    \centering
    \includegraphics[width=1.4\linewidth]{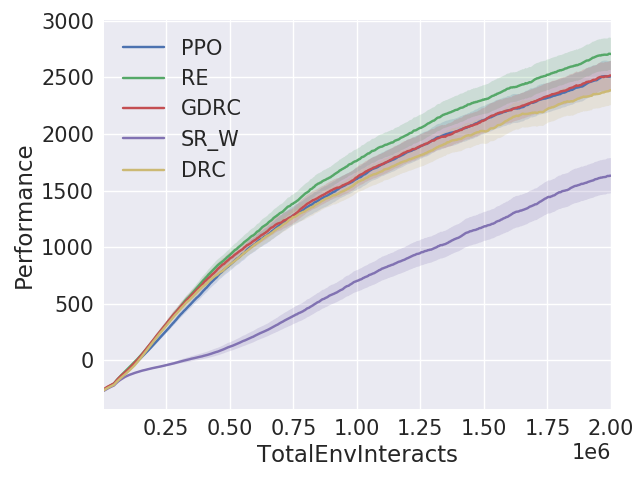}
    \label{fig:image2}
  \end{subfigure}
  \hspace{0.03\textwidth}
  \begin{subfigure}{0.2\textwidth}
    \centering
    \includegraphics[width=1.4\linewidth]{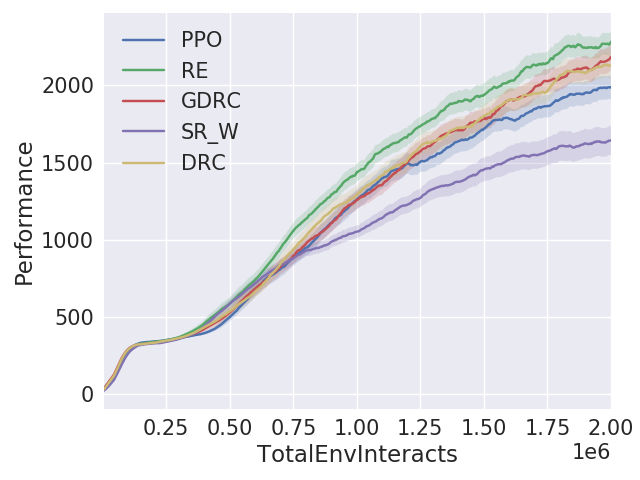}
    \label{fig:image3}
  \end{subfigure}
  \hspace{0.03\textwidth}
  \begin{subfigure}{0.2\textwidth}
    \centering
    \includegraphics[width=1.4\linewidth]{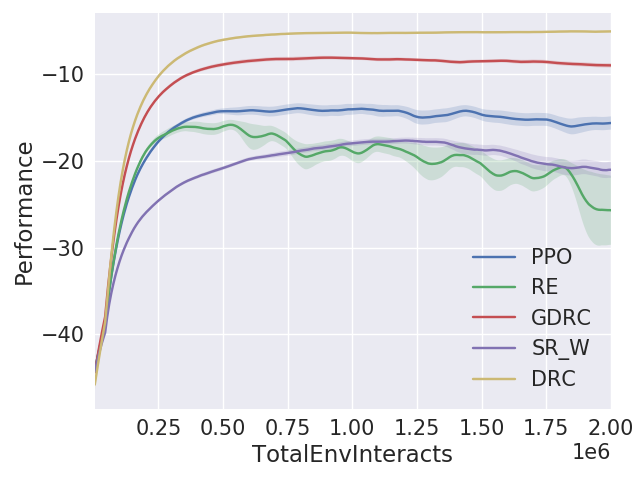}
    \label{fig:image4}
  \end{subfigure}
  \end{minipage}

  \begin{minipage}{0.05\textwidth}
        \centering
        \rotatebox{90}{$\omega=0.3$}
    \end{minipage}%
    \begin{minipage}{0.95\textwidth}
  \begin{subfigure}{0.2\textwidth}
    \centering
    \includegraphics[width=1.4\linewidth]{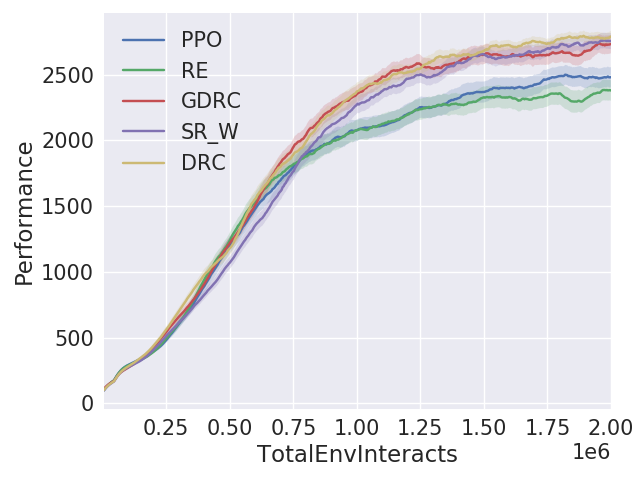}
    \label{fig:image1}
  \end{subfigure}
  \hspace{0.03\textwidth}
  \begin{subfigure}{0.2\textwidth}
    \centering
    \includegraphics[width=1.4\linewidth]{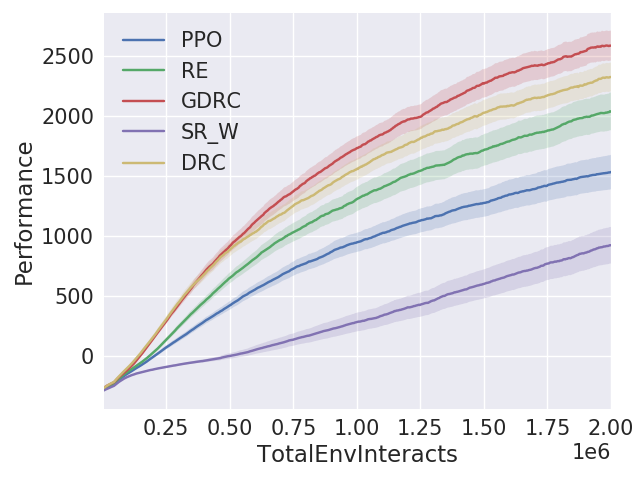}
    \label{fig:image2}
  \end{subfigure}
  \hspace{0.03\textwidth}
  \begin{subfigure}{0.2\textwidth}
    \centering
    \includegraphics[width=1.4\linewidth]{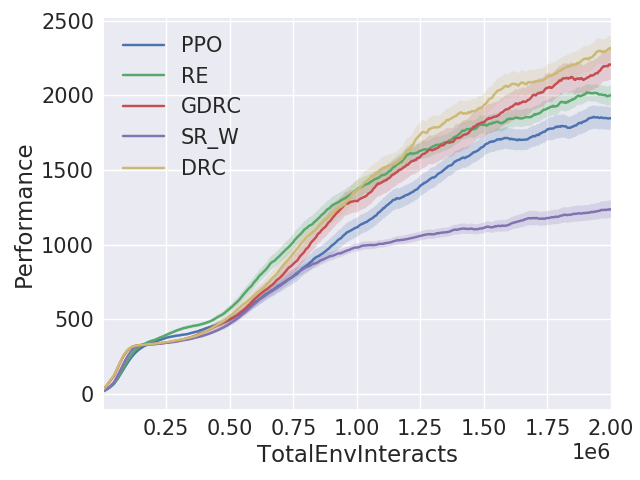}
    \label{fig:image3}
  \end{subfigure}
  \hspace{0.03\textwidth}
  \begin{subfigure}{0.2\textwidth}
    \centering
    \includegraphics[width=1.4\linewidth]{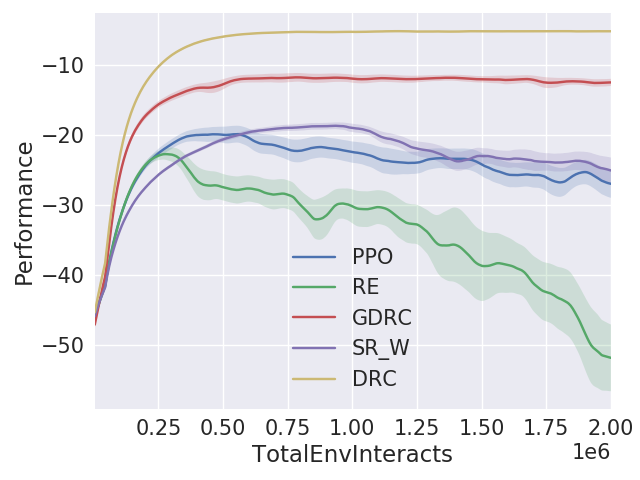}
    \label{fig:image4}
  \end{subfigure}
  \end{minipage}

  \begin{minipage}{0.05\textwidth}
        \centering
        \rotatebox{90}{$\omega=0.5$}
    \end{minipage}%
    \begin{minipage}{0.95\textwidth}
  \begin{subfigure}{0.2\textwidth}
    \centering
    \includegraphics[width=1.4\linewidth]{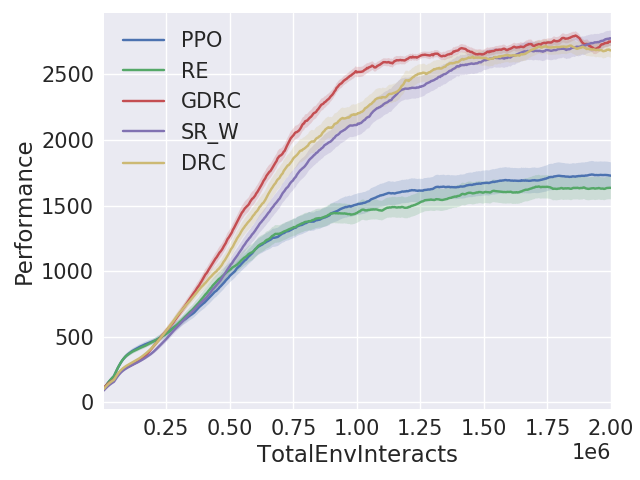}
    \label{fig:image1}
  \end{subfigure}
  \hspace{0.03\textwidth}
  \begin{subfigure}{0.2\textwidth}
    \centering
    \includegraphics[width=1.4\linewidth]{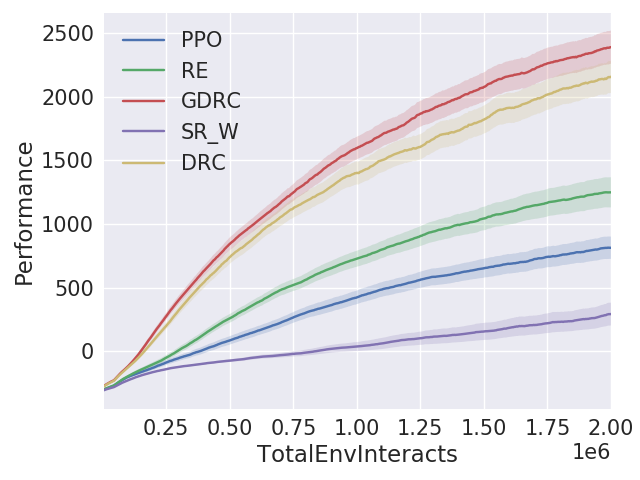}
    \label{fig:image2}
  \end{subfigure}
  \hspace{0.03\textwidth}
  \begin{subfigure}{0.2\textwidth}
    \centering
    \includegraphics[width=1.4\linewidth]{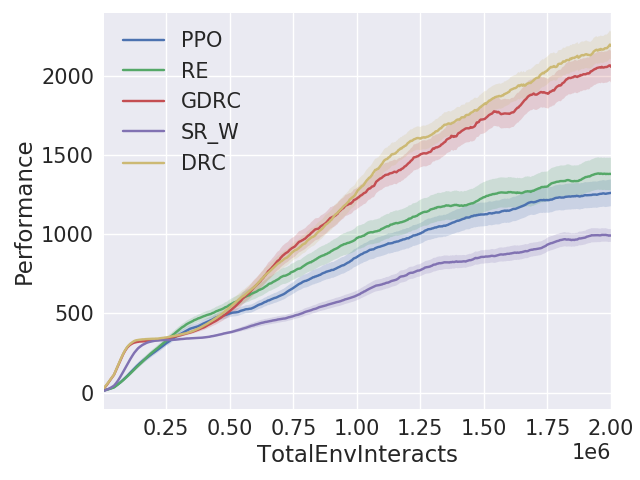}
    \label{fig:image3}
  \end{subfigure}
  \hspace{0.03\textwidth}
  \begin{subfigure}{0.2\textwidth}
    \centering
    \includegraphics[width=1.4\linewidth]{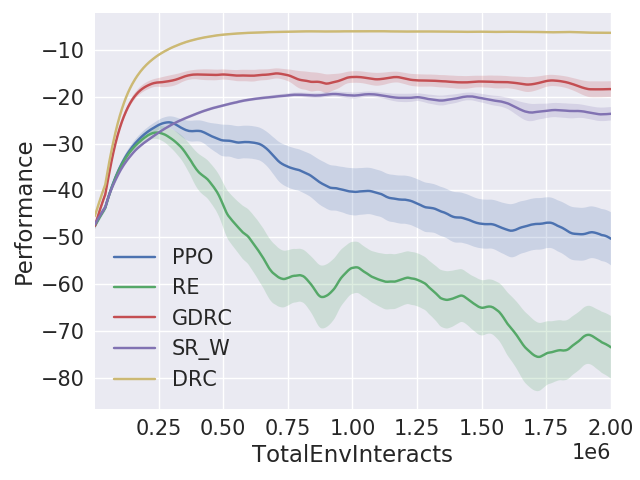}
    \label{fig:image4}
  \end{subfigure}
  \end{minipage}

    \begin{minipage}{0.05\textwidth}
        \centering
        \rotatebox{90}{$\omega=0.7$}
    \end{minipage}%
    \begin{minipage}{0.95\textwidth}
  \begin{subfigure}{0.2\textwidth}
    \centering
    \includegraphics[width=1.4\linewidth]{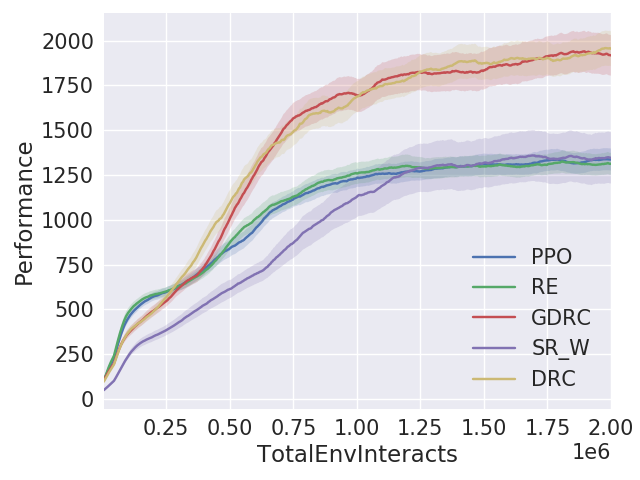}
    \caption{Hopper}
    \label{fig:image1}
  \end{subfigure}
  \hspace{0.03\textwidth}
  \begin{subfigure}{0.2\textwidth}
    \centering
    \includegraphics[width=1.4\linewidth]{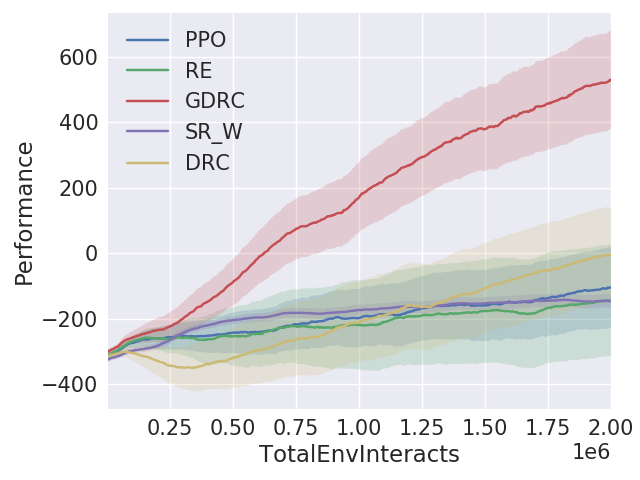}
    \caption{HalfCheetah}
    \label{fig:image2}
  \end{subfigure}
  \hspace{0.03\textwidth}
  \begin{subfigure}{0.2\textwidth}
    \centering
    \includegraphics[width=1.4\linewidth]{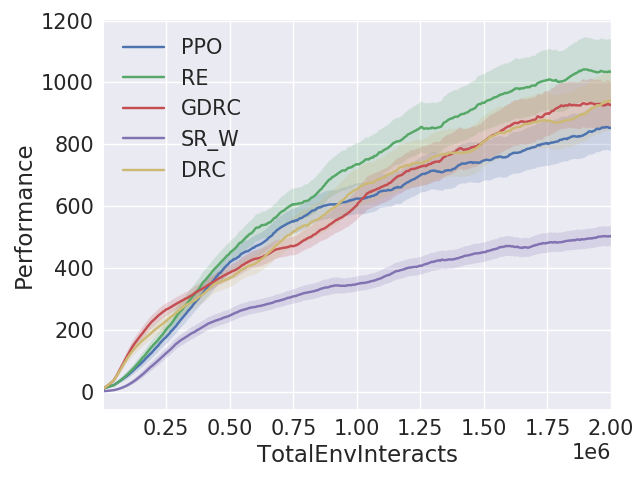}
    \caption{Walker2d}
    \label{fig:image3}
  \end{subfigure}
  \hspace{0.03\textwidth}
  \begin{subfigure}{0.2\textwidth}
    \centering
    \includegraphics[width=1.4\linewidth]{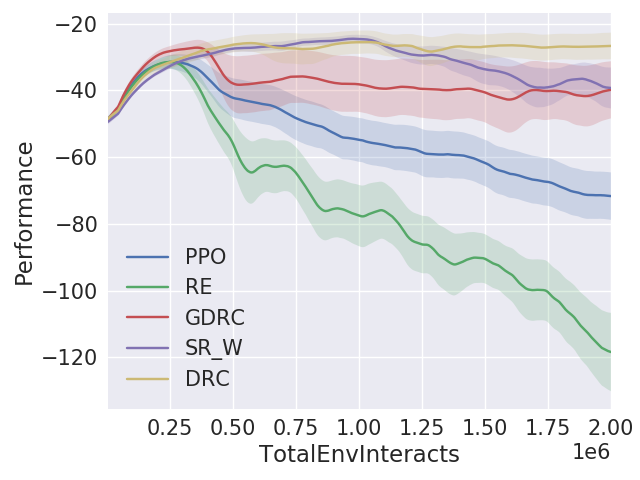}
    \caption{Reacher}
    \label{fig:image4}
  \end{subfigure}
  \end{minipage}

 \caption{Under GCM perturbations where $n_r=6$ and $\omega=0.1,~0.3,~0.5,~0.7$.}
 \label{appfig:n6}
\end{figure*}

\begin{figure*}
  \centering

  \begin{minipage}{0.05\textwidth}
        \centering
        \rotatebox{90}{$\omega=0.1$}
    \end{minipage}%
    \begin{minipage}{0.95\textwidth}
  \begin{subfigure}{0.2\textwidth}
    \centering
    \includegraphics[width=1.4\linewidth]{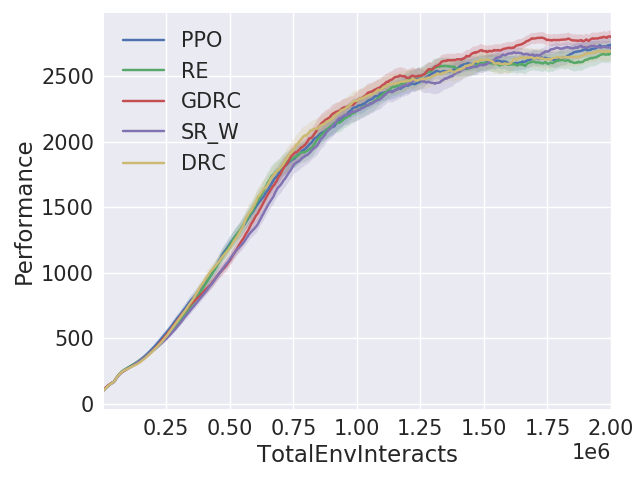}
    \label{fig:image1}
  \end{subfigure}
  \hspace{0.03\textwidth}
  \begin{subfigure}{0.2\textwidth}
    \centering
    \includegraphics[width=1.4\linewidth]{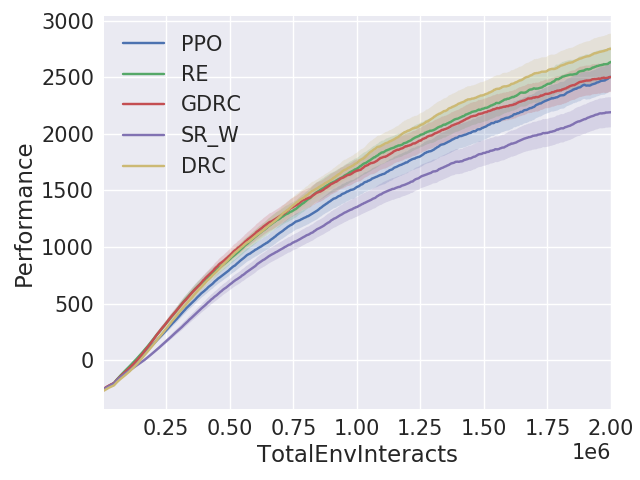}
    \label{fig:image2}
  \end{subfigure}
  \hspace{0.03\textwidth}
  \begin{subfigure}{0.2\textwidth}
    \centering
    \includegraphics[width=1.4\linewidth]{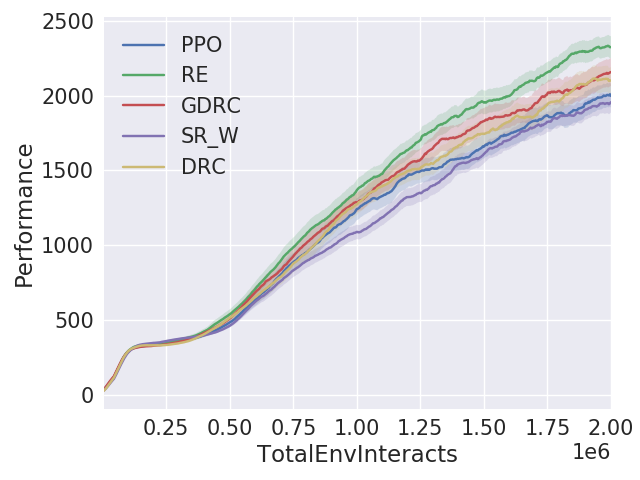}
    \label{fig:image3}
  \end{subfigure}
  \hspace{0.03\textwidth}
  \begin{subfigure}{0.2\textwidth}
    \centering
    \includegraphics[width=1.4\linewidth]{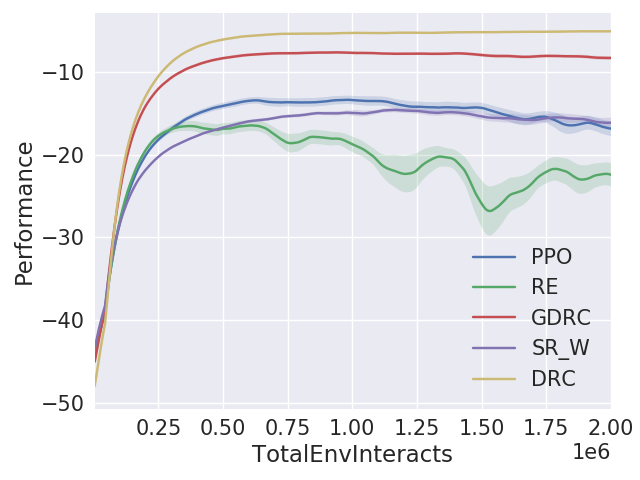}
    \label{fig:image4}
  \end{subfigure}
  \end{minipage}

  \begin{minipage}{0.05\textwidth}
        \centering
        \rotatebox{90}{$\omega=0.3$}
    \end{minipage}%
    \begin{minipage}{0.95\textwidth}
  \begin{subfigure}{0.2\textwidth}
    \centering
    \includegraphics[width=1.4\linewidth]{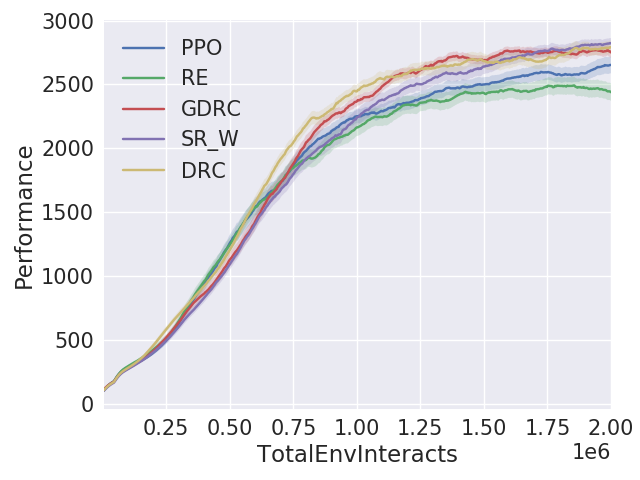}
    \label{fig:image1}
  \end{subfigure}
  \hspace{0.03\textwidth}
  \begin{subfigure}{0.2\textwidth}
    \centering
    \includegraphics[width=1.4\linewidth]{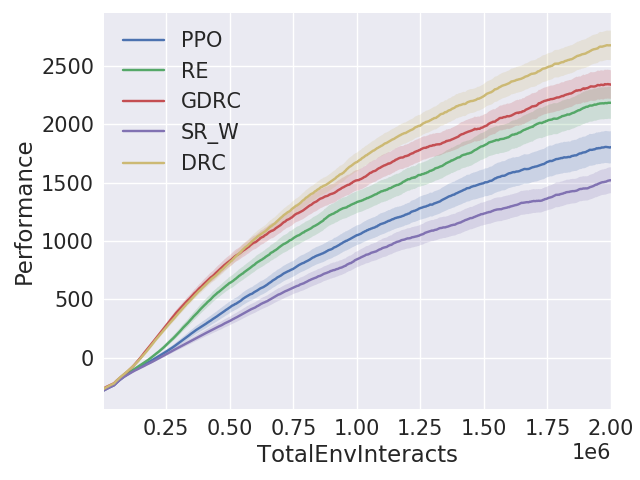}
    \label{fig:image2}
  \end{subfigure}
  \hspace{0.03\textwidth}
  \begin{subfigure}{0.2\textwidth}
    \centering
    \includegraphics[width=1.4\linewidth]{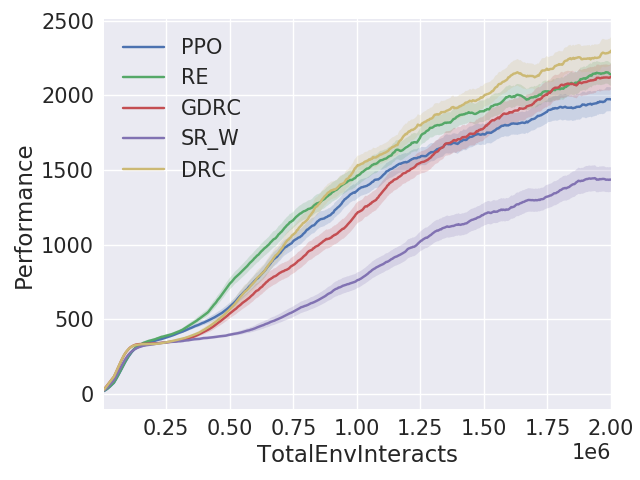}
    \label{fig:image3}
  \end{subfigure}
  \hspace{0.03\textwidth}
  \begin{subfigure}{0.2\textwidth}
    \centering
    \includegraphics[width=1.4\linewidth]{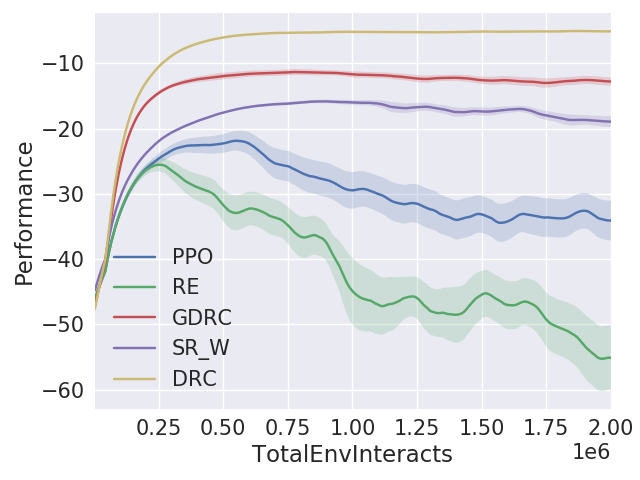}
    \label{fig:image4}
  \end{subfigure}
  \end{minipage}

  \begin{minipage}{0.05\textwidth}
        \centering
        \rotatebox{90}{$\omega=0.5$}
    \end{minipage}%
    \begin{minipage}{0.95\textwidth}
  \begin{subfigure}{0.2\textwidth}
    \centering
    \includegraphics[width=1.4\linewidth]{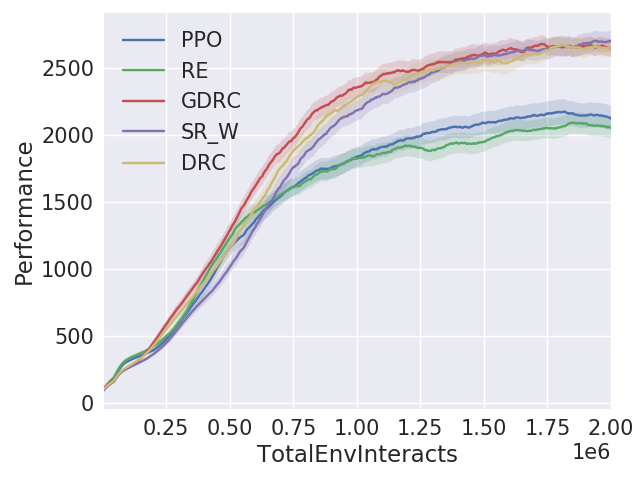}
    \label{fig:image1}
  \end{subfigure}
  \hspace{0.03\textwidth}
  \begin{subfigure}{0.2\textwidth}
    \centering
    \includegraphics[width=1.4\linewidth]{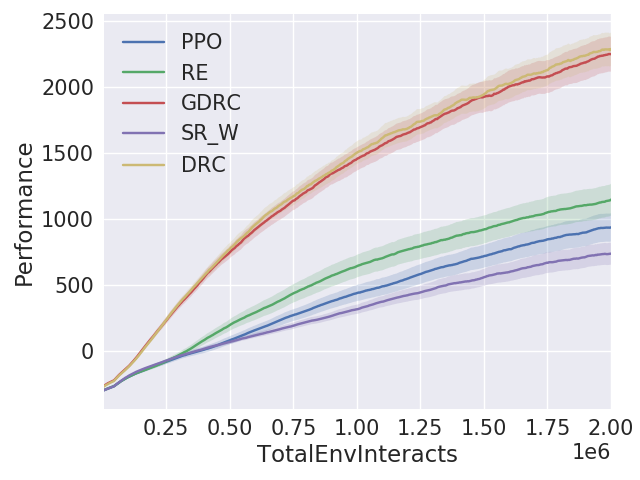}
    \label{fig:image2}
  \end{subfigure}
  \hspace{0.03\textwidth}
  \begin{subfigure}{0.2\textwidth}
    \centering
    \includegraphics[width=1.4\linewidth]{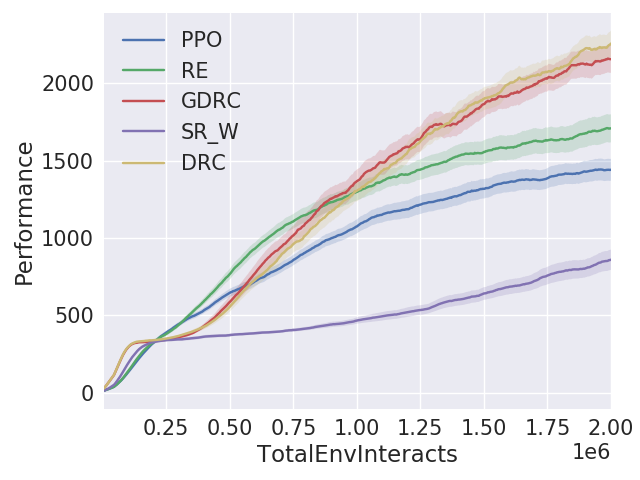}
    \label{fig:image3}
  \end{subfigure}
  \hspace{0.03\textwidth}
  \begin{subfigure}{0.2\textwidth}
    \centering
    \includegraphics[width=1.4\linewidth]{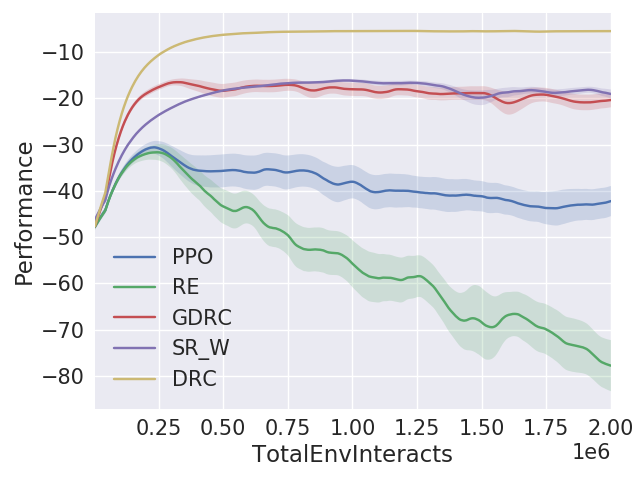}
    \label{fig:image4}
  \end{subfigure}
  \end{minipage}

    \begin{minipage}{0.05\textwidth}
        \centering
        \rotatebox{90}{$\omega=0.7$}
    \end{minipage}%
    \begin{minipage}{0.95\textwidth}
  \begin{subfigure}{0.2\textwidth}
    \centering
    \includegraphics[width=1.4\linewidth]{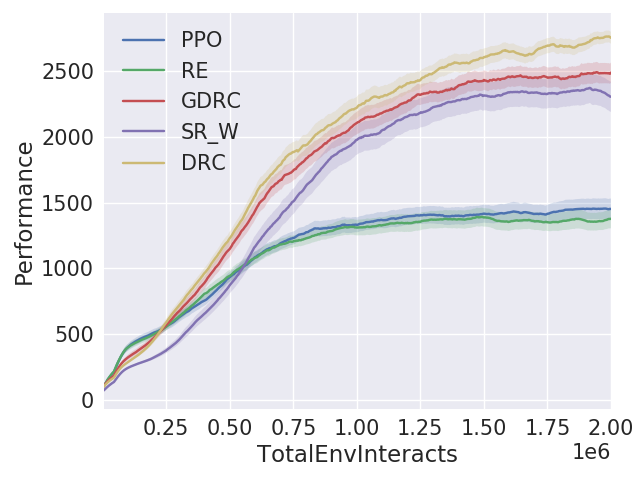}
    \caption{Hopper}
    \label{fig:image1}
  \end{subfigure}
  \hspace{0.03\textwidth}
  \begin{subfigure}{0.2\textwidth}
    \centering
    \includegraphics[width=1.4\linewidth]{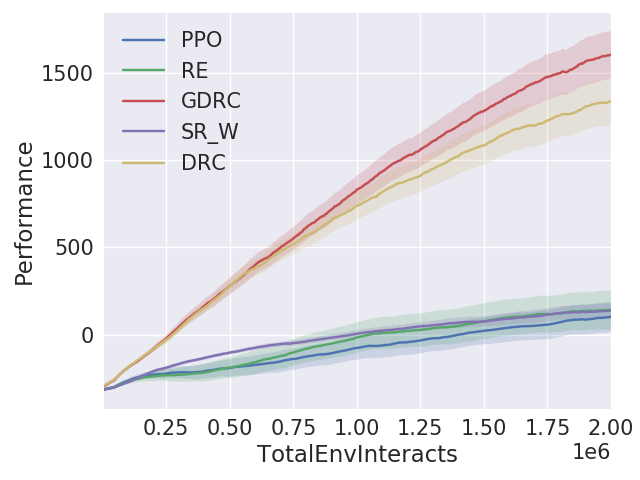}
    \caption{HalfCheetah}
    \label{fig:image2}
  \end{subfigure}
  \hspace{0.03\textwidth}
  \begin{subfigure}{0.2\textwidth}
    \centering
    \includegraphics[width=1.4\linewidth]{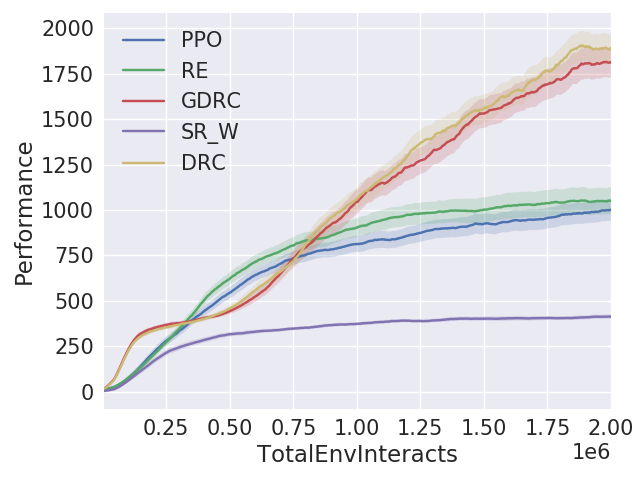}
    \caption{Walker2d}
    \label{fig:image3}
  \end{subfigure}
  \hspace{0.03\textwidth}
  \begin{subfigure}{0.2\textwidth}
    \centering
    \includegraphics[width=1.4\linewidth]{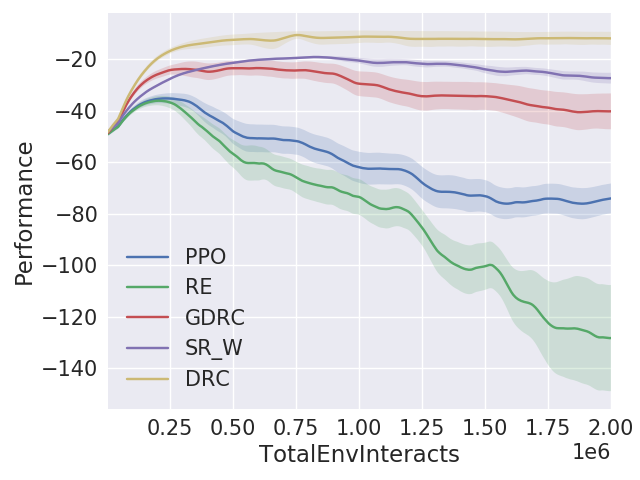}
    \caption{Reacher}
    \label{fig:image4}
  \end{subfigure}
  \end{minipage}

  \caption{Under GCM perturbations where $n_r=10$ and $\omega=0.1,~0.3,~0.5,~0.7$.}
  \label{appfig:n10}
\end{figure*}

\begin{figure*}
  \centering

  \begin{minipage}{0.05\textwidth}
        \centering
        \rotatebox{90}{$\omega=0.1$}
    \end{minipage}%
    \begin{minipage}{0.95\textwidth}
  \begin{subfigure}{0.2\textwidth}
    \centering
    \includegraphics[width=1.4\linewidth]{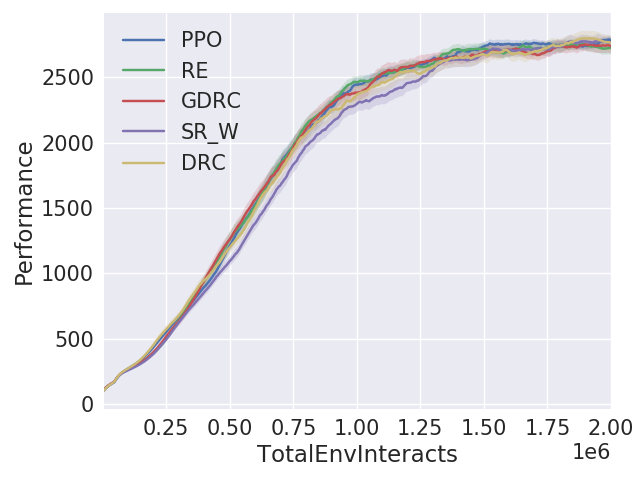}
    \label{fig:image1}
  \end{subfigure}
  \hspace{0.03\textwidth}
  \begin{subfigure}{0.2\textwidth}
    \centering
    \includegraphics[width=1.4\linewidth]{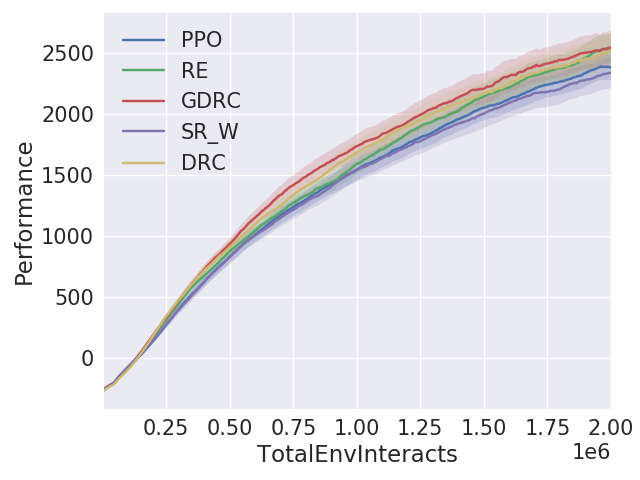}
    \label{fig:image2}
  \end{subfigure}
  \hspace{0.03\textwidth}
  \begin{subfigure}{0.2\textwidth}
    \centering
    \includegraphics[width=1.4\linewidth]{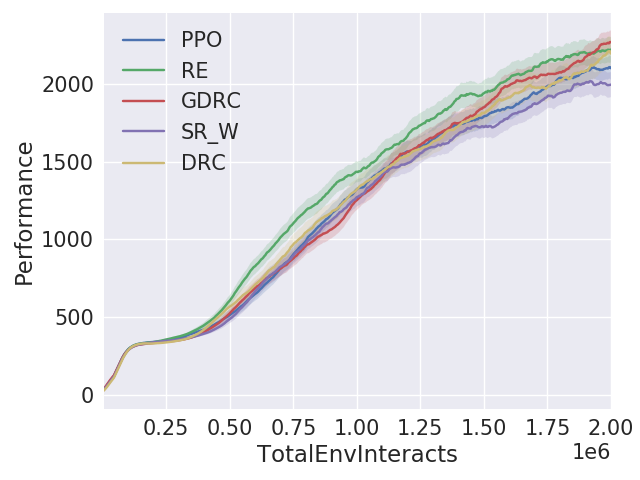}
    \label{fig:image3}
  \end{subfigure}
  \hspace{0.03\textwidth}
  \begin{subfigure}{0.2\textwidth}
    \centering
    \includegraphics[width=1.4\linewidth]{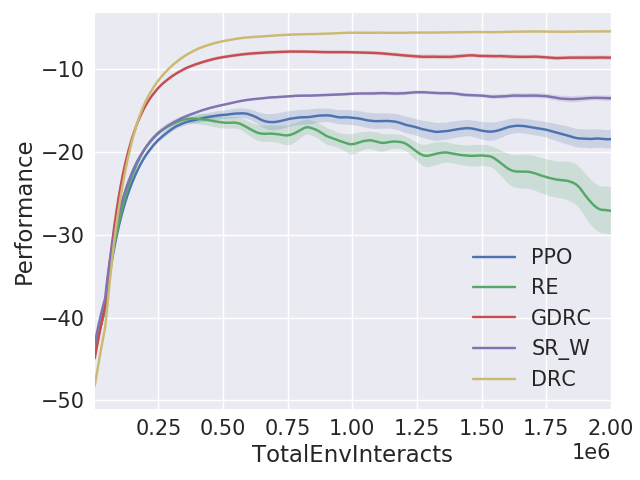}
    \label{fig:image4}
  \end{subfigure}
  \end{minipage}

  \begin{minipage}{0.05\textwidth}
        \centering
        \rotatebox{90}{$\omega=0.3$}
    \end{minipage}%
    \begin{minipage}{0.95\textwidth}
  \begin{subfigure}{0.2\textwidth}
    \centering
    \includegraphics[width=1.4\linewidth]{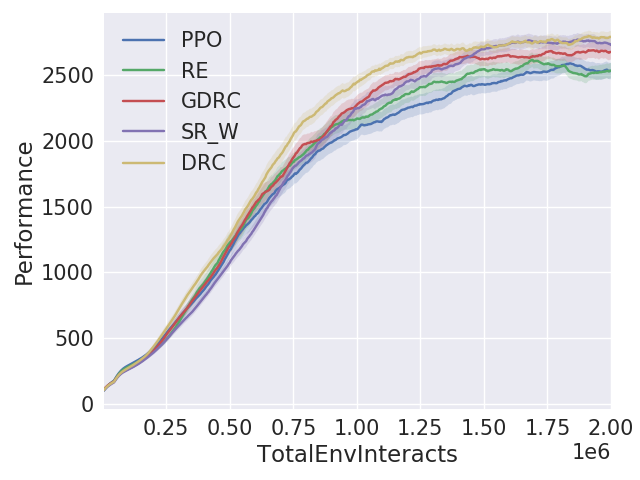}
    \label{fig:image1}
  \end{subfigure}
  \hspace{0.03\textwidth}
  \begin{subfigure}{0.2\textwidth}
    \centering
    \includegraphics[width=1.4\linewidth]{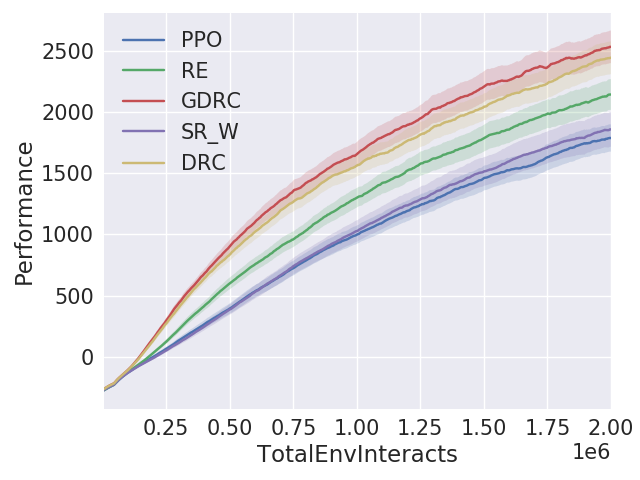}
    \label{fig:image2}
  \end{subfigure}
  \hspace{0.03\textwidth}
  \begin{subfigure}{0.2\textwidth}
    \centering
    \includegraphics[width=1.4\linewidth]{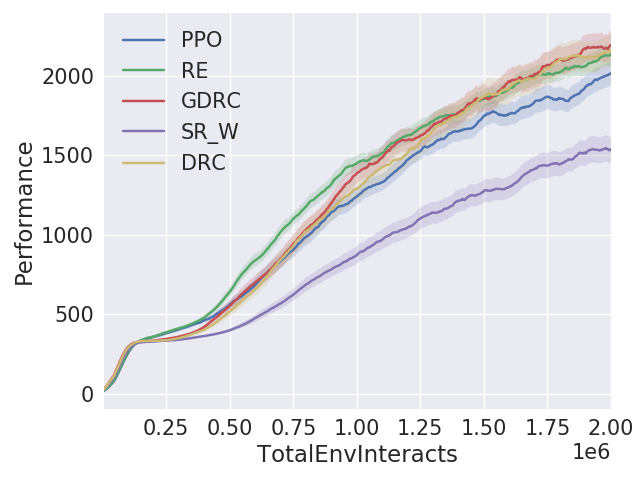}
    \label{fig:image3}
  \end{subfigure}
  \hspace{0.03\textwidth}
  \begin{subfigure}{0.2\textwidth}
    \centering
    \includegraphics[width=1.4\linewidth]{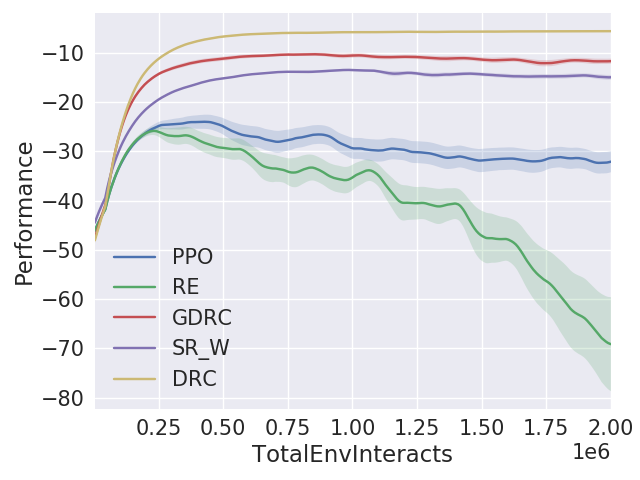}
    \label{fig:image4}
  \end{subfigure}
  \end{minipage}

  \begin{minipage}{0.05\textwidth}
        \centering
        \rotatebox{90}{$\omega=0.5$}
    \end{minipage}%
    \begin{minipage}{0.95\textwidth}
  \begin{subfigure}{0.2\textwidth}
    \centering
    \includegraphics[width=1.4\linewidth]{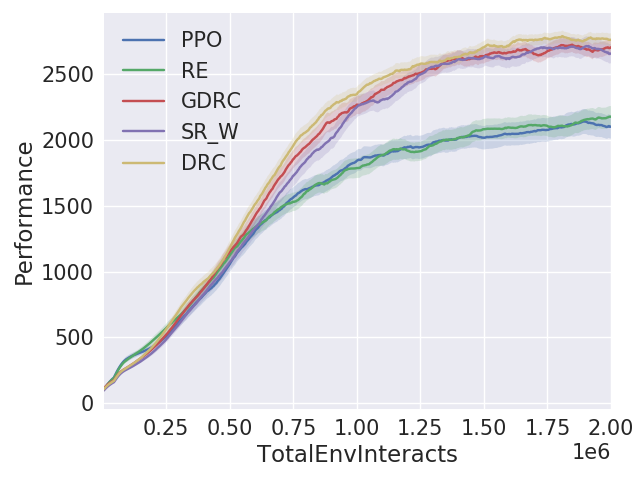}
    \label{fig:image1}
  \end{subfigure}
  \hspace{0.03\textwidth}
  \begin{subfigure}{0.2\textwidth}
    \centering
    \includegraphics[width=1.4\linewidth]{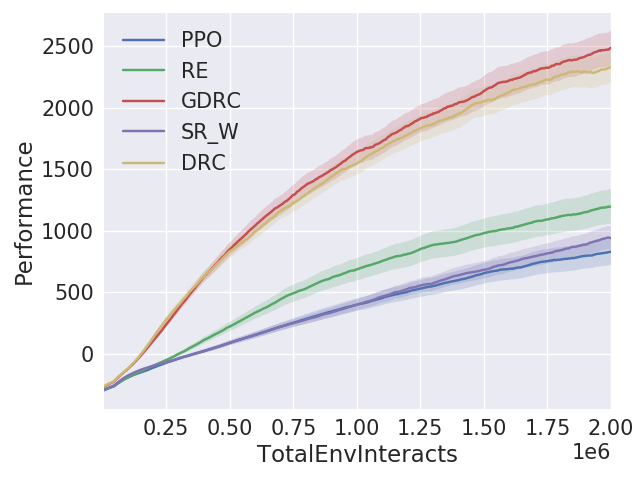}
    \label{fig:image2}
  \end{subfigure}
  \hspace{0.03\textwidth}
  \begin{subfigure}{0.2\textwidth}
    \centering
    \includegraphics[width=1.4\linewidth]{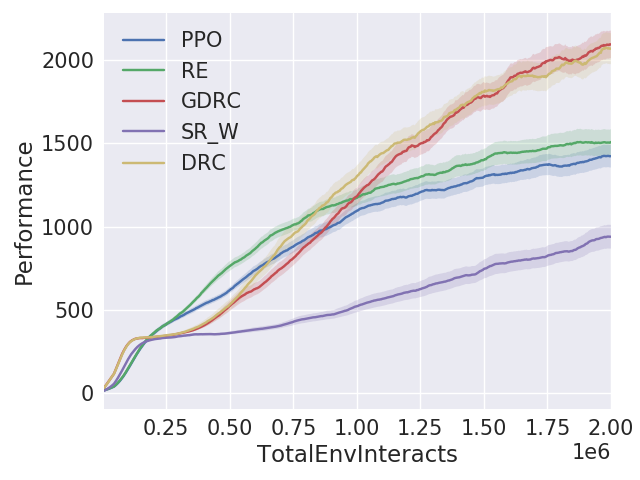}
    \label{fig:image3}
  \end{subfigure}
  \hspace{0.03\textwidth}
  \begin{subfigure}{0.2\textwidth}
    \centering
    \includegraphics[width=1.4\linewidth]{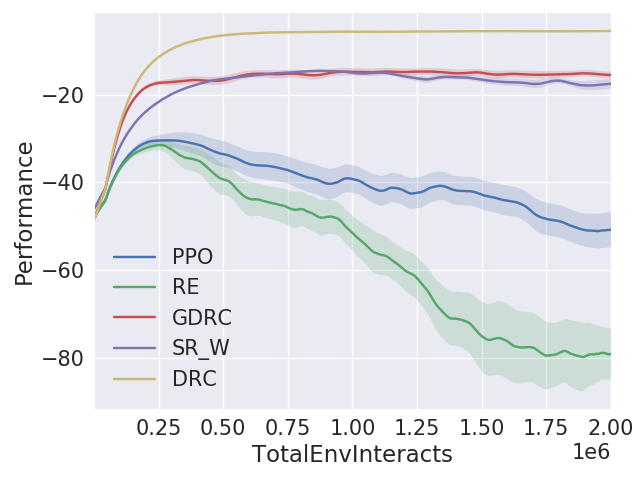}
    \label{fig:image4}
  \end{subfigure}
  \end{minipage}

    \begin{minipage}{0.05\textwidth}
        \centering
        \rotatebox{90}{$\omega=0.7$}
    \end{minipage}%
    \begin{minipage}{0.95\textwidth}
  \begin{subfigure}{0.2\textwidth}
    \centering
    \includegraphics[width=1.4\linewidth]{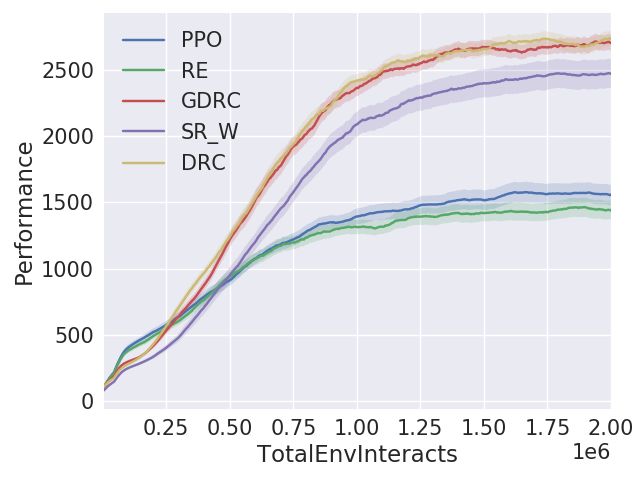}
    \caption{Hopper}
    \label{fig:image1}
  \end{subfigure}
  \hspace{0.03\textwidth}
  \begin{subfigure}{0.2\textwidth}
    \centering
    \includegraphics[width=1.4\linewidth]{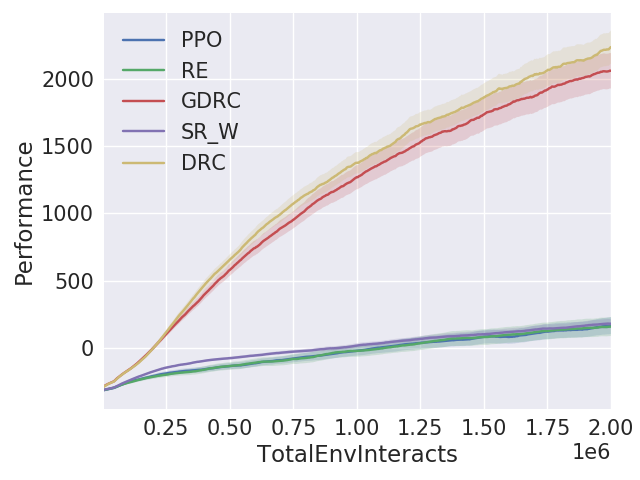}
    \caption{HalfCheetah}
    \label{fig:image2}
  \end{subfigure}
  \hspace{0.03\textwidth}
  \begin{subfigure}{0.2\textwidth}
    \centering
    \includegraphics[width=1.4\linewidth]{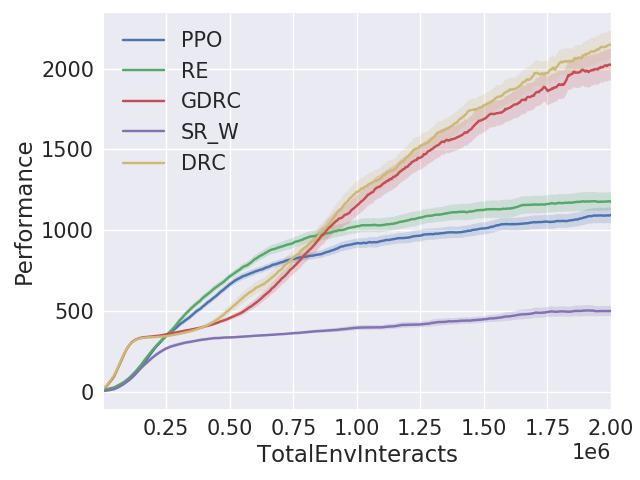}
    \caption{Walker2d}
    \label{fig:image3}
  \end{subfigure}
  \hspace{0.03\textwidth}
  \begin{subfigure}{0.2\textwidth}
    \centering
    \includegraphics[width=1.4\linewidth]{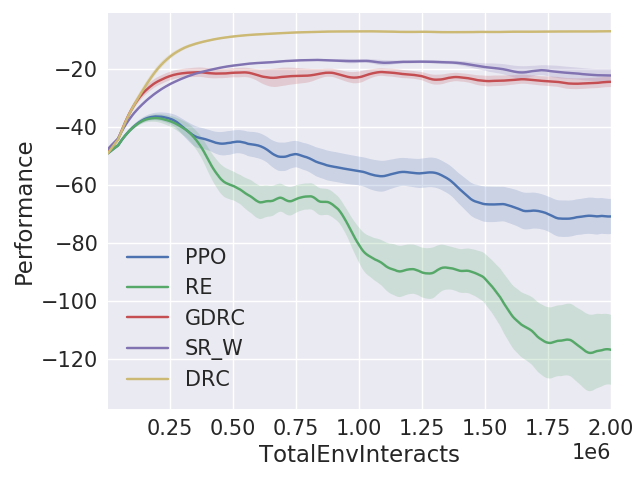}
    \caption{Reacher}
    \label{fig:image4}
  \end{subfigure}
  \end{minipage}

  \caption{Under GCM perturbations where $n_r=16$ and $\omega=0.1,~0.3,~0.5,~0.7$.}
  \label{appfig:n16}
\end{figure*}

\begin{figure*}
  \centering

  \begin{minipage}{0.05\textwidth}
        \centering
        \rotatebox{90}{$\sigma=1.0$}
    \end{minipage}%
  \begin{minipage}{0.95\textwidth}
      \begin{subfigure}{0.2\textwidth}
        \centering
        \includegraphics[width=1.4\linewidth]{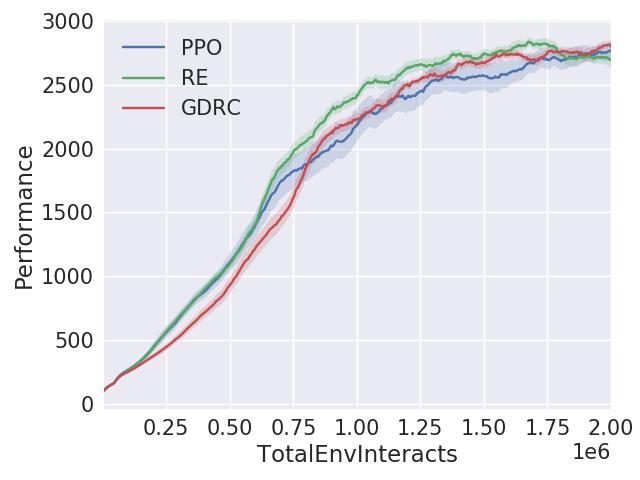}
        \label{fig:image1}
      \end{subfigure}
      \hspace{0.03\textwidth}
      \begin{subfigure}{0.2\textwidth}
        \centering
        \includegraphics[width=1.4\linewidth]{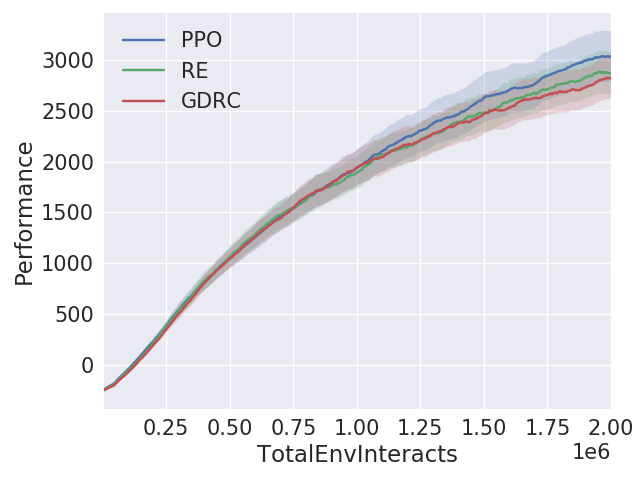}
        \label{fig:image2}
      \end{subfigure}
      \hspace{0.03\textwidth}
      \begin{subfigure}{0.2\textwidth}
        \centering
        \includegraphics[width=1.4\linewidth]{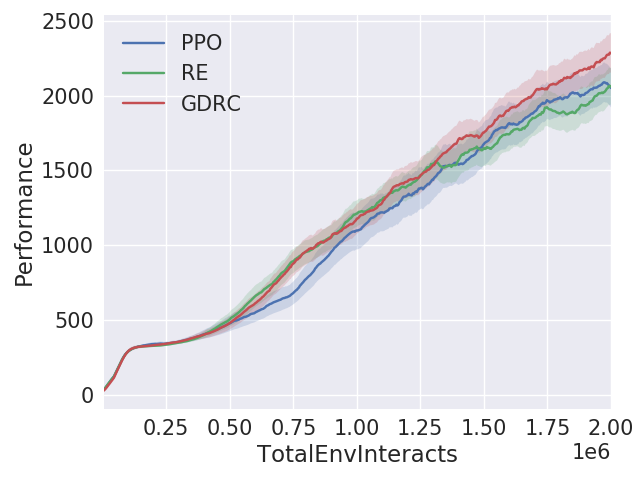}
        \label{fig:image3}
      \end{subfigure}
      \hspace{0.03\textwidth}
      \begin{subfigure}{0.2\textwidth}
        \centering
        \includegraphics[width=1.4\linewidth]{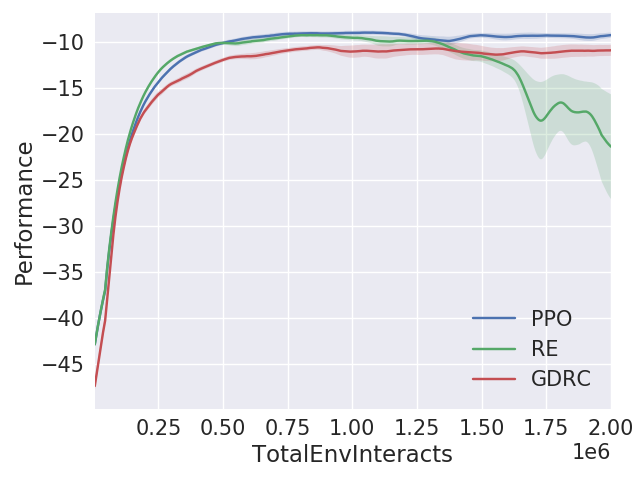}
        \label{fig:image4}
      \end{subfigure}
  \end{minipage}

  \begin{minipage}{0.05\textwidth}
        \centering
        \rotatebox{90}{$\sigma=1.5$}
    \end{minipage}%
  \begin{minipage}{0.95\textwidth}
      \begin{subfigure}{0.2\textwidth}
        \centering
        \includegraphics[width=1.4\linewidth]{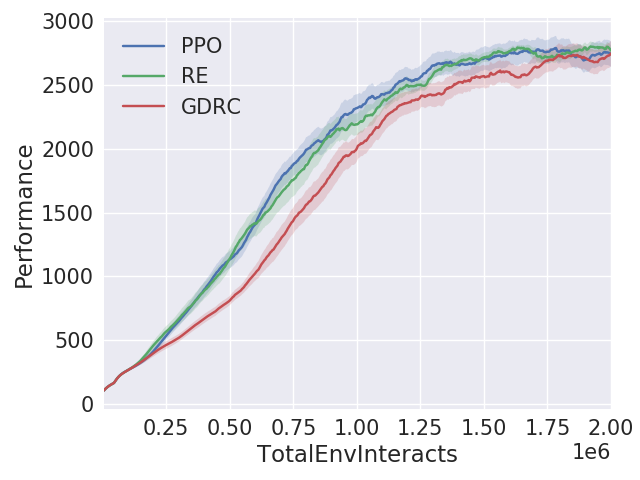}
        \label{fig:image1}
      \end{subfigure}
      \hspace{0.03\textwidth}
      \begin{subfigure}{0.2\textwidth}
        \centering
        \includegraphics[width=1.4\linewidth]{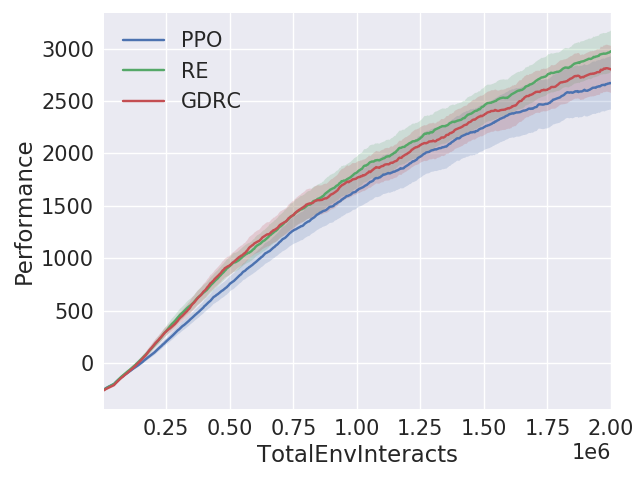}
        \label{fig:image2}
      \end{subfigure}
      \hspace{0.03\textwidth}
      \begin{subfigure}{0.2\textwidth}
        \centering
        \includegraphics[width=1.4\linewidth]{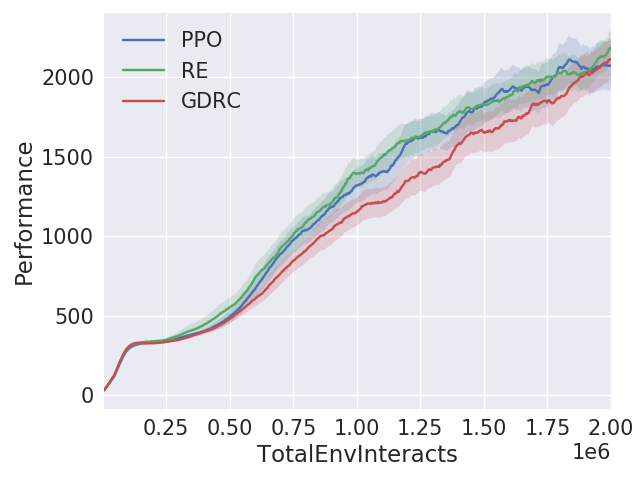}
        \label{fig:image3}
      \end{subfigure}
      \hspace{0.03\textwidth}
      \begin{subfigure}{0.2\textwidth}
        \centering
        \includegraphics[width=1.4\linewidth]{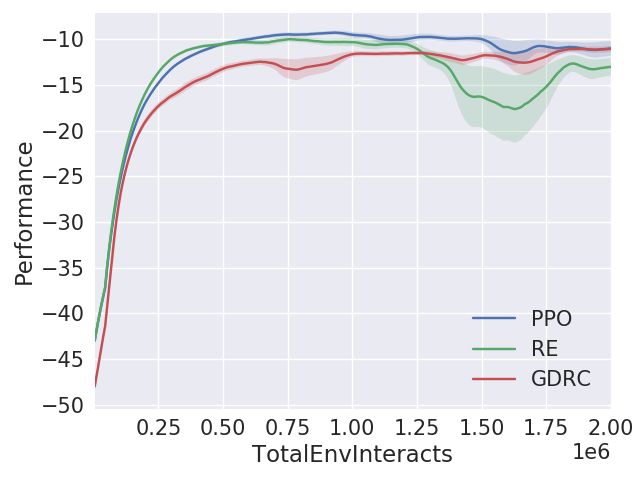}
        \label{fig:image4}
      \end{subfigure}
  \end{minipage}

    \begin{minipage}{0.05\textwidth}
        \centering
        \rotatebox{90}{$\sigma=2.0$}
    \end{minipage}%
  \begin{minipage}{0.95\textwidth}
  \begin{subfigure}{0.2\textwidth}
    \centering
    \includegraphics[width=1.4\linewidth]{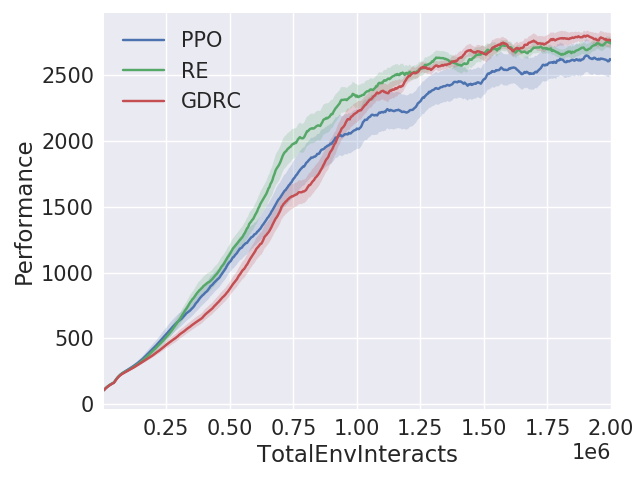}
    \caption{Hopper}
    \label{fig:image1}
  \end{subfigure}
  \hspace{0.03\textwidth}
  \begin{subfigure}{0.2\textwidth}
    \centering
    \includegraphics[width=1.4\linewidth]{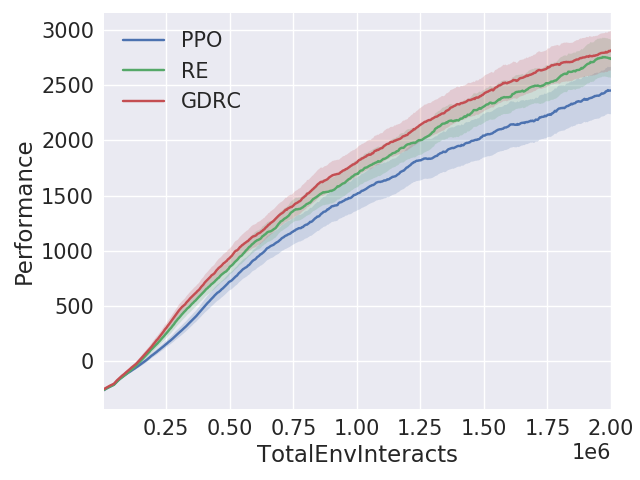}
    \caption{HalfCheetah}
    \label{fig:image2}
  \end{subfigure}
  \hspace{0.03\textwidth}
  \begin{subfigure}{0.2\textwidth}
    \centering
    \includegraphics[width=1.4\linewidth]{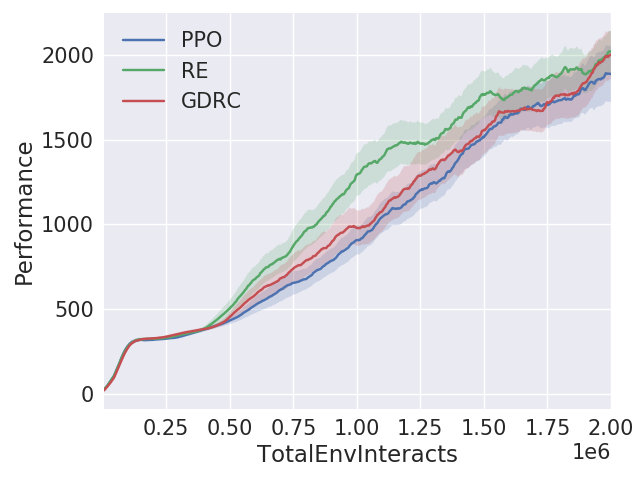}
    \caption{Walker2d}
    \label{fig:image3}
  \end{subfigure}
  \hspace{0.03\textwidth}
  \begin{subfigure}{0.2\textwidth}
    \centering
    \includegraphics[width=1.4\linewidth]{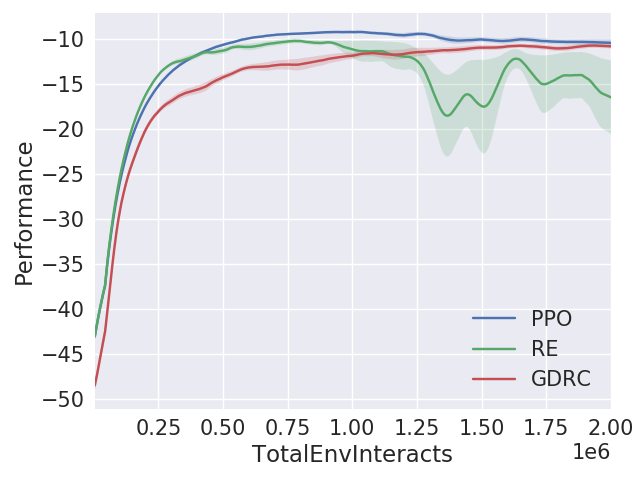}
    \caption{Reacher}
    \label{fig:image4}
  \end{subfigure}
  \end{minipage}

  \caption{Under Gaussian perturbations where $\sigma=1.0,~1.5,~2.0$.}
  \label{appfig:uni1}
\end{figure*}

\begin{figure*}
  \centering

  \begin{minipage}{0.05\textwidth}
        \centering
        \rotatebox{90}{$\omega=0.1$}
    \end{minipage}%
  \begin{minipage}{0.95\textwidth}
      \begin{subfigure}{0.2\textwidth}
        \centering
        \includegraphics[width=1.4\linewidth]{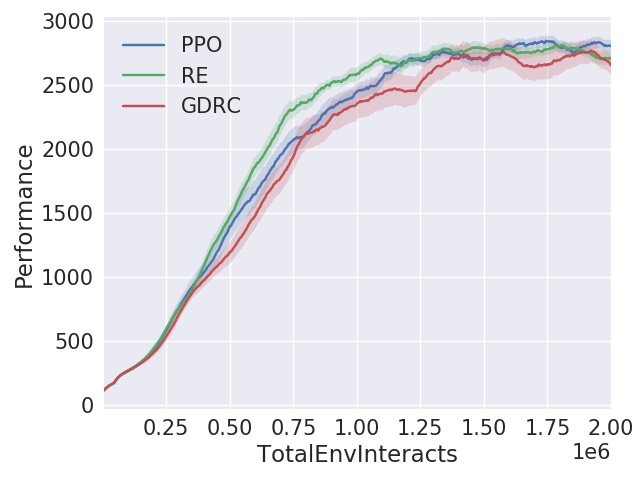}
        \label{fig:image1}
      \end{subfigure}
      \hspace{0.03\textwidth}
      \begin{subfigure}{0.2\textwidth}
        \centering
        \includegraphics[width=1.4\linewidth]{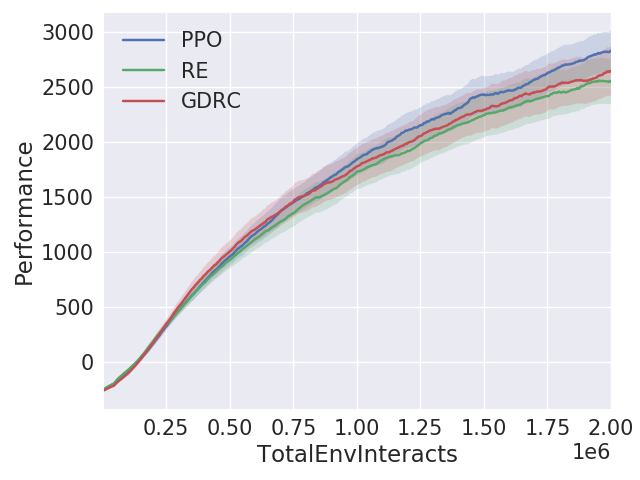}
        \label{fig:image2}
      \end{subfigure}
      \hspace{0.03\textwidth}
      \begin{subfigure}{0.2\textwidth}
        \centering
        \includegraphics[width=1.4\linewidth]{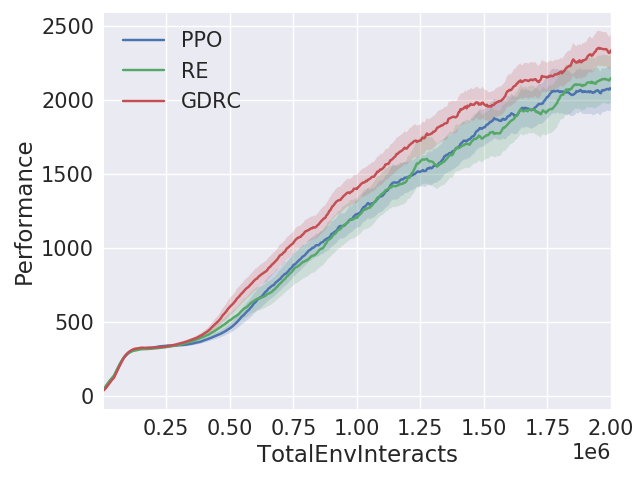}
        \label{fig:image3}
      \end{subfigure}
      \hspace{0.03\textwidth}
      \begin{subfigure}{0.2\textwidth}
        \centering
        \includegraphics[width=1.4\linewidth]{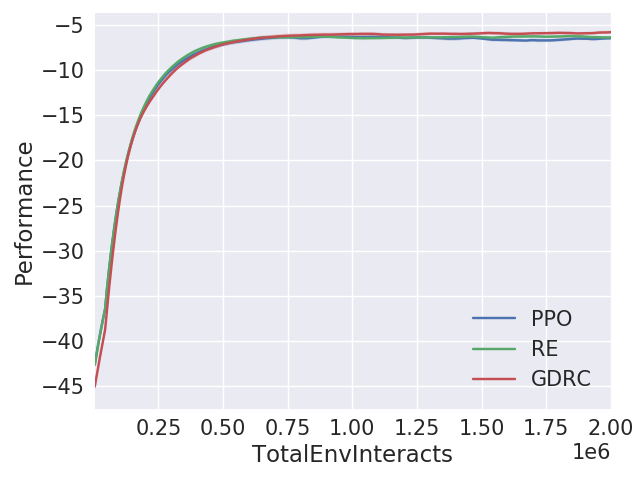}
        \label{fig:image4}
      \end{subfigure}
  \end{minipage}

  \begin{minipage}{0.05\textwidth}
        \centering
        \rotatebox{90}{$\omega=0.2$}
    \end{minipage}%
  \begin{minipage}{0.95\textwidth}
      \begin{subfigure}{0.2\textwidth}
        \centering
        \includegraphics[width=1.4\linewidth]{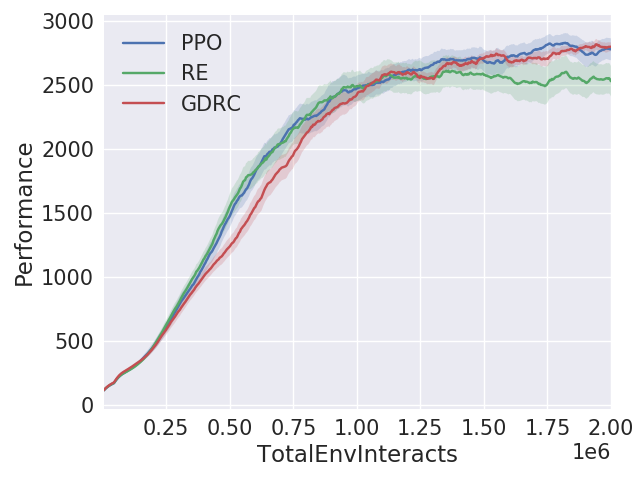}
        \label{fig:image1}
      \end{subfigure}
      \hspace{0.03\textwidth}
      \begin{subfigure}{0.2\textwidth}
        \centering
        \includegraphics[width=1.4\linewidth]{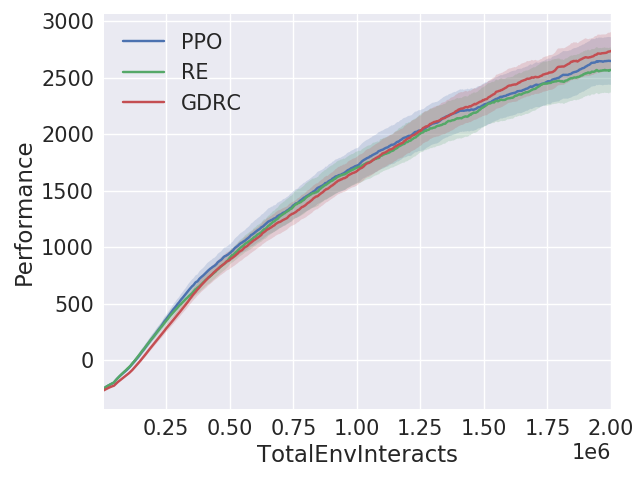}
        \label{fig:image2}
      \end{subfigure}
      \hspace{0.03\textwidth}
      \begin{subfigure}{0.2\textwidth}
        \centering
        \includegraphics[width=1.4\linewidth]{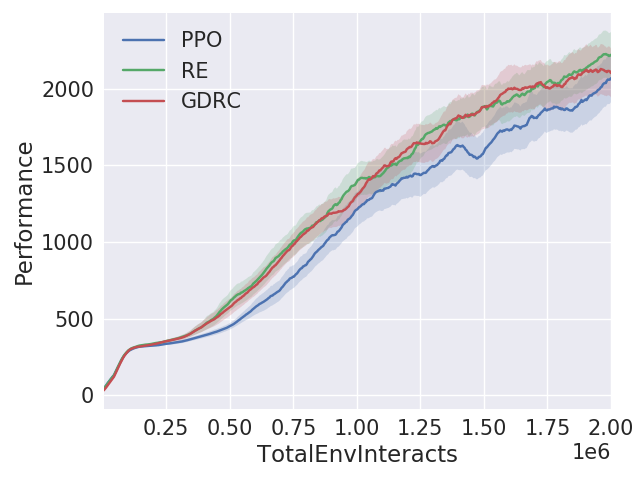}
        \label{fig:image3}
      \end{subfigure}
      \hspace{0.03\textwidth}
      \begin{subfigure}{0.2\textwidth}
        \centering
        \includegraphics[width=1.4\linewidth]{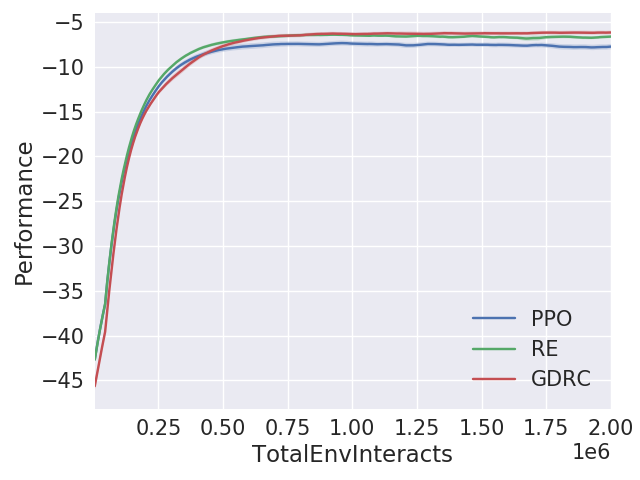}
        \label{fig:image4}
      \end{subfigure}
  \end{minipage}

  \begin{minipage}{0.05\textwidth}
        \centering
        \rotatebox{90}{$\omega=0.3$}
    \end{minipage}%
  \begin{minipage}{0.95\textwidth}
      \begin{subfigure}{0.2\textwidth}
        \centering
        \includegraphics[width=1.4\linewidth]{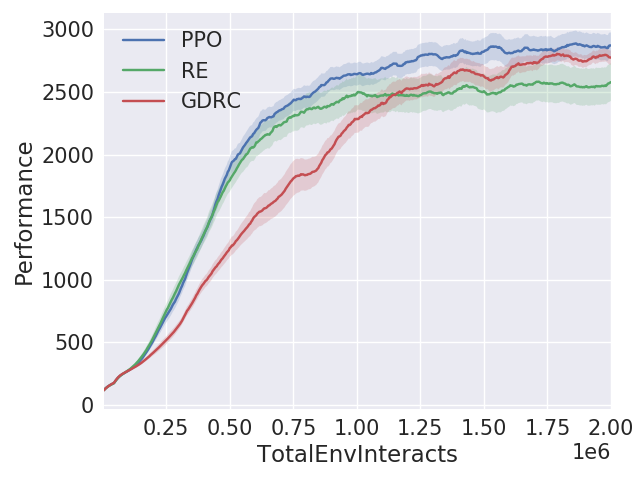}
        \label{fig:image1}
      \end{subfigure}
      \hspace{0.03\textwidth}
      \begin{subfigure}{0.2\textwidth}
        \centering
        \includegraphics[width=1.4\linewidth]{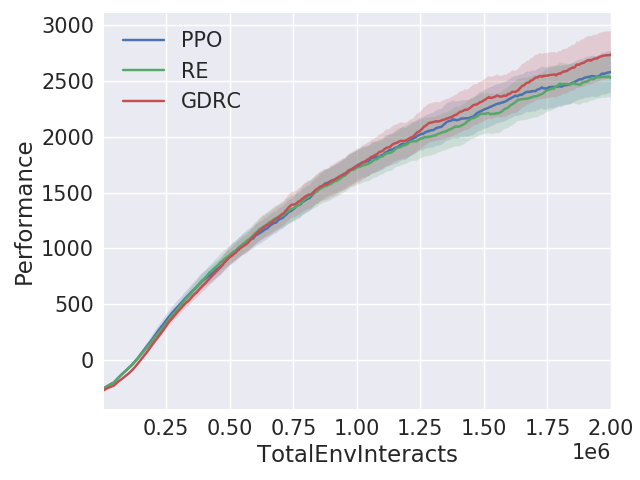}
        \label{fig:image2}
      \end{subfigure}
      \hspace{0.03\textwidth}
      \begin{subfigure}{0.2\textwidth}
        \centering
        \includegraphics[width=1.4\linewidth]{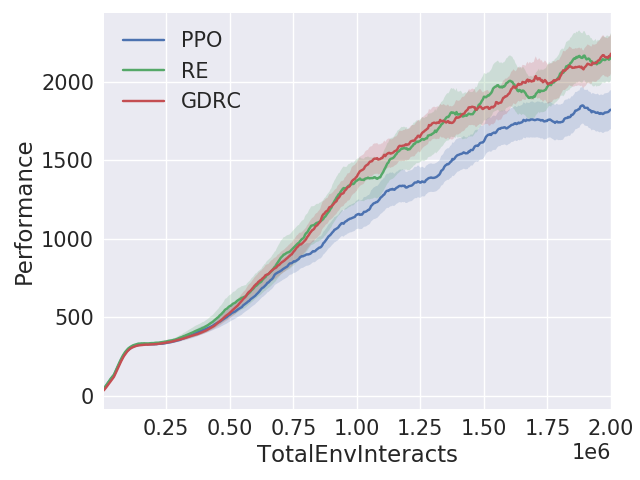}
        \label{fig:image3}
      \end{subfigure}
      \hspace{0.03\textwidth}
      \begin{subfigure}{0.2\textwidth}
        \centering
        \includegraphics[width=1.4\linewidth]{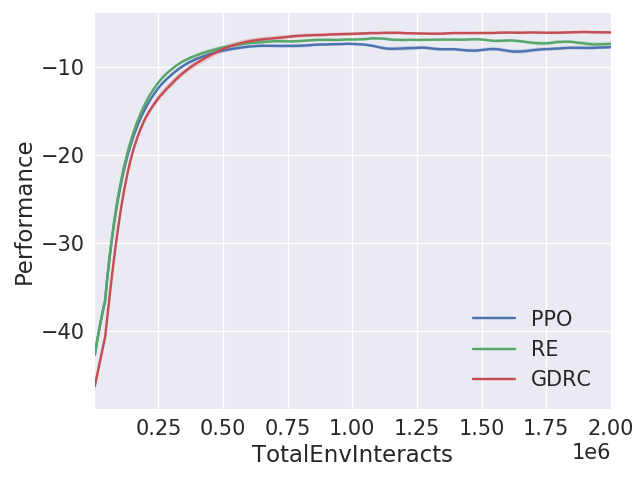}
        \label{fig:image4}
      \end{subfigure}
  \end{minipage}

  \begin{minipage}{0.05\textwidth}
        \centering
        \rotatebox{90}{$\omega=0.4$}
    \end{minipage}%
  \begin{minipage}{0.95\textwidth}
      \begin{subfigure}{0.2\textwidth}
        \centering
        \includegraphics[width=1.4\linewidth]{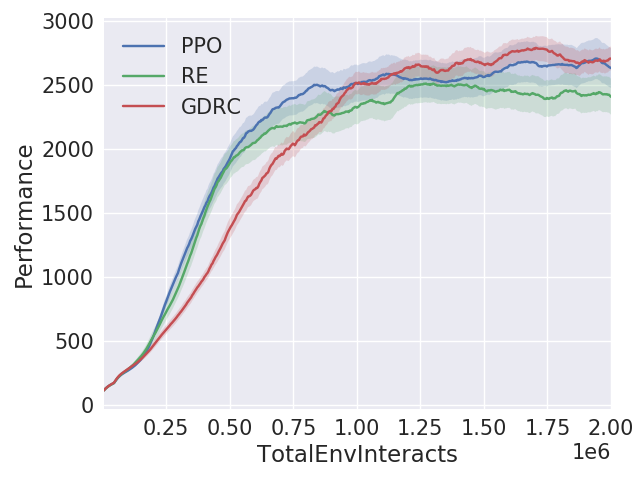}
        \label{fig:image1}
        \caption{Hopper}
      \end{subfigure}
      \hspace{0.03\textwidth}
      \begin{subfigure}{0.2\textwidth}
        \centering
        \includegraphics[width=1.4\linewidth]{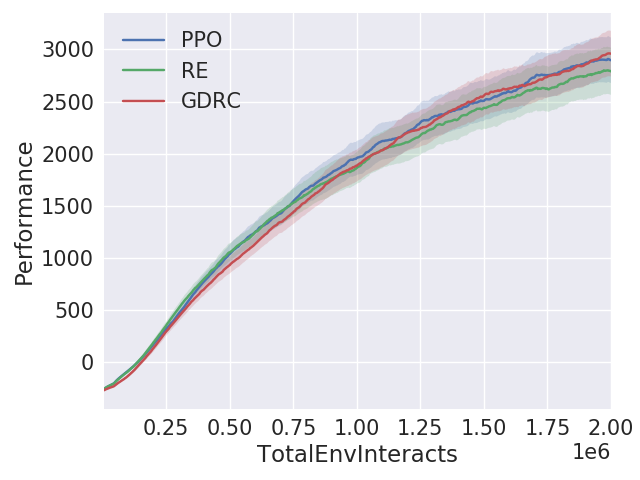}
        \label{fig:image2}
        \caption{HalfCheetah}
      \end{subfigure}
      \hspace{0.03\textwidth}
      \begin{subfigure}{0.2\textwidth}
        \centering
        \includegraphics[width=1.4\linewidth]{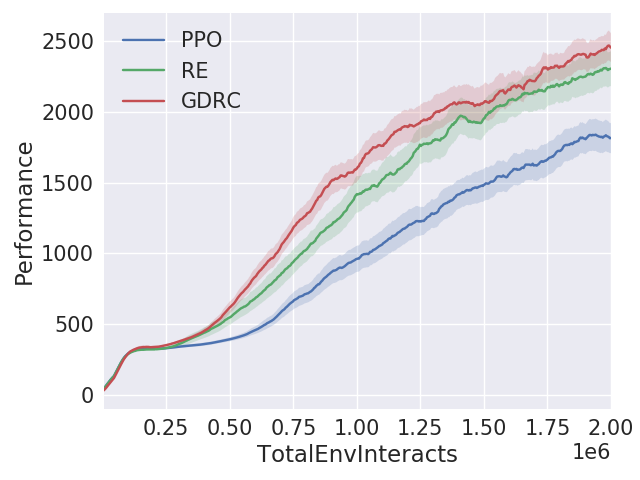}
        \label{fig:image3}
        \caption{Walker2d}
      \end{subfigure}
      \hspace{0.03\textwidth}
      \begin{subfigure}{0.2\textwidth}
        \centering
        \includegraphics[width=1.4\linewidth]{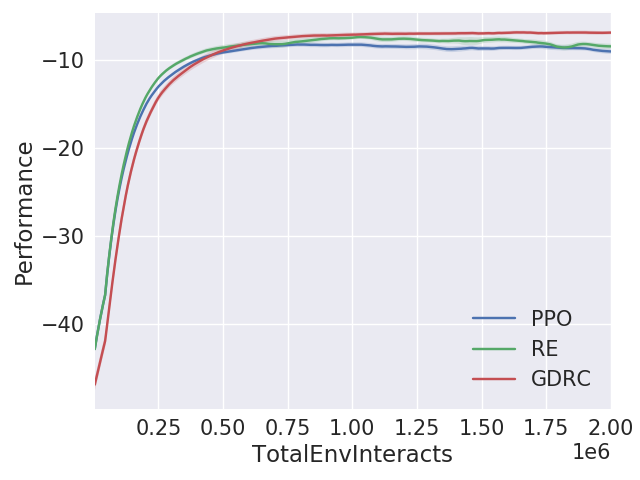}
        \label{fig:image4}
        \caption{Reacher}
      \end{subfigure}
  \end{minipage}

  \caption{Under Uniform perturbations where $\omega=0.1,~0.2,~0.3,~0.4$.}
  \label{appfig:uni2}
\end{figure*}

\begin{figure*}
  \centering
  
    \begin{minipage}{0.05\textwidth}
        \centering
        \rotatebox{90}{$\omega=0.1$}
    \end{minipage}%
  \begin{minipage}{0.95\textwidth}
      \begin{subfigure}{0.2\textwidth}
        \centering
        \includegraphics[width=1.4\linewidth]{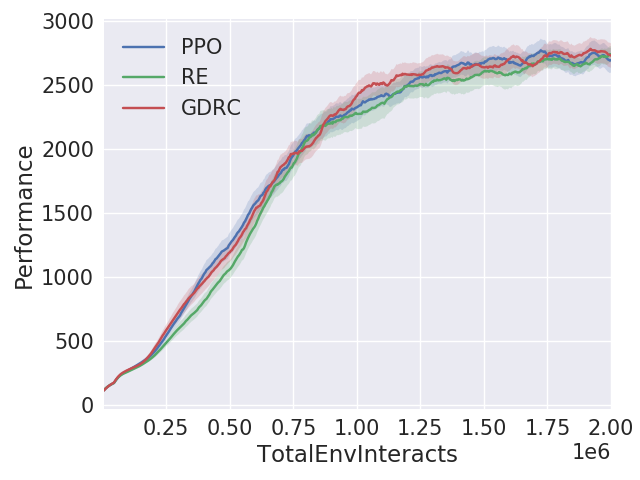}
        \label{fig:image1}
      \end{subfigure}
      \hspace{0.03\textwidth}
      \begin{subfigure}{0.2\textwidth}
        \centering
        \includegraphics[width=1.4\linewidth]{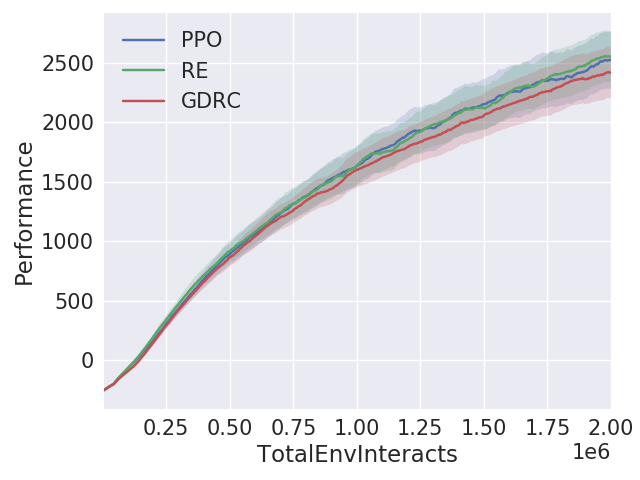}
        \label{fig:image2}
      \end{subfigure}
      \hspace{0.03\textwidth}
      \begin{subfigure}{0.2\textwidth}
        \centering
        \includegraphics[width=1.4\linewidth]{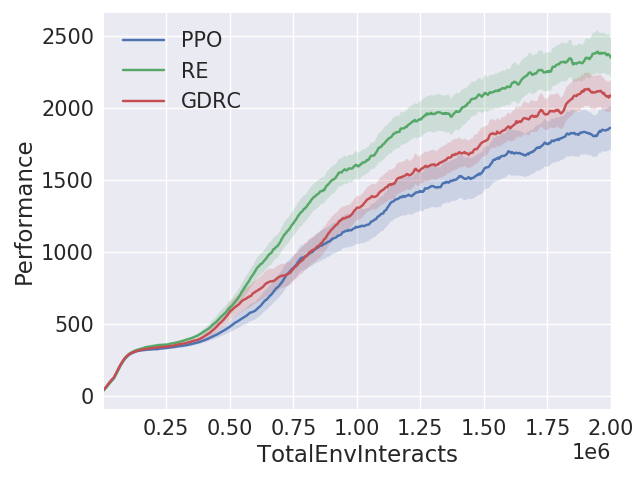}
        \label{fig:image3}
      \end{subfigure}
      \hspace{0.03\textwidth}
      \begin{subfigure}{0.2\textwidth}
        \centering
        \includegraphics[width=1.4\linewidth]{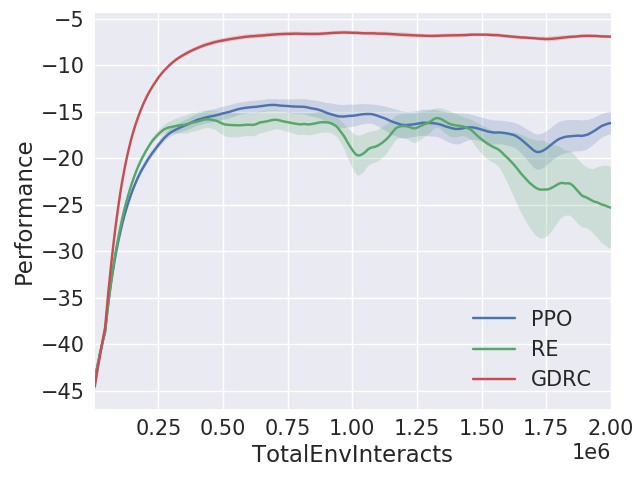}
        \label{fig:image4}
      \end{subfigure}
  \end{minipage}
  
    \begin{minipage}{0.05\textwidth}
        \centering
        \rotatebox{90}{$\omega=0.2$}
    \end{minipage}%
  \begin{minipage}{0.95\textwidth}
      \begin{subfigure}{0.2\textwidth}
        \centering
        \includegraphics[width=1.4\linewidth]{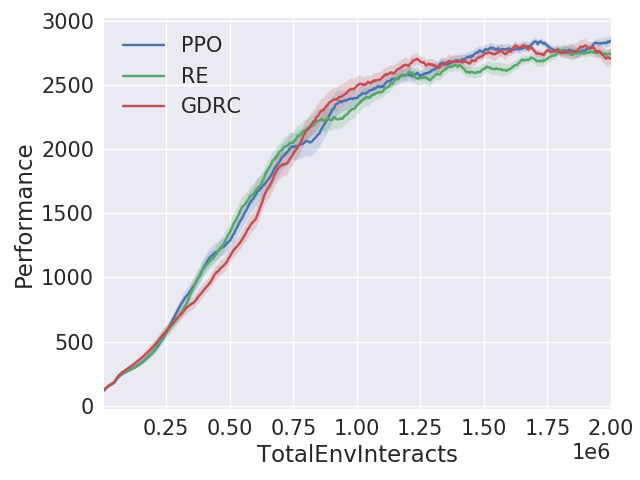}
        \label{fig:image1}
      \end{subfigure}
      \hspace{0.03\textwidth}
      \begin{subfigure}{0.2\textwidth}
        \centering
        \includegraphics[width=1.4\linewidth]{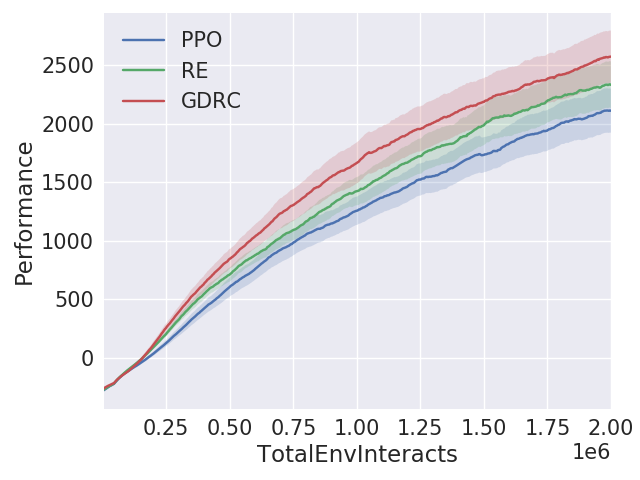}
        \label{fig:image2}
      \end{subfigure}
      \hspace{0.03\textwidth}
      \begin{subfigure}{0.2\textwidth}
        \centering
        \includegraphics[width=1.4\linewidth]{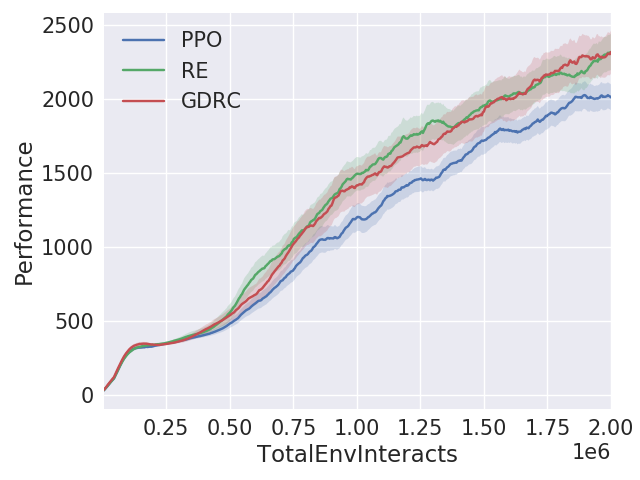}
        \label{fig:image3}
      \end{subfigure}
      \hspace{0.03\textwidth}
      \begin{subfigure}{0.2\textwidth}
        \centering
        \includegraphics[width=1.4\linewidth]{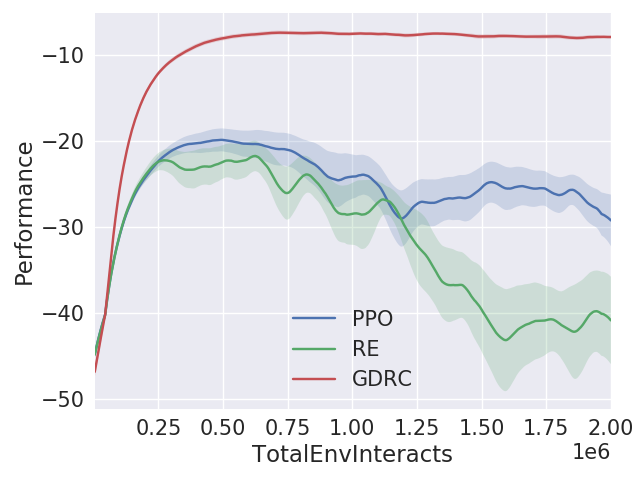}
        \label{fig:image4}
      \end{subfigure}
  \end{minipage}

    \begin{minipage}{0.05\textwidth}
        \centering
        \rotatebox{90}{$\omega=0.3$}
    \end{minipage}%
  \begin{minipage}{0.95\textwidth}
      \begin{subfigure}{0.2\textwidth}
        \centering
        \includegraphics[width=1.4\linewidth]{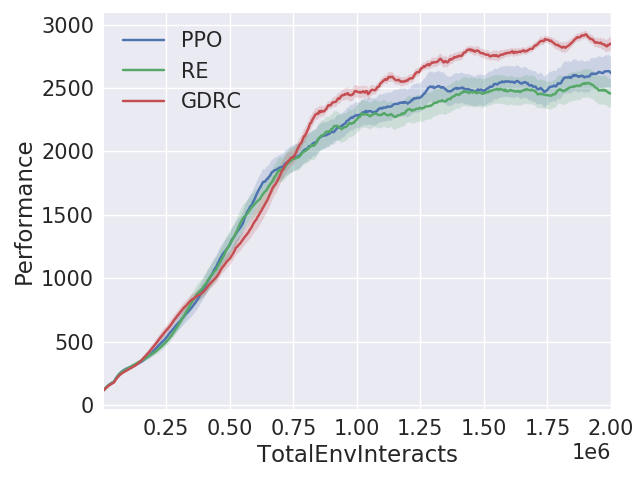}
        \label{fig:image1}
      \end{subfigure}
      \hspace{0.03\textwidth}
      \begin{subfigure}{0.2\textwidth}
        \centering
        \includegraphics[width=1.4\linewidth]{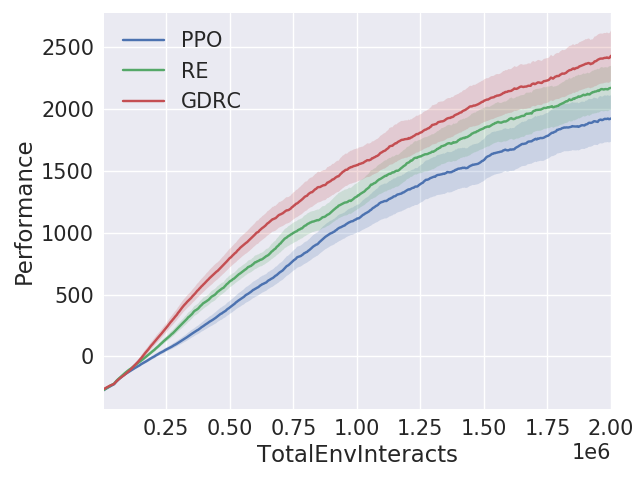}
        \label{fig:image2}
      \end{subfigure}
      \hspace{0.03\textwidth}
      \begin{subfigure}{0.2\textwidth}
        \centering
        \includegraphics[width=1.4\linewidth]{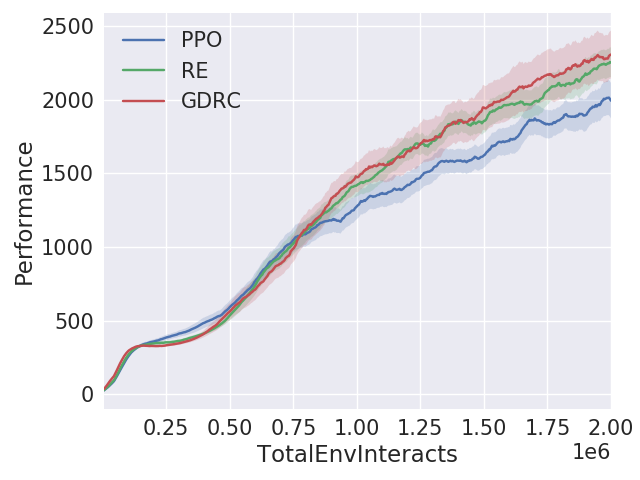}
        \label{fig:image3}
      \end{subfigure}
      \hspace{0.03\textwidth}
      \begin{subfigure}{0.2\textwidth}
        \centering
        \includegraphics[width=1.4\linewidth]{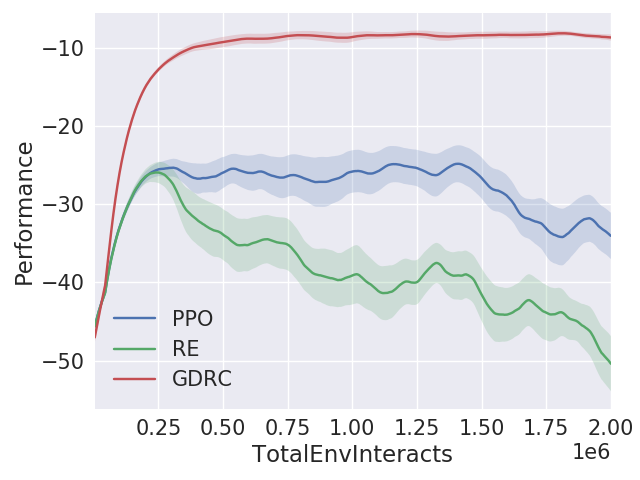}
        \label{fig:image4}
      \end{subfigure}
  \end{minipage}

    \begin{minipage}{0.05\textwidth}
        \centering
        \rotatebox{90}{$\omega=0.4$}
    \end{minipage}%
  \begin{minipage}{0.95\textwidth}
      \begin{subfigure}{0.2\textwidth}
        \centering
        \includegraphics[width=1.4\linewidth]{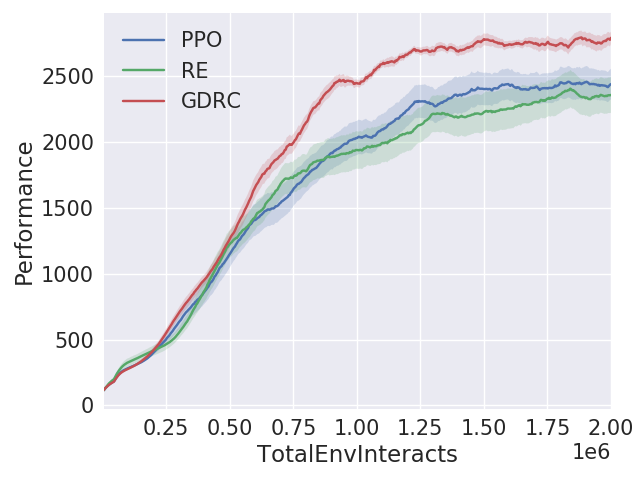}
        \label{fig:image1}
        \caption{Hopper}
      \end{subfigure}
      \hspace{0.03\textwidth}
      \begin{subfigure}{0.2\textwidth}
        \centering
        \includegraphics[width=1.4\linewidth]{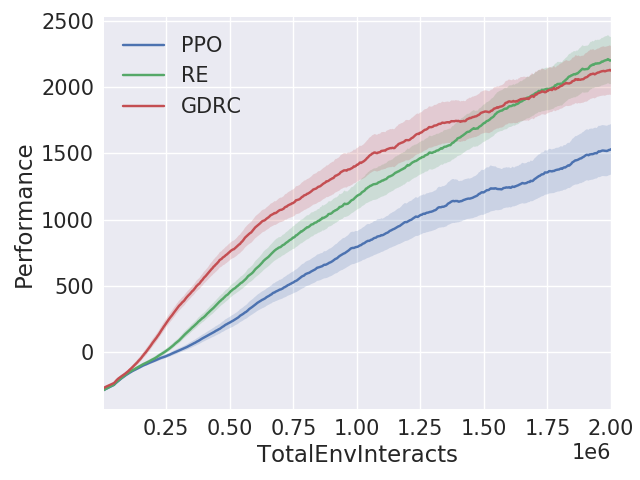}
        \label{fig:image2}
        \caption{HalfCheetah}
      \end{subfigure}
      \hspace{0.03\textwidth}
      \begin{subfigure}{0.2\textwidth}
        \centering
        \includegraphics[width=1.4\linewidth]{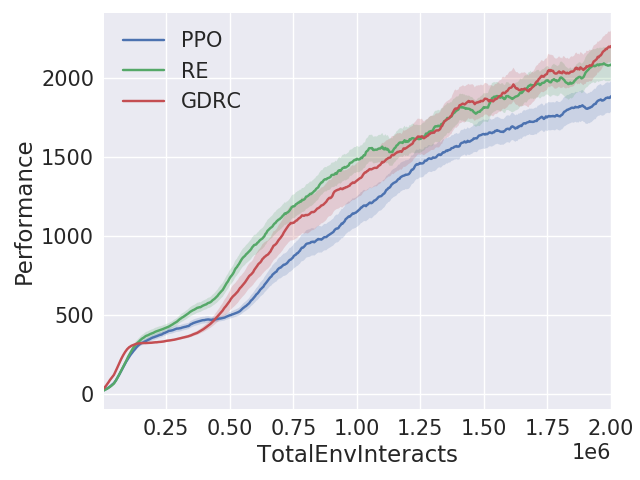}
        \label{fig:image3}
        \caption{Walker2d}
      \end{subfigure}
      \hspace{0.03\textwidth}
      \begin{subfigure}{0.2\textwidth}
        \centering
        \includegraphics[width=1.4\linewidth]{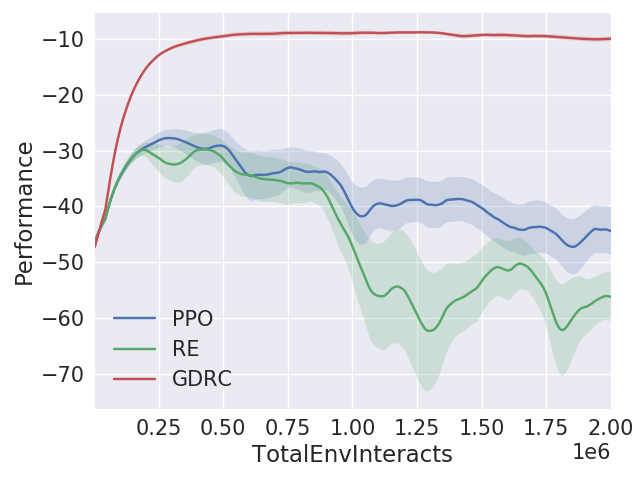}
        \label{fig:image4}
        \caption{Reacher}
      \end{subfigure}
  \end{minipage}

  \caption{Under Reward Range Uniform perturbations where $\omega=0.1,~0.2,~0.3,~0.4$.}
  \label{appfig:uni3}
\end{figure*}

We provide the training curves in Fig.~\ref{appfig:n6}, \ref{appfig:n10}, \ref{appfig:n16}, \ref{appfig:uni1}, \ref{appfig:uni2}, and \ref{appfig:uni3} of the experiments under Generalized Confusion Matrix (GCM) perturbations and continuous perturbations in Mujoco tasks.

\begin{table}[ht]
  \centering
  \caption{Experiement Results in Pendulum Under GCM perturbations. Bolded methods are our methods. Blue methods can be applied without any information about the perturbations. The values after $+/-$ are standard errors calculated with 50 seeds.}
  \label{table:n1_sr}
  \begin{adjustbox}{width=0.7\columnwidth}
  \small
  \begin{tabular}{c c c c c }
    \hline
    $\mathbf{\omega}$ & \textbf{Method} & \textbf{Pendulum}\\
    \hline
                   & \textcolor{blue}{DDPG} &  -157.0 +/- 3.5 \\
                   & \textcolor{blue}{RE}   &  -158.8 +/- 6.0 \\
     $\omega =0.3$ & \textcolor{blue}{\textbf{GDRC}} & -149.8 +/- 2.7  \\
                   & \textbf{DRC}  &  -146.5 +/- 2.4 \\
                   & SR & -154.0 +/- 3.3  \\
                   & SR\_W  &  -147.3 +/- 2.8 \\
    \hline
                   & \textcolor{blue}{DDPG} & -166.4 +/- 7.6  \\
                   & \textcolor{blue}{RE}   & -216.7 +/- 29.7  \\
     $\omega =0.5$ & \textcolor{blue}{\textbf{GDRC}} &  -147.5 +/- 2.5 \\
                   & \textbf{DRC}  & -153.8 +/- 2.9  \\
                   & SR & -167.8 +/- 11.3  \\
                   & SR\_W  & -188.4 +/- 17.0  \\
    \hline
                   & \textcolor{blue}{DDPG} & -322.8 +/- 34.4  \\
                   & \textcolor{blue}{RE}   & -444.1 +/- 54.1  \\
     $\omega =0.7$ & \textcolor{blue}{\textbf{GDRC}} & -177.8 +/- 26.0 \\
                   & \textbf{DRC}  & -149.3 +/- 2.9 \\
                   & SR & -347.8 +/- 37.1  \\
                   & SR\_W  & -291.2 +/- 31.5  \\
    \hline
                   & \textcolor{blue}{DDPG} & -786.0 +/- 62.8  \\
                   & \textcolor{blue}{RE}   & -862.4 +/- 61.8 \\
     $\omega =0.9$ & \textcolor{blue}{\textbf{GDRC}} & -600.2 +/- 75.1  \\
                   & \textbf{DRC}  &  -438.2 +/- 59.1 \\
                   & SR & -942.9 +/- 52.8  \\
                   & SR\_W  &  -1192.2 +/- 43.5 \\
    \hline
  \end{tabular}
  \end{adjustbox}
\end{table}

\begin{table}[ht]
  \centering
  \caption{Experiement Results in CartPole Under GCM perturbations. Bolded methods are our methods. Blue methods can be applied without any information about the perturbations. The values after $+/-$ are standard errors calculated with 50 seeds.}
  \label{table:n2_sr}
  \begin{adjustbox}{width=0.7\columnwidth}
  \small
  \begin{tabular}{c c c c c }
    \hline
    $\mathbf{\omega}$ & \textbf{Method} & \textbf{CartPole}\\
    \hline
                   & \textcolor{blue}{DQN} & 394 +/- 26  \\
                   & \textcolor{blue}{RE}  & 372 +/- 26  \\
     $\omega =0.1$ & \textcolor{blue}{\textbf{DRC}} &  455 +/- 18 \\
                   & SR  &  340 +/- 26 \\
                   & SR\_W &  264 +/- 26 \\
    \hline
                   & \textcolor{blue}{DQN} &  313 +/- 24 \\
                   & \textcolor{blue}{RE}   & 436 +/- 18  \\
     $\omega =0.2$ & \textcolor{blue}{\textbf{DRC}} & 455 +/- 18  \\
                   & SR  & 364 +/- 24  \\
                   & SR\_W &  382 +/- 11 \\
    \hline
                   & \textcolor{blue}{DQN} &  278 +/- 26 \\
                   & \textcolor{blue}{RE}   & 385 +/- 22  \\
     $\omega =0.3$ & \textcolor{blue}{\textbf{DRC}} & 435 +/- 26  \\
                   & SR  &  324 +/- 28 \\
                   & SR\_W & 352 +/- 26 \\
    \hline
                   & \textcolor{blue}{DQN} &  252 +/- 24 \\
                   & \textcolor{blue}{RE}   & 285 +/- 26  \\
     $\omega =0.4$ & \textcolor{blue}{\textbf{DRC}} & 332 +/- 26  \\
                   & SR  &  252 +/- 28 \\
                   & SR\_W & 228 +/- 26  \\
    \hline
  \end{tabular}
  \end{adjustbox}
\end{table}

\FloatBarrier

\section{Time Complexity}
\label{appendix:g}

\begin{table}[ht]
\caption{The time required per seed for each method built on top of PPO when running in Hopper, HalfCheetah, Walker2d, and Reacher, using an Intel Xeon E5-2680 v4 processor.}
\label{table:time}
\centering
\begin{tabular}{c|ccccc} \toprule
    Method      & PPO & RE & SR/SR\_W &  DRC & GDRC \\ \midrule
    Time (min)  & 26 & 30 & 28 & 30 & 34 \\ \bottomrule
\end{tabular}
\end{table}

\begin{table*}[h]
  \centering
  \caption{Experiement Results in Mujoco tasks Under GCM perturbations. Bolded methods are our methods. Blue methods can be applied without any information about the perturbations. The values after $+/-$ are standard errors calculated with 50 seeds.}
  \label{table:n1_2C}
  \begin{adjustbox}{width=1.7\columnwidth,center}
  \small
  \begin{tabular}{c c c c c c c }
    \hline
    \textbf{$n_r$} & $\mathbf{\omega}$ & \textbf{Method} & \textbf{Hopper} & \textbf{HalfCheetah} & \textbf{Walker2d} & \textbf{Reacher}\\
    \hline
             &               & \textcolor{blue}{PPO}           & 2699.0 +/- 76.3&	2534.5 +/- 138.9&	1973.3 +/- 71.8&	-15.8 +/- 0.8  \\
             &               & \textcolor{blue}{RE}            & 2782.6 +/- 51.9&	2735.8 +/- 142.5&	2312.6 +/- 89.4	&   -25.2 +/- 3.7 \\
             & $\omega =0.1$ & \textcolor{blue}{\textbf{GDRC}} & 2768.9 +/- 50.7&	2514.2 +/- 127.8&	2169.7 +/- 88.7&	-9.1 +/- 0.3  \\
             &               & SR\_W                           & 2644.2 +/- 62.2&	1644.5 +/- 152.8&	1627.0 +/- 92.6&	-21.1 +/- 1.0 \\
             &               & \textbf{DRC}                    & 2786.9 +/- 60.9&	2396.9 +/- 129.4&	2116.2 +/- 94.1&	-5.1 +/- 0.1  \\
    \cline{2-7}
             &               & \textcolor{blue}{PPO}           & 2483.6 +/- 78.6&	1552.3 +/- 145.6&	1825.6 +/- 82.8&	-27.1 +/- 2.0  \\
             &               & \textcolor{blue}{RE}            & 2308.5 +/- 78.1&	2026.0 +/- 149.0&	1960.5 +/- 69.1&	-52.8 +/- 4.9 \\
             & $\omega =0.3$ & \textcolor{blue}{\textbf{GDRC}} & 2740.5 +/- 78.4&	2612.4 +/- 128.1&	2107.6 +/- 98.5&	-12.4 +/- 0.6 \\
             &               & SR\_W                           & 2768.6 +/- 64.6&	920.0 +/- 150.9&	1263.9 +/- 63.0&	-24.4 +/- 1.7 \\
    \multirow{2}{*}{$n_r=6$}  &   & \textbf{DRC}               & 2763.8 +/- 61.6&	2333.8 +/- 120.2&	2345.8 +/- 94.8&	-5.2 +/- 0.1  \\
    \cline{2-7}
             &               & \textcolor{blue}{PPO}           & 1731.5 +/- 103.8&	821.4 +/- 86.5&	1288.5 +/- 85.8&	-50.9 +/- 5.8 \\
             &               & \textcolor{blue}{RE}            & 1654.5 +/- 90.1&	1257.5 +/- 120.8&	1377.9 +/- 102.2&	-74.6 +/- 6.9\\
             & $\omega =0.5$ & \textcolor{blue}{\textbf{GDRC}} & 2707.5 +/- 45.2&	2392.0 +/- 129.9&	2061.5 +/- 103.0&	-18.7 +/- 1.7  \\
             &               & SR\_W                           & 2746.2 +/- 68.1&	290.5 +/- 87.6&	1000.5 +/- 37.5&	-23.5 +/- 1.5\\
             &               & \textbf{DRC}                    & 2722.9 +/- 56.1&	2154.6 +/- 125.6&	2186.7 +/- 96.2&	-6.5 +/- 0.2\\
    \cline{2-7}
             &               & \textcolor{blue}{PPO}           & 1333.7 +/- 61.3&	-99.3 +/- 123.6&	848.0 +/- 74.0&	-72.2 +/- 7.2  \\
             &               & \textcolor{blue}{RE}            & 1318.5 +/- 61.2&	-152.2 +/- 170.6&	1021.2 +/- 100.2&	-119.6 +/- 11.9  \\
             & $\omega =0.7$ & \textcolor{blue}{\textbf{GDRC}} & 1935.7 +/- 116.3&	539.8 +/- 150.2&	967.0 +/- 79.3&	-40.1 +/- 8.6 \\
             &               & SR\_W                           &  1318.7 +/- 141.2&	-138.7 +/- 21.2&	500.7 +/- 32.9&	-39.1 +/- 6.0 \\
             &               & \textbf{DRC}                    &  1966.0 +/- 99.2&	-3.0 +/- 144.3&	976.0 +/- 100.2&	-26.3 +/- 4.0\\
    \hline
             &               & \textcolor{blue}{PPO}           & 2709.7 +/- 71.7&	2522.2 +/- 131.8&	2030.4 +/- 87.0&	-17.0 +/- 1.0 \\
             &               & \textcolor{blue}{RE}            & 2706.9 +/- 71.0&	2624.8 +/- 130.9&	2310.8 +/- 93.3&	-22.6 +/- 1.4  \\
             & $\omega =0.1$ & \textcolor{blue}{\textbf{GDRC}} & 2842.8 +/- 49.0&	2527.9 +/- 134.2&	2097.4 +/- 89.8& -8.2 +/- 0.2 \\
             &               & SR\_W                           & 2644.7 +/- 60.4&	2195.5 +/- 132.6&	1935.0 +/- 76.6&	-16.1 +/- 0.6  \\
             &               & \textbf{DRC}                    & 2628.9 +/- 75.2&	2738.8 +/- 133.9&	1997.0 +/- 84.6&	--5.1 +/- 0.1  \\
    \cline{2-7}
             &               & \textcolor{blue}{PPO}           & 2623.2 +/- 68.5&	1803.1 +/- 136.4&	1986.8 +/- 89.5&	-33.4 +/- 2.8  \\
             &               & \textcolor{blue}{RE}            & 2486.4 +/- 75.3&	2211.2 +/- 138.1&	2199.8 +/- 84.1&	-55.1 +/- 5.0  \\
             & $\omega =0.3$ & \textcolor{blue}{\textbf{GDRC}} & 2748.2 +/- 56.7&	2399.2 +/- 129.6&	2124.0 +/- 91.4&	-12.7 +/- 0.6  \\
             &               & SR\_W                           & 2822.8 +/- 46.9&	1522.1 +/- 109.7&	1471.9 +/- 91.3&	-18.8 +/- 0.9  \\
    \multirow{2}{*}{$n_r=10$}  &   & \textbf{DRC}              & 2764.8 +/- 54.6&	 2705.4 +/- 126.5&	 2321.4 +/- 94.0&	 -5.1 +/- 0.1  \\
    \cline{2-7}
             &               & \textcolor{blue}{PPO}           & 2188.1 +/- 94.1&	947.9 +/- 109.8&	1463.7 +/- 75.3&	 -42.4 +/- 3.3  \\
             &               & \textcolor{blue}{RE}            & 2103.9 +/- 80.0&	1166.2 +/- 121.2&	1726.6 +/- 94.6&	 -77.8 +/- 5.8 \\
             & $\omega =0.5$ & \textcolor{blue}{\textbf{GDRC}} & 2663.9 +/- 69.1&	2259.4 +/- 135.9&	2170.2 +/- 90.3&	-20.5 +/- 1.6 \\
             &               & SR\_W                           & 2631.5 +/- 83.2&	763.3 +/- 87.5&	     884.1 +/- 69.6&	-19.0 +/- 1.1 \\
             &               & \textbf{DRC}                    & 2686.8 +/- 69.6&	2278.5 +/- 122.5&	2237.5 +/- 86.3&	-5.4 +/- 0.1  \\
    \cline{2-7}
             &               & \textcolor{blue}{PPO}           & 1447.8 +/- 77.4&	102.2 +/- 87.8&	1010.8 +/- 61.4& -73.8 +/- 5.9	\\
             &               & \textcolor{blue}{RE}            & 1379.9 +/- 75.0&	149.7 +/- 112.8&	1050.5 +/- 74.8&    -126.0 +/- 20.6 \\
             & $\omega =0.7$ & \textcolor{blue}{\textbf{GDRC}} & 2438.4 +/- 80.4&	1618.0 +/- 136.7&	1848.5 +/- 89.4&     -40.0 +/- 7.0\\
             &               & SR\_W                           & 2338.3 +/- 117.6&	133.8 +/- 41.6&	     409.0 +/- 9.8&	 -27.1 +/- 2.1 \\
             &               & \textbf{DRC}                    & 2714.4 +/- 46.2&	1372.5 +/- 139.7&	1959.6 +/- 90.4&	-11.8 +/- 2.4 \\
    \hline
             &               & \textcolor{blue}{PPO}           & 2762.5 +/- 40.7&	2390.8 +/- 106.5& 	2117.9 +/- 79.9&	-18.4 +/- 1.1 \\
             &               & \textcolor{blue}{RE}            & 2736.8 +/- 52.3&	2529.0 +/- 125.4&	2251.7 +/- 93.7&	-27.1 +/- 2.9 \\
             & $\omega =0.1$ & \textcolor{blue}{\textbf{GDRC}} & 2697.4 +/- 61.1&	2567.2 +/- 140.9&	2292.1 +/- 80.4&	-8.6 +/- 0.3  \\
             &               & SR\_W                           & 2731.6 +/- 59.3&	2378.8 +/- 131.1&	1988.7 +/- 79.9&	-13.6 +/- 0.4\\
             &               & \textbf{DRC}                    & 2704.3 +/- 63.3&	2512.6 +/- 135.3&	2168.5 +/- 101.4&	-5.4 +/- 0.1  \\
    \cline{2-7}
             &               & \textcolor{blue}{PPO}           & 2582.8 +/- 67.8&	1814.6 +/- 118.8&	1989.3 +/- 83.6&	-31.8 +/- 2.1  \\
             &               & \textcolor{blue}{RE}            & 2553.7 +/- 62.2&	2151.9 +/- 123.6&	2151.5 +/- 71.0&	-69.3 +/- 10.2 \\
             & $\omega =0.3$ & \textcolor{blue}{\textbf{GDRC}} & 2698.1 +/- 61.0&	2554.4 +/- 134.6&	2204.0 +/- 100.5&	-11.5 +/- 0.5 \\
             &               & SR\_W                           &  2709.9 +/- 56.4&	1886.6 +/- 144.4&	1550.1 +/- 85.0&	-14.7 +/- 0.4 \\
    \multirow{2}{*}{$n_r=16$}&    & \textbf{DRC}               &  2811.2 +/- 53.7&	2443.3 +/- 136.8&	2243.2 +/- 113.0&	-5.6 +/- 0.2 \\
    \cline{2-7}
             &               & \textcolor{blue}{PPO}           & 2126.0 +/- 93.3&	833.7 +/- 110.0&	1453.2 +/- 72.4&	-49.7 +/- 3.9 \\
             &               & \textcolor{blue}{RE}            & 2146.0 +/- 81.9&	1187.2 +/- 136.8&	1499.7 +/- 79.5&	-78.0 +/- 5.7 \\
             & $\omega =0.5$ & \textcolor{blue}{\textbf{GDRC}} & 2766.6 +/- 62.0&	2496.1 +/- 138.1&	2140.1 +/- 92.7&	-15.4 +/- 0.8  \\
             &               & SR\_W                           & 2649.3 +/- 82.6&	945.7 +/- 93.2&	947.6 +/- 72.9&	-17.4 +/- 1.1 \\
             &               & \textbf{DRC}                    & 2806.0 +/- 51.4&	2345.1 +/- 126.5&	2051.5 +/- 92.1&	-5.5 +/- 0.1 \\
    \cline{2-7}
             &               & \textcolor{blue}{PPO}           & 1570.4 +/- 79.3&	165.3 +/- 62.5&	1077.1 +/- 48.4&	-69.8 +/- 5.8 \\
             &               & \textcolor{blue}{RE}            & 1404.5 +/- 68.4&	164.1 +/- 72.4&	1176.4 +/- 59.7&	-117.7 +/- 11.8 \\
             & $\omega =0.7$ & \textcolor{blue}{\textbf{GDRC}} & 2707.0 +/- 64.6&	2053.6 +/- 133.2&	2042.7 +/- 107.8&	-24.0 +/- 1.7 \\
             &               & SR\_W                           & 2493.8 +/- 111.2&	   184.1 +/- 35.5&	493.0 +/- 31.6&	-22.1 +/- 2.3 \\
             &               & \textbf{DRC}                    & 2773.8 +/- 66.4&	2249.6 +/- 123.1&	2100.0 +/- 88.4&	-7.0 +/- 0.2 \\
    \hline
  \end{tabular}
  \end{adjustbox}
\end{table*}
\FloatBarrier

\begin{table*}[h]
  \centering
  \caption{Experiement Results in Mujoco tasks Under continuous perturbations. Bolded methods are our methods. The values after $+/-$ are standard errors calculated with 20 seeds.}
  \label{table:n2_continuous}
  \begin{adjustbox}{center}
  \small
  \begin{tabular}{c c c c c c c }
    \hline
    \textbf{Perturbation} & $\mathbf{\sigma/\omega}$ & \textbf{Method} & \textbf{Hopper} & \textbf{HalfCheetah} & \textbf{Walker2d} & \textbf{Reacher}\\
    \hline
     & \multirow{3}{*}{$\sigma =1.0$} & \textcolor{blue}{PPO} & 2702.2 +/- 81.2 &  3096.5 +/- 252.5&  2116.5 +/- 109.1&  -9.2 +/- 0.3\\
    & & \textcolor{blue}{RE} & 2577.0 +/- 117.9 &  2912.6 +/- 213.9& 2096.8 +/- 139.8& -23.0 +/- 6.1\\
    & & \textcolor{blue}{\textbf{GDRC}} & 2829.0 +/- 64.4 &  2821.4 +/- 212.8&  2218.8 +/- 144.9&  -10.9 +/- 0.6\\
    \cline{2-7}
    & \multirow{3}{*}{$\sigma =1.5$} & \textcolor{blue}{PPO} &  2716.9 +/- 109.5&  2659.7 +/- 255.8& 2096.8 +/- 169.4&	-10.9 +/- 0.7\\
    Gaussian & & \textcolor{blue}{RE} & 2755.0 +/- 56.7&	2881.3 +/- 185.7& 2187.1 +/- 118.9&	 -12.7 +/- 0.9\\
    & & \textcolor{blue}{\textbf{GDRC}} & 2793.5 +/- 102.2&	2743.6 +/- 213.5& 2069.4 +/- 135.3&	 -11.0 +/- 0.3\\
    \cline{2-7}
     & \multirow{3}{*}{$\sigma =2.0$} & \textcolor{blue}{PPO} & 2678.5 +/- 127.0 &	2452.7 +/- 216.6&  1933.0 +/- 158.0&   -10.3 +/- 0.5\\
    & & \textcolor{blue}{RE} & 2794.0 +/- 70.9&	2820.9 +/- 194.1&	1969.2 +/- 112.7& -15.2 +/- 2.8\\
    & & \textcolor{blue}{\textbf{GDRC}} & 2846.1 +/- 48.2&  2831.6 +/- 187.6&	2015.3 +/- 148.9&	-10.9 +/- 0.3\\
    \hline
    \multirow{12}{*}{Uniform} & \multirow{3}{*}{$\omega =0.1$} & \textcolor{blue}{PPO} & 2858.9 +/- 37.8&	2888.7 +/- 188.7&	2211.6 +/- 157.4&	-6.5 +/- 0.2\\
    & & \textcolor{blue}{RE} & 2705.1 +/- 63.6&	2569.7 +/- 213.5& 2134.9 +/- 156.2&	-6.5 +/- 0.2\\
    & & \textcolor{blue}{\textbf{GDRC}} & 2651.9 +/- 80.2&	2678.1 +/- 217.2&	2044.6 +/- 134.5&	-5.7 +/- 0.2\\
    \cline{2-7}
    & \multirow{3}{*}{$\omega =0.2$} & \textcolor{blue}{PPO} &  2680.2 +/- 98.8&	2672.3 +/- 215.1&	2017.3 +/- 149.0&	-7.6 +/- 0.3\\
    & & \textcolor{blue}{RE} &  2588.3 +/- 125.7&	2601.6 +/- 212.2&	2319.0 +/- 179.2&	-6.5 +/- 0.2\\
    & & \textcolor{blue}{\textbf{GDRC}} &  2846.7 +/- 58.9&	2783.4 +/- 176.2&	2255.0 +/- 121.2&	-6.0 +/- 0.1\\
    \cline{2-7}
     & \multirow{3}{*}{$\omega =0.3$} & \textcolor{blue}{PPO} &  2824.0 +/- 106.7&	2625.2 +/- 185.3&	1815.0 +/- 134.9&	-7.7 +/- 0.2 \\
    & & \textcolor{blue}{RE} &  2565.5 +/- 156.3&	2638.6 +/- 203.2&	2188.9 +/- 155.7&	-7.5 +/- 0.4 \\
    & & \textcolor{blue}{\textbf{GDRC}} &  2782.8 +/- 78.0&	2758.0 +/- 228.4&	2025.5 +/- 146.1&	-6.0 +/- 0.2  \\
    \cline{2-7}
     & \multirow{3}{*}{$\omega =0.4$} & \textcolor{blue}{PPO} &  2612.4 +/- 160.2&	2928.0 +/- 223.6&	1776.2 +/- 107.7&	-9.0 +/- 0.4\\
    & & \textcolor{blue}{RE} & 2442.2 +/- 143.5&	2824.7 +/- 240.8&	2202.4 +/- 151.2&	-8.4 +/- 0.4 \\
    & & \textcolor{blue}{\textbf{GDRC}} &   2751.2 +/- 76.3&	2915.6 +/- 217.4&	2223.5 +/- 148.3&	-6.4 +/- 0.2 \\
    \hline
     \multirow{12}{*}{} & \multirow{3}{*}{$\omega =0.1$} & \textcolor{blue}{PPO} &  2723.5 +/- 96.2 & 2530.2 +/- 241.5&	1897.3 +/- 175.5&	-16.1 +/- 1.2 \\
    & & \textcolor{blue}{RE} & 2733.1 +/- 58.7&	2533.9 +/- 210.1&	2287.4 +/- 113.4&	-25.5 +/- 4.1 \\
    & & \textcolor{blue}{\textbf{GDRC}} & 2877.0 +/- 55.1&	2410.2 +/- 185.5&	2303.2 +/- 169.6	& -6.7 +/- 0.2  \\
    \cline{2-7}
    & \multirow{3}{*}{$\omega =0.2$} & \textcolor{blue}{PPO} & 2801.4 +/- 58.9&	2134.6 +/- 191.4&	1887.2 +/- 91.7& -30.8 +/- 4.0 \\
    & & \textcolor{blue}{RE} &  2742.5 +/- 90.7&	2377.8 +/- 200.1&	2307.8 +/- 142.1&	-41.6 +/- 5.4 \\
    Reward Range & & \textcolor{blue}{\textbf{GDRC}} &  2793.2 +/- 76.8&	2518.5 +/- 193.2&	2179.5 +/- 162.8&	-7.8 +/- 0.2 \\
    \cline{2-7}
    Uniform & \multirow{3}{*}{$\omega =0.3$} & \textcolor{blue}{PPO} & 2635.8 +/- 128.5&	1945.9 +/- 191.0&	1996.0 +/- 122.7&	-33.9 +/- 3.0 \\
    & & \textcolor{blue}{RE} &  2409.4 +/- 118.6&	2163.7 +/- 171.3&	2217.7 +/- 123.9&	-50.9 +/- 3.9\\
    & & \textcolor{blue}{\textbf{GDRC}} &  2714.5 +/- 83.9&	2439.6 +/- 214.9&	2169.5 +/- 121.1	& -8.8 +/- 0.5\\
    \cline{2-7}
     & \multirow{3}{*}{$\omega =0.4$} & \textcolor{blue}{PPO} & 2433.8 +/- 116.3&	1510.9 +/- 179.0& 1936.3 +/- 96.4&	-44.5 +/- 4.3 \\
    & & \textcolor{blue}{RE} & 2401.2 +/- 137.3&	2218.8 +/- 189.2& 2128.5 +/- 108.4&	-55.8 +/- 4.2 \\
    & & \textcolor{blue}{\textbf{GDRC}} & 2679.1 +/- 105.4&	2627.2 +/- 214.2&	2119.4 +/- 118.2	& -8.7 +/- 0.4 \\
    \hline
  \end{tabular}
  \end{adjustbox}
\end{table*}
\FloatBarrier

\clearpage

\end{document}